\documentclass{ieeeaccess-nologo}
\usepackage{amsmath,amssymb,amsfonts}
\usepackage{algorithmic}
\usepackage{graphicx}
\usepackage{textcomp}
\usepackage{calligra}
\usepackage[numbers]{natbib}
\usepackage[utf8]{inputenc}

\usepackage{amsmath,amssymb, amsthm, bm}
\usepackage{algorithm,algorithmic}
\usepackage{subfigure}
\usepackage{graphicx}
\usepackage{enumerate}
\usepackage{setspace}
\usepackage{hyperref}
 
\usepackage{wrapfig}
\usepackage{comment}
\usepackage{mathtools}
\usepackage{color,soul}
\usepackage{xcolor}         

\newcommand{\Raed}[1]{{\color{red}[Raed: #1]}}
\newtheorem{definition}{Definition}

\def\BibTeX{{\rm B\kern-.05em{\sc i\kern-.025em b}\kern-.08em
    T\kern-.1667em\lower.7ex\hbox{E}\kern-.125emX}}
\begin{document}
\doi{}

\title{The Internet of Federated Things (IoFT): \\ \Large{A Vision for the Future and In-depth Survey of Data-driven Approaches for Federated Learning}}
\author{\uppercase{Raed Kontar\authorrefmark{1},
Naichen Shi\authorrefmark{1}, Xubo Yue\authorrefmark{1}, Seokhyun Chung\authorrefmark{1}, Eunshin Byon\authorrefmark{1}, Mosharaf Chowdhury\authorrefmark{2}, Judy Jin\authorrefmark{1}, Wissam Kontar\authorrefmark{3}, Neda Masoud\authorrefmark{4}, Maher Noueihed\authorrefmark{5}, Chinedum E. Okwudire\authorrefmark{6}, Garvesh Raskutti\authorrefmark{7}, Romesh Saigal\authorrefmark{1}, Karandeep Singh\authorrefmark{8}, and Zhisheng Ye\authorrefmark{9}} 
}
\address[1]{Department of Industrial \& Operations Engineering, University of Michigan, Ann Arbor, 48105}
\address[2]{Electrical Engineering \& Computer Science, University of Michigan, Ann Arbor, 48105}
\address[3]{Civil \& Environmental Engineering, University of Wisconsin Madison, Madison, 53715}
\address[4]{Civil \& Environmental Engineering, University of Michigan, Ann Arbor, 48105}
\address[5]{Industrial Engineering, American University of Beirut, Beirut, 1107 2020}
\address[6]{Mechanical Engineering, University of Michigan, Ann Arbor, 48105}
\address[7]{Statistics, University of Wisconsin Madison, Madison, 53715}
\address[8]{Learning Health Sciences, University of Michigan, Ann Arbor, 48105}
\address[9]{Industrial Systems Engineering \& Management, National University of Singapore, Singapore 119077}
\tfootnote{ ``This work was supported in part by the National Science Foundation under Grant CPS 1931950.''}

\corresp{Corresponding author: Raed Kontar (e-mail: alkontar@umich.edu).}

\begin{abstract}
The Internet of Things (IoT) is on the verge of a major paradigm shift. In the IoT system of the future, IoFT, the “cloud'' will be substituted by the ``crowd'' where model training is brought to the edge, allowing IoT devices to collaboratively extract knowledge and build smart analytics/models while keeping their personal data stored locally. This paradigm shift was set into motion by the tremendous increase in computational power on IoT devices and the recent advances in decentralized and privacy-preserving model training, coined as federated learning (FL). This article provides a vision for IoFT and a systematic overview of current efforts towards realizing this vision. Specifically, we first introduce the defining characteristics of IoFT and discuss FL data-driven approaches, opportunities, and challenges that allow decentralized inference within three dimensions: (i) a global model that maximizes utility across all IoT devices, (ii) a personalized model that borrows strengths across all devices yet retains its own model, (iii) a meta-learning model that quickly adapts to new devices or learning tasks. We end by describing the vision and challenges of IoFT in reshaping different industries through the lens of domain experts. Those industries include manufacturing, transportation, energy, healthcare, quality \& reliability, business, and computing. 
\end{abstract}


\titlepgskip=-15pt

\maketitle
\setcounter{tocdepth}{2}
\itshape

\tableofcontents


\section{Introduction} \label{sec:introduction}
\subsection{\emph{Preamble}} \label{sec:preamble}

At the early stages of the COVID-19 pandemic, companies that mass-produce personal protective equipment (PPE) required long ramp-up times to fulfill the urgent demand \cite{demandsurge,3drapidresponse}. The ramp-up time took longer than expected as supply chains across the globe were critically disrupted, with entire countries in lockdown and essential workers succumbing to the virus \cite{fordcorona}. Realizing this, many citizens and small businesses tried to bridge the supply gap using readily available, and low-cost 3D printers \cite{civilian3d,12yearold}. This attempt at so-called massively distributed manufacturing \cite{MDM} helped fill PPE production gaps to some extent \cite{civilian3d,12yearold}. However, it also revealed critical impediments to realizing massively distributed manufacturing in terms of standardizing production requirements, guaranteeing quality and reliability, and attaining high production efficiencies that can rival those of mass production \cite{MDM}. For example, a large percentage of parts printed by citizens did not meet the quality requirements \cite{fillgap, faultymasks}. Even when following standard 3D printing guidelines, several prints failed \cite{defectivemasks} while others experienced recurrent defects due to the use of models or methods that did not account for the specific environment in which the 3D printer is operating \cite{salmi20203d}. On the other hand, citizens that succeeded struggled to effectively broadcast their improved models or methods to other users to help improve quality across the network of manufacturers \cite{cant3dprint}.  

Now imagine an alternative future based on a cyber-physical operating system for massively distributed manufacturing. All 3D printers are IoT-enabled through wifi and smart sensors. In addition, printers now have computation power through AI chips (many 3D printers nowadays have such capabilities, ex: Raspberry Pi's \cite{baumann2017additive,okwudire2018low}). The printers collaboratively learn a model for 3D printing PPE accurately with the help of a central orchestrator, guiding the production to the desired quality level. To preserve privacy and intellectual property and allow for massive parallelization,  raw data from each 3D printer is never shared with the central server; instead, printers exploit their compute resources at the edge by running small local computations and only sharing the minimal information needed to learn the model. This model, despite having a global state, is personalized to form a local model that accounts for individual-level external factors affecting each 3D printer.

In this alternative reality, responders can 3D print PPE at the desired quality level with little or no defects. Responders act quickly due to the massively parallelized efforts from many 3D printers and the effective utilization of network bandwidth. In addition, with their personalized 3D printing models, the responders are able to push 3D printers at faster speeds to shorten printing time while maintaining quality \cite{UMtech,duan2018limited,okwudire2018low}. Accordingly, the PPE supply gap is successfully filled until mass production ramps up.

In this future, not only manufacturing benefits. Take healthcare wearable devices as an example. Compute power on such devices has been immensely increasing over the years. Now, personal data need not be uploaded to a central cloud system to learn an anomaly detection model for health signals. Instead, the ``cloud'' is replaced by the ``crowd'', where wearable devices store necessary data, perform local computations and send only needed model updates to the central authority. This decouples the ability to learn the model from storing data in the cloud by bringing training to the device as well, where a model can be learned across thousands of millions of wearable devices in geographically dispersed locations.

Let us now switch paradigms and replace smart devices with ``smart'' institutes. Different medical institutions can join efforts and collaboratively learn diagnostic models without directly sharing their electronic health records, as imposed by the Health Insurance Portability and Accountability Act (HIPAA). Now, diagnostic models can leverage largely diverse datasets and promote fairness through a decentralized learning framework that mitigates privacy risks and costs associated with centralized modeling. Learning can be done across institutes and individuals at multiple scales and in areas that this has not been possible or allowed before. 

The future described above is not a far cry away. It has already been set into action as the immediate yet bold next step for the Internet of Things (IoT). It is the cultivation of Industry 4.0. A cultivation of advances in interdisciplinary fields in the past two decades: ranging from data science, edge computing, machine learning, operations research, optimization, data acquisition technologies, physics-guided modeling, and privacy, amongst many others. 


In this article, we term this future of IoT as the \textbf{Internet of Federated Things (IoFT)}. The term ``federated'' refers to some level of internal autonomy of IoT devices and is inspired by the explosive interest during the past two years in \textbf{Federated Learning (FL)}: an approach that allows decentralized and privacy-preserving training of models \citep{fedavg}. With the help of FL, the decentralized paradigm in IoFT exploits edge compute resources in order to enable devices to collaboratively extract knowledge and build smart analytics/models while keeping their personal data stored locally. This paradigm shift not only reduces privacy concerns but also sets forth many intrinsic advantages including cost efficiency, diversity, and reduced computation, amongst many others to be detailed in the following sections.

\subsection{\emph{Purpose and Uniqueness}} \label{sec:overview}
This paper is a joint effort of researchers across a wide variety of expertise to address the three questions below:

 \begin{enumerate}
     \item What are the defining characteristics of IoFT?
     \item What are key recent advances and potential data-driven methods in IoFT that allow learning in one of the three dimensions stated below? what modeling, optimization, and statistical challenges do they face? and what are potential promising solutions?
     \begin{itemize}
         \item \textbf{A Global model}: that maximizes utility across all devices. The global model aims at capturing the commonalities and intrinsic relatedness across data from all devices to improve prediction and learning accuracy. 
         \item \textbf{A Personalized model}: that tries to personalize and adapt the global model to data and external conditions from each device. This embodies the principle of multi-task learning, \citep{pan2009survey} where each device retains its own model while borrowing strength across all IoFT devices.
         \item \textbf{A Meta-learning model}: that learns a global model which can quickly adapt to a new task with only a small amount of training samples and learning steps. This embodies the principle of ``learning to learn fast," \citep{snell2017prototypical} where the goal of the global model is not to perform well on all tasks in expectation, instead to find a good initialization that can directly adapt to a specific task.   
     \end{itemize}
     \item How will IoFT shape different industries and what are the domain specific challenges it faces for it to become the standard practice? Through the lens of domain experts, we shed light on the following sectors: \textbf{manufacturing}, \textbf{transportation}, \textbf{energy}, \textbf{healthcare}, \textbf{quality \& reliability}, \textbf{business} and \textbf{computing}.
 \end{enumerate}

Besides defining the central characteristics of IoFT, our paper's focus is summarized in two folds. The first is \textbf{data-driven modeling} where we categorize FL approaches in IoFT into learning a global, personalized, and meta-learning model and then provide an in-depth analysis on modeling techniques, recent advances, possible alternative, and statistical/optimization challenges. 
The second focus is a \textbf{vision of IoFT}'s potential use cases, application-specific models, and obstacles within different application domains. Our overarching goal is to encourage researchers across different industries to explore the transformation from IoT to IoFT so that critical societal impacts brought by this emerging technology can be fully realized.

We note here that some excellent surveys on FL have been recently released. Most notably, \citet{lim2020federated} address FL challenges in mobile edge networks with a focus on communication cost, privacy and security, \citet{niknam2020federated} discuss FL application in wireless communications, especially under 5G networks,  \citet{li2020federated} provide a thorough overview of implementation challenges in FL, \citet{yang2019federated} then categorize different architectures for FL, \citet{rahman2020survey} discuss the evolution of the deployment architectures with an in-depth discussion on privacy and security, while \citet{aledhari2020federated} highlight necessary protocols and platforms needed for such architectures, \citet{kairouz2019advances} study open problems in FL and recent initiatives while providing a remarkable survey on privacy-preserving mechanisms. Along this line, \citet{lyu2020threats} highlight threats and major attacks in FL. While our focus is on \textbf{data-driven modeling} for IoFT and how various application fields will be affected by the shift from IoT to IoFT, the surveys above serve as excellent complementary work for a bird's eye view of FL and hence IoFT.  

The remainder of this paper is organized as follows. Sec. \ref{sec:iot} highlights the past and  present features of IoT-enabled systems leading to IoFT. Secs. \ref{sec:global} - \ref{sec:metalearning} provide data-driven modeling approaches for learning a global, personalized, and meta-learning model, along with their challenges and promising solutions. Finally, Sec. \ref{sec:statopt} poses central statistical and optimization open problems in IoFT. These open problems are from both a theoretical and applied perspective. Finally, Sec. \ref{sec:applications} provides a vision for IoFT within manufacturing, transportation, energy, healthcare, quality \& reliability, business, and computing. 

\textbf{Throughout this paper, we use IoFT to denote the future IoT system we envision, while FL denotes the underlying data analytics approach for data-driven model learning within IoFT}. Also, edge device, local device, node, user, or client are used interchangeably to denote the end-user based on the problem context. 

\subsection{\emph{I\lowercase{O}FT Website and Central Directory}} \label{sec:website}

While exploring data-driven modeling approaches to FL in IoFT, it became clear that real-life datasets (in engineering, health sciences, etc..) are pressingly needed to fully explore the disruptive potential of IoFT. While few already exist, they are based on artificial examples, and the few non-artificial datasets are mostly focused on mobile applications. However, for IoFT to become a norm in different industries, real-life datasets with defining features of the underlying system are needed to unveil the potential challenges and opportunities faced within different domains. Only with a deep understanding of the underlying system and domain, one formulates the right analytics. Towards this end, this paper features a supplementary website (\url{https://ioft-data.engin.umich.edu/}) managed by the University of Michigan. The website will serve as a central directory for IoFT-based datasets and will feature brief descriptions of each dataset categorized by its respective field with a link to the repository (research lab website, GitHub account, papers, etc..) where the data is contained. Our hope is to provide a means for model validation within different domains, encourage researchers to develop real-life datasets for IoFT and help with the outreach and visibility of their datasets and corresponding papers.  

\begin{center}
    \noindent\fcolorbox{white}[rgb]{0.95,0.95,0.95}{\begin{minipage}{0.8\columnwidth}
		\begin{center}
 Website: \url{https://ioft-data.engin.umich.edu/}
\end{center}
\end{minipage}}
\end{center}

\section{Internet of Things: The Past, Present and Future} \label{sec:iot}

\begin{figure}[!htb]
	\centering
	\includegraphics[width=0.5\textwidth]{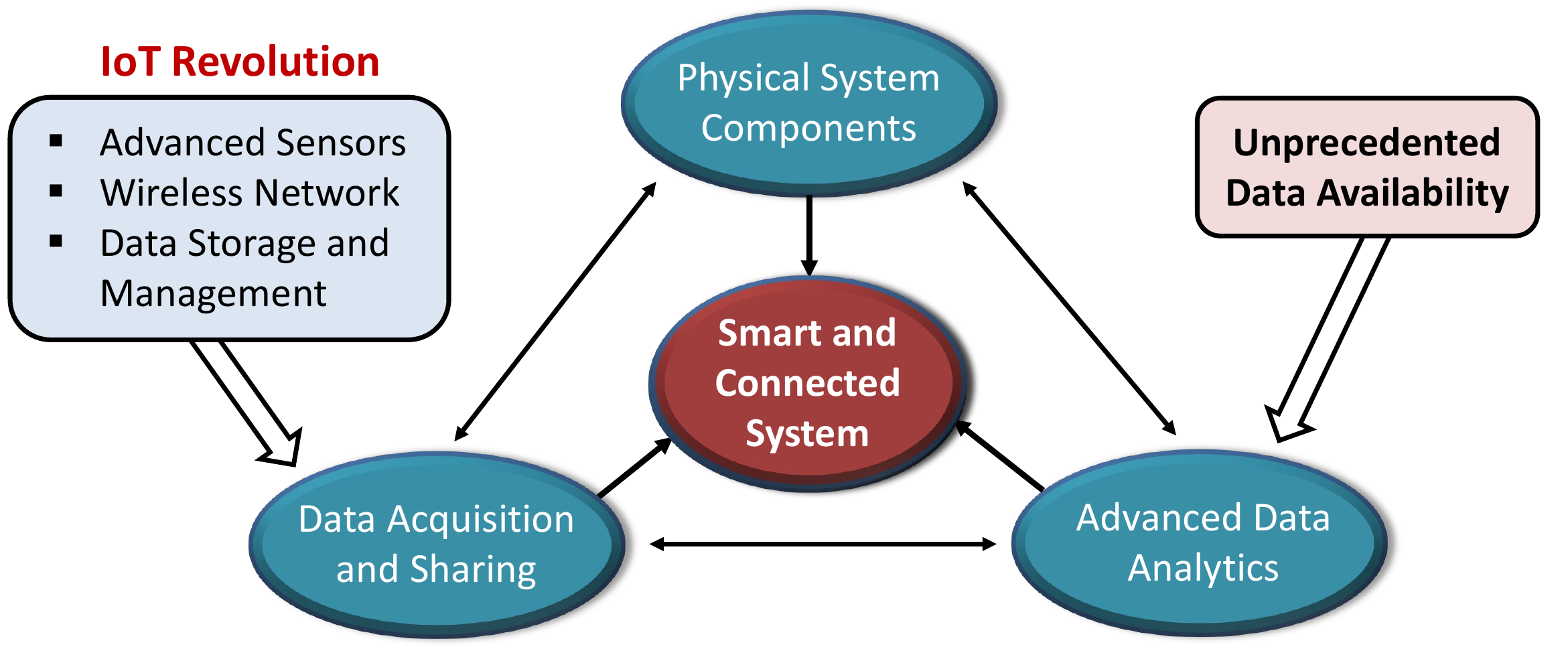}
	\caption{\label{fig:IoT_System} Key components of an IoT enabled system}
\end{figure}

IoT-enabled systems possess three defining characteristics: tangible physical components that comprise the system, connectivity among components that enable data acquisition and sharing, and data analytics and decision-making capabilities that transform a merely ``connected'' system into a ``smart and connected'' system. These defining features of IoT enabled systems \citep{porter2014smart,analytics2016age,chen2015data} are shown in Fig. \ref{fig:IoT_System}. IoT has brought broad disruptive societal impacts, particularly on economic competitiveness, quality of life, public health, and essential infrastructure \citep{madakam2015internet}. Companies around the globe have invested heavily in IoT, including: Google's Cloud IoT \citep{googleiot}, Samsung's Active wearable device \citep{sumsung}, Amazon's Webservices solutions \citep{awsamazon}, Rockwell's Connected Enterprise \citep{rockwell}, Welbilt's Smart Home Appliances, to name a few. The value at stake is more than 15 trillion dollars, a number expected to triple in the next decade \citep{atzori2010internet}. 

The essential feature of an IoT system is that data from multiple similar units and across multiple components within the system are collected during their operation, often in real-time. Since we have observations from potentially a large number of similar units, we can compare their operations, share information, and extract common knowledge to enable accurate prediction and control. One can argue that such a notion of IoT dates back a long time before the Industrial Revolution, to the time when artisans producing crafts in geographically close locations used to gather to share knowledge and perfect/standardize the quality of their crafted product \citep{srai2016distributed}. \textit{A lot has changed since then}.

\subsection{\emph{I\lowercase{O}T: The Present}}

 \begin{figure}[!htb]
	\centering
	\includegraphics[width=0.5\textwidth]{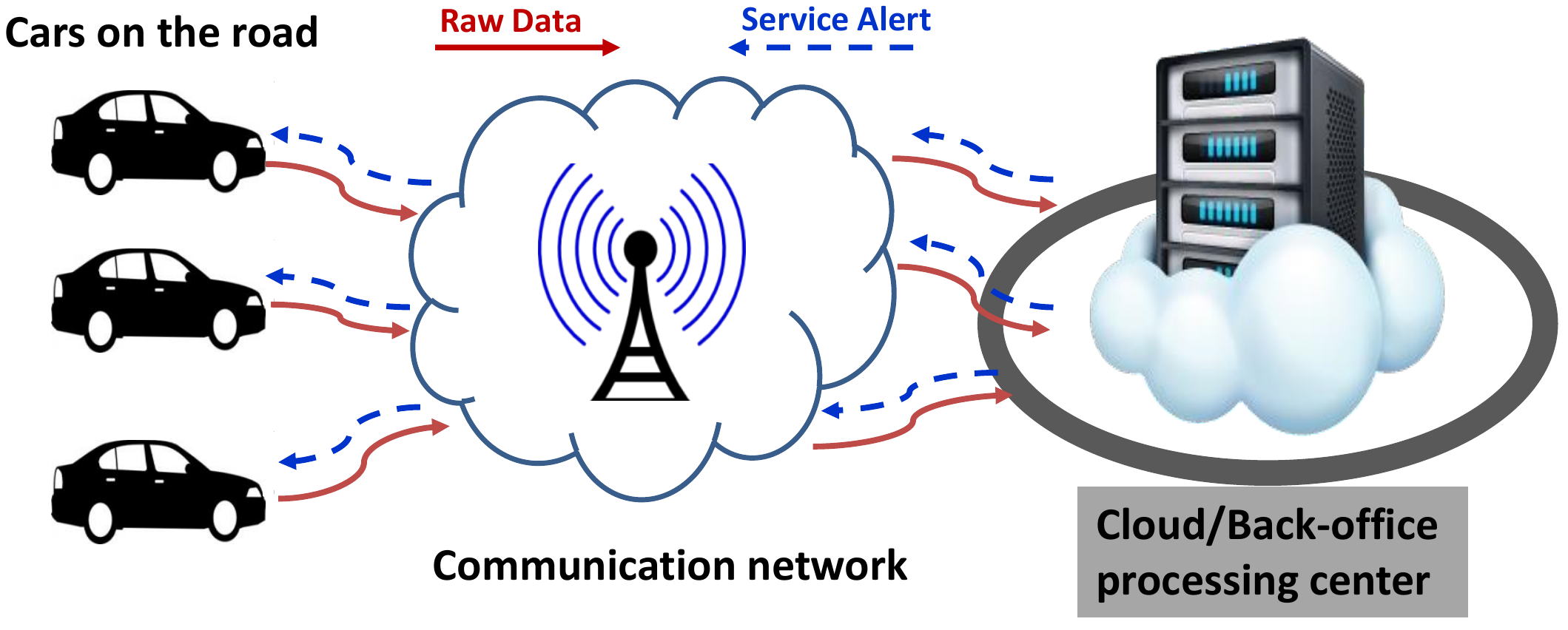}
	\caption{\label{fig:IoT_Present} Present Day IoT system }
\end{figure}

Starting with the industrial revolution came rapid advances in connectivity, automation, data science, cloud-based systems, among many others \citep{manufcaturingtrends,predictivemaintanence}. An IoT sensor price dropped to \$0.48 on average, and wide-area communication became readily available with around \$36.13 billion connected IoT devices in 2018 \cite{sensorprice}. Distributed computing allowed handling larger datasets than what was previously thought possible and cloud-based solutions for data storage and processing have become widely available for commercial use (ex: Amazon's AWS \cite{amazonaws} or Microsoft's Azure \cite{azurevideo}). This ushered in the present-day era of Industry 4.0 characterized by IoT-enabled systems \cite{atzori2010internet}. In this present era, a typical IoT-enabled system structure is shown in Fig. \ref{fig:IoT_Present}. Take for example GM's OnStar\textregistered \, or Ford's SYNC\textregistered \, teleservice systems \cite{onstar,GMIOT, sync}. Vehicles enrolled for this service have their data in the form of condition monitoring (CM) signals uploaded to the cloud regularly. The cloud then acts as a back-office or data center that processes the data to keep drivers informed about the health of their vehicle. In the cloud, GM and Ford train models that can monitor and predict maintenance needs, amongst others. The data is also used to cross-validate the behavior of their learned models for continuous improvement. When the need arrives, service alerts are then sent to drivers.

Much like other IoT giants such as Google, Amazon and Facebook, GM and Ford have long adopted this centralized approach towards IoT: (i) gigantic amounts of data are uploaded and stored in the cloud (ii) models (such as predictive maintenance, diagnostics, text prediction) are trained in these data centers (iii) the models are then deployed to the edge devices. Needless to say, the need to upload large amounts of data to the cloud raises privacy concerns, incurs high costs, and benefits large enterprises capable of building their own private cloud infrastructures at the expense of smaller entities. 

Here, distributed learning is often implemented in centralized systems to alleviate the huge computational burden via parallelization. In such systems, the clients are computing nodes within this centralized framework. Nodes can then access any part of the dataset, as data partitions can be continuously adjusted. In contrast, as described in the following sections, in IoFT, the data resides at the edge and is not centrally stored. As a result, data partitions are fixed and cannot be changed, shuffled, nor randomized.   

\subsection{\emph{I\lowercase{O}T: The Future}} \label{secfuture}
 
With the tremendous increase in computational power on edge devices, IoT is on its way to move from the cloud/datacenter to the edge device, hence the aforementioned notion of substituting the ``cloud'' by the ``crowd''. In this IoT system of the future (IoFT), devices collaboratively extract knowledge from each other and achieve the ``smart'' component of IoT, often with the orchestration of a central server, while keeping their personal data stored locally. This paradigm shift is based on one simple yet powerful idea: with the availability of computing resources at the edge, clients can execute small computations locally instead of learning models on the cloud and then only share the minimum information needed to learn that model. As a result, IoFT decouples the ability to do analytics from storing data in the cloud by bringing training to the edge device as well. The underlying premise is that IoFT devices have computational (ex: AI chips) and communication (ex: wifi) capabilities. 

\begin{figure}[!htb]
    \centering
    \includegraphics[width=0.4\textwidth]{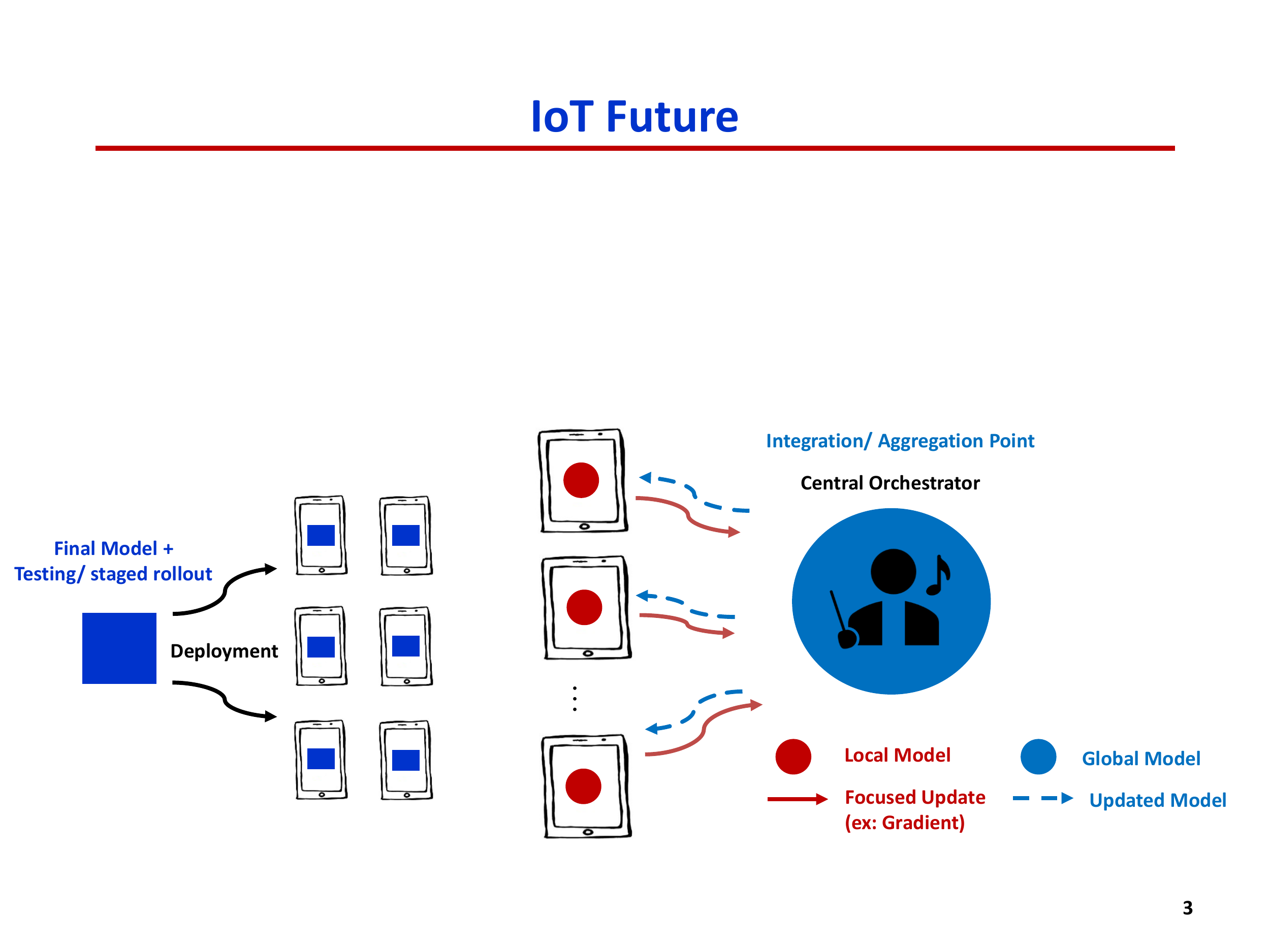}
    \caption{\label{fig:IoT_Future} IoFT: IoT system of the Future}
\end{figure}

Let us start with a simple example, assume the central orchestrator in Fig. \ref{fig:IoT_Future} wants to learn the mean ($\bar y$) of a single feature ($y$) over all clients. Now assume that clients have some computational capabilities. To calculate $\bar y$, client $i$ only needs to run a small calculation to compute their own mean ($\bar{y}_i$) and share it, rather than sharing their entire feature vector ($\bm{y}_i$). $\bar{y}_i$ is a sufficient statistic to learn $\bar y$.

In reality, models are often more complicated and require multiple communications between the central orchestrator and clients. For instance, and without loss of generality, assume that IoFT devices cooperate to learn a deep learning model through borrowing strength from each other, rather than using their own knowledge in isolation. In the decentralized realm of IoFT, model learning is often administered by a central orchestrator and follows the cycle shown in Fig. \ref{fig:IoT_Future}. (i) The orchestrator (i.e., the central server) selects a set of IoFT devices meeting certain eligibility requirements and broadcasts an initial model to the selected clients. This model contains the neural network (NN) architecture, initial weights, and a training program. (ii) IoFT devices perform local computations by executing the program on their local data, and each device reports its focused update to the orchestrator. Here the program can be running stochastic gradient descent (SGD) on local data, and the focused update can be updated weights or a gradient. It is worth noting that the client might choose to encrypt their focused update or add noise to it for enhanced privacy at this stage. (iii) The central orchestrator collects the focused updates from clients and aggregates them to update the global model. (iv) This procedure is then iterated over several rounds until a stopping criterion, such as validation accuracy, is met. Through this process, the global model can account for knowledge from all IoFT clients, and each client can indirectly make use of the knowledge from other clients. Finally, the learned global model goes through a testing phase such as quality-A/B testing on held-out devices and a staged rollout on a gradually increasing number of devices.

This decentralized paradigm shift, made possible by compute resources at the edge, sets forth many intrinsic advantages that include:

\begin{itemize}
	\setlength\itemsep{0em}
	\item \underline{{\it \textbf{Privacy}}}: By bringing training to the edge device, users no longer have to share their valuable information, instead, it is kept local and never shared. 
	\item \underline{{\it\textbf{Autonomy}}}: IoFT devices can be under independent control and opt-out of the collaborative training process at any time. Yet, with enhanced privacy in IoFT, clients will be more inclined to collaborate and build better models. 
	\item \underline{{\it\textbf{Computation}}}: As the number of IoT devices skyrockets, computational and storage needs accumulated from these devices (say smartphones) is far beyond what any data center or cloud computing system can handle \citep{FLfuture}. Instead, by exploiting compute and storage capacity at the edge, massive parallelization becomes a reality \citep{singh2019detailed,huo2018training}.  
	\item \underline{{\it\textbf{Cost}}}: Focused updates embody the principle of data minimization and contain the minimum information needed for a specific learning task. As a result, less information is transmitted to the orchestrator, which reduces communication costs and efficiently utilizes network bandwidth.  Also, compute power at the edge device is now utilized. Hence storage and computational needs of the orchestrator are minimal. This is in contrast to distributed systems where massive utilization and synchronization of GPU and CPU power in the cloud is needed.
	\item \underline{{\it\textbf{Fast Alerts and Decisions}}}: In IoFT, upon deployment of the final model to clients, real-time decisions or service alerts are achieved locally at the edge. In contrast, cloud-based systems incur a lag in deployment, as decisions made in the cloud need to be transmitted to the clients (as shown in Fig. \ref{fig:IoT_Present}).
	\item \underline{{\it\textbf{Minimal Infrastructure}}}: With the increase in computing power of IoT devices and the gradual market penetration of AI chips \citep{aichipsbillion}, minimal hardware is required to achieve the transition to IoFT.
	\item \underline{\textbf{Fast encryption: }} Encryption of focused updates can be done readily and with better guarantees compared to encrypting entire datasets.
	\item \underline{\textbf{Resilience}}: Edge devices are resilient to failures at the orchestrator level due to the existence of a local model.
	\item \underline{{\it\textbf{Diversity and Fairness}}}: IoFT allows integrating information across uniquely diverse datasets, some of which have been restricted to be shared previously (recall medical institutes example). This diversity and ability to learn across geographically disperse locations promotes fairness by combining data across boundaries \citep{brisimi2018federated,chang2018distributed}.
\end{itemize}

Having recently realized its disruptive potential to traditional IoT, industries are eagerly trying to exploit IoFT in their operating systems and production. However, these efforts are in their infancy phase, awaiting broad implementations. Google pioneered some of the IoFT applications in their mobile keyboard ``Gboard'' \citep{hard2018federated,chen2019federated,yang2018applied,ramaswamy2019federated} and Android messaging \citep{googlesupport} to improve next-word predictions and preserve privacy. Additionally, they introduced a decentralized framework to update android models on their Pixel phones \citep{mcmahan2017communication}. In this framework, each android phone updates its model parameters locally and sends out the updated parameters to the Android cloud, which trains its central model from the aggregated parameters. BigTech giants have since started to catch up and utilize FL in their systems. Most notably Apple adopted FL in their QuickType keyboard, ``Siri'' and privacy protection protocols \citep{bhowmick2018protection,appleprivacy}. As well as Microsoft in their device's telemetry data \citep{ding2017collecting}. Further, FL has seen some application in optimizing mobile edge computing and communication \citep{wang2019edge, lim2020federated}, computational offloading \citep{wang2019edge} and reliable network communication \citep{samarakoon2019distributed}. 

Most of the current IoFT applications are present within the technology industry and specifically tailored for mobile applications and few others. However, IoFT is expected to infiltrate all industries that benefit from knowledge sharing, data analytics, and decision-making. Indeed, the gradual use of FL in the technology industry has set in motion a timid yet insuppressible momentum for IoFT application in other sectors. For instance, in the healthcare field, FL is lately being used as a medium of collaboration between hospitals to share patients' electronic records and other medical data \citep{brisimi2018federated, nvidiaclara, huang2020loadaboost, futurefl}. In Sec. \ref{sec:applications}, we will present a deeper vision into how IoFT and FL will shape the future of various industries; those include manufacturing, transportation, energy, healthcare, quality \& reliability, business, and computing.

\subsubsection{Challenges}
IoFT as an emerging technology poses significant intellectual challenges. Interdisciplinary skills across diverse fields are needed to bring the great promise of IoFT into reality. Below we highlight some of the challenges and shed light on their uniqueness compared to centralized IoT systems. This is by no means an exhaustive list as IoFT challenges vary widely across different application sectors as highlighted in Sec. \ref{sec:applications}. 

\begin{itemize}
	\setlength\itemsep{0em}
	\item\underline{{\it\textbf{Statistical Heterogeneity}}}: IoFT devices often have local datasets that differ in both size and distribution. Recent papers have shown the unfortunate wide gap in the global model's performance across different devices due to their heterogeneity in distribution \citep{zhao2018federated, wang2019federated} and size \citep{duan2019astraea}. For instance, IoFT devices may have (i) unique outputs, labels, or features only observed within certain IoFT devices. (ii)  Similar outputs but with dissimilar features (i.e., feature distribution skew) or vice versa. This statistical heterogeneity directly consequences IoFT's ability to reach out to many devices operating under different external factors and subject to geographic, cultural, and socio-economic differences. In contrast, traditional IoT systems offer a key, yet often subtle fundamental advantage: the ability to handle nonindependent or identically distributed (\textit{i.i.d}) data by shuffling/randomizing the raw data collected in the cloud before learning; be it through distributed computing or learning on a single machine. This is not a luxury that IoFT possesses; rather, it is a price to pay for enhanced privacy.
	\item\underline{{\it\textbf{Personalization and Negative Transfer}}}: 
	In the IoFT process described in Sec. \ref{secfuture} all clients collaborate to learn a global model; ``one model that fits all''. This integrative analysis of multiple clients implicitly assumes that these local datasets share some commonalities. However, with heterogeneity, negative transfer of knowledge may occur, which leads to decreased performance relative to learning tasks separately \citep{kontar2020minimizing, li2020negative}. One possible solution is through personalized modeling where global models are adapted for local clients (refer to Sec. \ref{sec:personlized} for data-driven personalization approaches). Indeed, personalization may be the fundamental tool to overcome the heterogeneity barrier intrinsic to IoFT. Yet developing validation techniques to identify negative transfer and minimize it is a critical problem in FL. 
	\item\underline{{\it\textbf{Communication Efficiency and Resource Management}}}: Communication can be a critical bottleneck for IoFT, especially with a large number of participants. Unlike cloud datacenters, edge devices in IoFT often have limited communication bandwidth with unstable and slow connection \citep{konevcny2016federated}. As a result, IoFT devices are often unreliable and can drop out due to battery loss or connectivity loss. Besides that, devices themselves are heterogeneous in their computational capabilities and memory budgets. Therefore, resource management in IoFT is of critical importance.  Methods such as compressed communication  \citep{tang2019doublesqueeze, koloskova2019decentralized}, client selection \citep{xu2020client} and optimal trade-offs between convergence rates, accuracy, energy consumption, latency and communications \citep{nguyen2020resource,reisizadeh2019robust} are of high future relevance. Another possible approach is through incentive design to encourage reliable clients to participate in the training process and minimize dropout rates \citep{kang2019incentive}. 
	\item\underline{{\it\textbf{Privacy}}}: Privacy remains one of the key challenges and motivators behind IoFT. IoFT systems are prone to poisoning attacks on both edge devices and the central server. Targeted data perturbations \citep{bagdasaryan2020backdoor,chen2017targeted,liu2017trojaning} to specific labels/instances or corrupting a large number of devices (i.e., fake devices) can immensely reduce accuracy.
	Further, a malicious server might be able to reconstruct raw data even through a focused update. As a result, secure computation, aggregation, and communication are needed in IoFT \citep{beimel2019power, bittau2017prochlo}. So is adversarial data modeling to ensure robustness against corrupted data in case breaches are inevitable  \citep{madry2017towards}. 
	\item\underline{{\it\textbf{Bias and Fairness}}}: IoFT systems can raise bias and fairness concerns. For example, sampling reliable phones with a larger bandwidth (i.e., more expensive phones) can lead to models mostly representative of people with certain socioeconomic statuses. Further, it is often important to build models that are competitive over different groups or attributes. This becomes a bigger challenge if such sensitive attributes are not shared. Therefore, fair FL is an important challenge to tackle within IoFT \citep{yue2021gifair,li2019fair}
	\item\underline{{\it\textbf{Other Statistical and Optimization Challenges}}}: We also refer readers to Sec. \ref{sec:statopt} for both statistical and optimization challenges/opportunities and Sec. \ref{sec:applications} for domain-specific challenges in different sectors. 
\end{itemize}

We here note that Secs. \ref{sec:global}, \ref{sec:personlized}, \ref{sec:metalearning} shed light on data-driven modeling approaches (global, personalized and meta-learning) aimed to tackle of the challenges above. However, we exclude (i) privacy and communication efficiency: since there are excellent surveys focused mainly on these challenges (refer to Sec. \ref{sec:overview}) (ii)  resource management: since literature in that area is still scarce. 

\subsubsection{IoFT structures}
The underlying structure and overall architecture of IoFT should be tailored to fit certain applications and overcome specific challenges. Current IoFT architectures are influenced by the data composition and the FL learning process. For instance, in the situation where multiple clients collaborate to learn a global model with the orchestration of a central server (as seen in Fig. \ref{fig:IoT_Future}), it is implicitly assumed that local datasets share a common feature space but have a different sample space - i.e. different clients. Such data composition is technically referred to as Horizontally partitioned data \citep{yang2019federated}. A typical FL system architecture for Horizontally portioned data (also known as Horizontal FL (HFL)), would exploit the availability of a common feature space. 
Notably, horizontally partitioned data are very common across different applications, making HFL the common practice in IoFT \citep{hard2018federated,chen2019federated,yang2019federated,yang2018applied,ramaswamy2019federated}. 

\begin{wrapfigure}{R}{0.3\textwidth}
	\centering
	\includegraphics[width=0.3\textwidth]{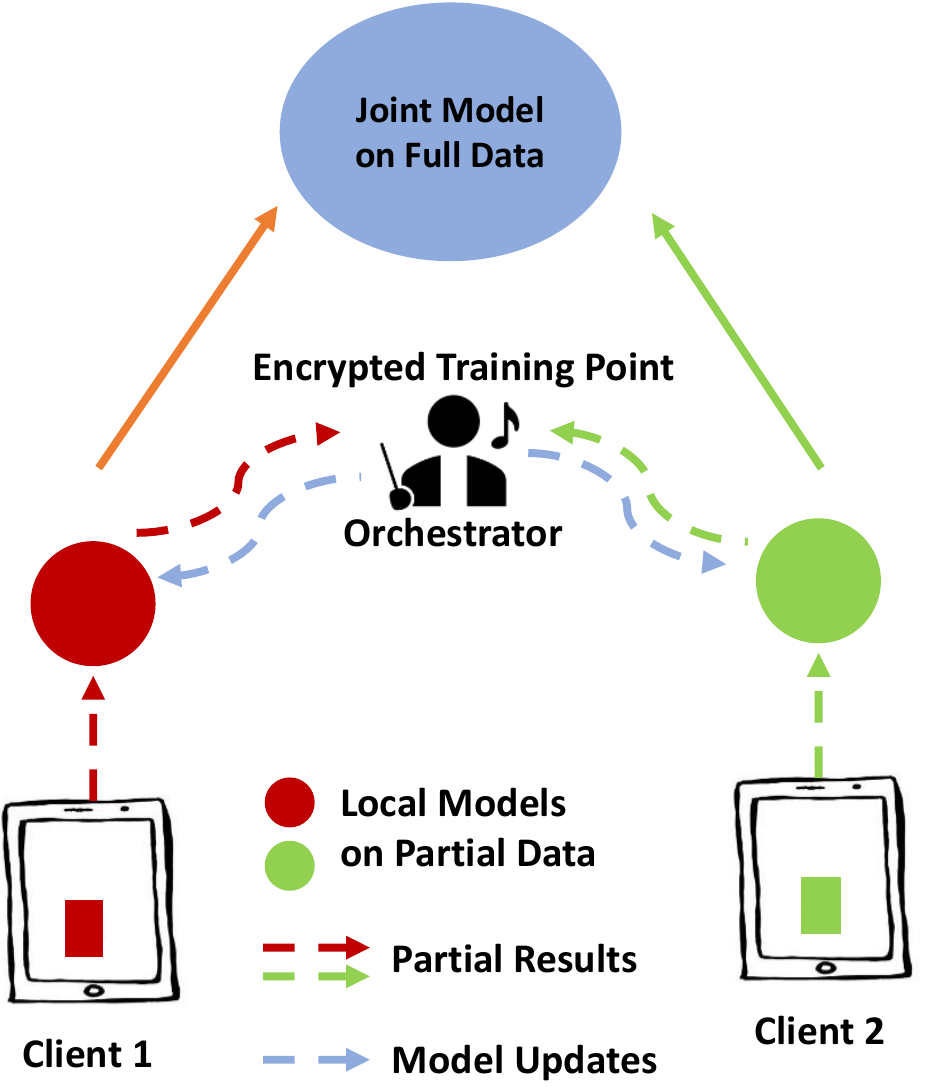}
	\caption{\label{fig:vfl} IoFT with vertically partitioned data}
\end{wrapfigure}

However, not all datasets share a common feature space which naturally poses the need for a different architecture. Vertically partitioned data, which refers to datasets sharing a different feature space but similar sample space, is another familiar theme in various applications. Such datasets mostly appear in scenarios that involve joint collaboration between large enterprises. Consider as an example two different health institutes, each owning different health records yet sharing the same patients. Suppose you wish to build a predictive model for a patient's health using a complete portfolio of medical records from both healthcare institutes. Unlike HFL where each client trains a local model using their own data, training a local model requires data owned by other clients since each client holds a disjoint subset of the data. Accordingly, a typical FL system architecture for Vertically partitioned data (also known as Vertical FL (VFL) \citep{yang2019federated}) is designed to introduce secure communication channels between clients to share the needed training data, while preserving privacy and preventing data leakage from one provider to another. For this, VFL architecture may involve a trusted, neutral party to orchestrate the federation. The orchestrator aligns and aggregates data from participants to allow for collaborative model building using the joint data, see Fig. \ref{fig:vfl}.
Nonetheless, VFL remains less explored than HFL, and most of the currently developed structures can only handle two participants \citep{nock2018entity,hardy2017private,yang2019parallel}. More challenging scenarios can occur when clients have datasets that share only partial overlap in the feature and sample spaces. FL in these cases can leverage transfer learning techniques to allow for collaborative model training \citep{pan2010survey,yang2019federated}. 

\begin{wrapfigure}{R}{0.23\textwidth}
	\centering
	\includegraphics[width=0.23\textwidth]{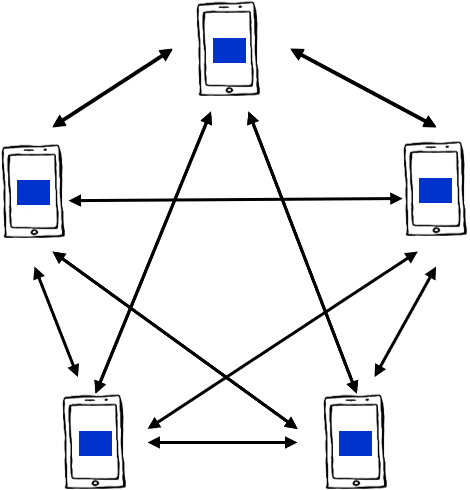}
	\caption{\label{fig:peer} Peer-to-peer network}
\end{wrapfigure}

The structures described above are designed to handle challenges arising from dataset partitioning. However, different challenges require new structures. One notable commonality of the above structures is the usage of a central orchestrator that coordinates the FL process in IoFT. The caveat, however, is that a central orchestrator is a single point of failure and can lead to a communication bottleneck with a large number of clients \citep{lian2017can}. Accordingly, fully decentralized solutions can be explored to nullify the dependency on a central orchestrator. In fully decentralized architectures, communication with the central server is replaced by peer-to-peer communication, as seen in Fig. \ref{fig:peer}. In this setting, no central location receives model updates/data or maintains a global model over all clients, however, clients are set to communicate with each other to reach desired solutions. Notably, such peer-to-peer networks are better able to achieve scalability in situations with a large number of clients, thanks to their fully decentralized mechanism \citep{kermarrec2015want}; the current success of blockchains is a clear demonstration of this.
Further, they offer additional security guarantees as it is difficult to observe the system's full state \citep{bellet2018personalized}. However, such architecture yields performance concerns. Some clients could be malicious in peer-to-peer networks and potentially corrupt the network (e.g., violate data privacy). Others could be unreliable and thus disrupt the communication channels. Consequently, a level of trust in a central authority in a peer-to-peer architecture can be of benefit in regulating the network's protocols.   

The structures discussed here are by no means comprehensive, and several others exist in the literature (see \cite{yang2019federated,kairouz2019advances,rahman2020survey,li2020federated}). However, the common denominator here is that IoFT structures spawn from challenges of FL applicability to different scenarios. As IoFT is poised to infiltrate more and more fields, domain-specific challenges will dictate its architecture. 

\section{Learning a Global Model}  \label{sec:global}

Hereon, we discuss data-driven approaches for FL within IoFT. As aforementioned, we classify model building in FL into three categories: (i) a global model, (ii) a personalized model (iii) a meta-learning model. We then provide an in-depth overview of data-driven models, open challenges, and possible alternatives within these three categories. 

As will become clear shortly, the current FL techniques mostly focus on predictive modeling using deep learning and first-order optimization techniques, specifically stochastic gradient descent (SGD). This is understandable as the immense data collected within IoFT often necessitates such an approach. Yet, as we discuss in the statistical/optimization perspective (Sec. \ref{sec:statopt}) and applications (Sec. \ref{sec:applications}) sections, exploring FL beyond predictive models and deep learning is critical for its wide-scale implementation. Topics such as graphical models, correlated inference, zeroth and second order distributed optimization, validation \& hypothesis testing, uncertainty quantification, design of experiments, Bayesian optimization, optimization under conflicting objectives (see in Sec. \ref{sec:business}), game theory and reinforcement learning, amongst others, are yet to be explored in the IoFT realm.

\subsection{\emph{A General Framework for FL}}

As highlighted in Fig. \ref{fig:IoT_Future}, IoFT allows multiple clients to collaborate and learn a shared model while keeping their personal data stored locally. This shared model is referred to as the global model as it aims to maximize utility across all devices. One can view the global model as: ``one model that fits all'', where the goal is to yield better performance in expectation across all clients relative to each client learning a separate model using its own data.

We start by constructing the objective function of a global model. Assume there are $N$ clients (or local IoFT devices) and each client $i$ has $n_i$ number of observations. The general objective of training a global model is to minimize the average over the objective of all clients: 
\begin{equation}\label{eq:obj}
\min_{\bm{w}} F(\bm{w})\coloneqq \frac{1}{N}\sum_{i=1}^NF_i(\bm{w}) \, ,
\end{equation}
where $F_i(\bm{w})$ is usually a risk function on client $i$. This risk function can be expressed as
\[
F_i(\bm{w})=\mathbb{E}_{(x_i,y_i)\sim \mathcal{D}_i}\left[\ell(f_{\bm{w}}(x_i),y_i)\right], 
\]
where $\mathcal{D}_i$ indicates the data distribution of the $i$-th client's data observations $(x_i, y_i)$, $f_{\bm{w}}$ is the model to be learned  parametrized by weights $\bm{w}$, and $\ell(\cdot, \cdot)$ is a loss function. 

The risk function is usually approximated by the empirical risk given as $F_i(\bm{w})=\mathbb{E}_{(x_i,y_i)\sim \mathcal{D}_i}\left[\ell(f_{\bm{w}}(x_i),y_i)\right] \approx \frac{1}{n_i}\sum_{j=1}^{n_i} \left[\ell(f_{\bm{w}}(x_j),y_j)\right]$. Therefore, learning a global model in FL aims at minimizing the average of risks over all clients.  However, unlike centralized training, in IoFT client $i$ can only evaluate its own risk function $F_i(\bm{w})$ and the central server does not have access to the data from the clients. Client and central server training are thus decoupled. 

Given this setting, Algorithm \ref{alg::generalfl} is a general ``computation then aggregation'' \citep{zhang2020fedpd} framework for FL. In each communication round, a central orchestrator selects a subset of clients ($\mathcal{S}\subseteq [N]$) and broadcasts the global model information to the subset. Each client then updates the global model using its own local data. Afterwards, clients send their updated models back to the central orchestrator/server. The orchestrator aggregates and revises the global model based on input from clients. The process repeats for several communication rounds until a stopping criterion, such as validation accuracy, is met.  Note that we use $[N]$ to denote the set $\{1,2,...N\}$, $D_i$ a client's dataset and the superscript $^t$ to represent the $t$-th communication round between the central server and selected clients, where $t\in \{1,..,T\}$.

\begin{algorithm}[!htb]
   \caption{Framework for Learning a Global Model}
   \label{alg::generalfl}
\begin{algorithmic}[1]
   \STATE {\bfseries Input:} Client datasets $\{D_i\}_{i=1}^N$, $T$, initialization for $\bm{w}$

   \FOR{$t$ = 1, 2, $\cdots T$}
   \STATE Orchestrator selects a subset of clients $\mathcal{S}\subseteq [N]$, broadcasts global model $\bm{w}^t$, or a part of it, to clients in $\mathcal{S}$.
   
   \FOR {each $i\in\mathcal{S}$}
   
   \STATE Clients update model parameters  $\bm{w}^{t+1}_i=\textit{client\_update}\left(\bm{w}^t,D_i\right)$
   \STATE Clients send updated parameters $\bm{w}^{t+1}_i$ to server. 
   \ENDFOR
   \STATE Orchestrator updates $\bm{w}^t$ by aggregating client updates $\bm{w}^{t+1}=\textit{server\_update}\left(\left\{\bm{w}^t_i\right\}\right)$
   \ENDFOR
\end{algorithmic}
\end{algorithm}

One of the simplest FL algorithms is \texttt{FedSGD} \citep{parallelsgd,fedsgd}, a distributed version of SGD. \texttt{FedSGD} was initially used for distributed computing in a centralized regime. \texttt{FedSGD} partitions the data across multiple computing nodes. In every communication round, each node calculates the gradient from its local data using a single SGD step. The calculated weights are then averaged across all nodes. As a data-parallelization approach, \texttt{FedSGD} utilizes the computation power of several compute nodes instead of one. This approach accelerates vanilla SGD and has been widely used due to the growing size of datasets collected nowadays. Furthermore, since \texttt{FedSGD} only performs one step of SGD on a local node, averaging updated weights is equivalent to averaging gradients ($\eta$ denotes steps size):
\[
\mathbb{E}_i\left[\bm{w}^{t}-\eta \nabla F_i(\bm{w}^t)\right]=\bm{w}^{t}-\eta \mathbb{E}_i\left[ \nabla F_i(\bm{w}^t)\right].
\]

Despite being a viable option, traditional distributed optimization algorithms are often unsuitable in IoFT due to the large communication cost and the presence of heterogeneity. \texttt{FedSGD} transmits the gradient vector from one machine to the other after each single local optimization iterate. This issue is not critical in centralized distributed training when computation nodes are usually connected by large bandwidth infrastructure. However in IoFT, data lives on the edge device and not on a computing node. Communication with the central orchestrator at each gradient calculation is not feasible and may suffer immensely when the edge devices have limited communication bandwidth with unstable or slow connection.   

\begin{figure}[htbp]
\centering
\includegraphics[width=0.5\textwidth]{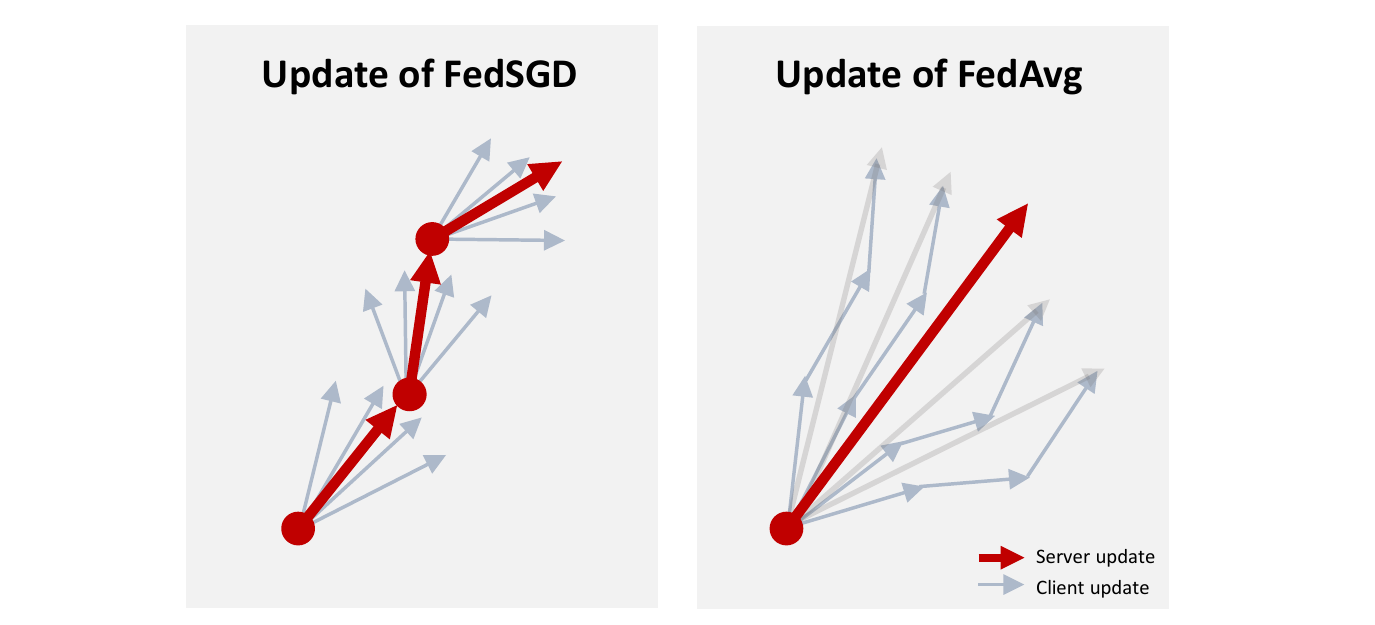}
\caption{An illustration of \texttt{FedAvg} and \texttt{FedSGD}. Grey arrows represent gradients evaluated on the local client. Bold red arrows represent a global model update on the central server in one communication round. In \texttt{FedSGD}, each client performs one step of SGD, and sends the update to the server, while \texttt{FedAvg} allows each client to perform multiple SGD steps before averaging.}
\label{fig::framework}
\end{figure}

To resolve this challenge, the seminal work of \citet{fedavg} proposed a simple solution: \texttt{FedAvg}. The fundamental idea is that clients run multiple updates of model parameters before passing the updated weights to the central orchestrator. Specifically, in \texttt{FedAvg}, clients update local models by running multiple steps (e.g., $E$ local steps) of SGD on their local objective $\min_{\bm{w}_i^t} F_i(\bm{w}_i^t)$. Upon receiving updated weights from clients, the \textit{server\_update} function simply calculates the average of the client models: $\bm{w}^{t+1}=\frac{1}{|\mathcal{S}|}\sum_{i\in \mathcal{S}}\bm{w}^t_i$. An illustration contrasting \texttt{FedAvg} and \texttt{FedSGD} is shown in Fig \ref{fig::framework}. Here one can also add flexibility by re-scaling the global update with a step size $\eta_g$,  $\bm{w}^{t+1}=\bm{w}^{t}+\eta_g\left(\frac{1}{|\mathcal{S}|}\sum_{i\in \mathcal{S}} [\bm{w}^t_i-\bm{w}^{t}]\right)$.

Indeed, despite its simplicity, \texttt{FedAvg} has seen wide empirical success within FL due to its communication efficiency and strong predictive performance on several datasets. To this day, \texttt{FedAvg} remains a standard benchmark that is often hard to beat. However, a major observed challenge was that the performance of \texttt{FedAvg} and \texttt{FedSGD} degrades significantly \citep{fedavg} when data across clients are heterogeneous, i.e. non-\textit{i.i.d.} data. Here one should note that empirical results have shown that \texttt{FedAvg} requires fewer communication rounds 
than \texttt{FedSGD} even in the presence of heterogeneity \citep{fedavg}.

\subsection{\emph{Tackling Heterogeneity}} \label{sec:heterogeniety}

As previously discussed, an intrinsic property of IoFT is that the data distribution across clients is often imbalanced and heterogeneous. Unlike centralized systems, data cannot be randomized or shuffled prior to inference as it resides on the edge. For example, wearable devices collect data on users' health conditions such as heartbeats and blood pressure. Due to the many differences across users, the amount of data collected can significantly vary, and statistical patterns of these data are not alike, often with unique or conflicting trends. This heterogeneity degrades the performance of \texttt{FedAvg}. The reason is that minimizing the local empirical risk $F_i(\bm{w})$ is sometimes fundamentally inconsistent with minimizing the global empirical risk $F(\bm{w})$ when data are non-\textit{i.i.d}.  Mathematically, it also implies that $F(\bm{w}^*)\neq\frac{1}{N}\sum_{i=1}^NF_i(\bm{w}_i^*)$, where superscript $^*$ indicates an optimal parameter. This phenomenon is known as client-drift \citep{karimireddy2020scaffold}. Notice that if local datasets are \textit{i.i.d.}, when the size of local datasets approaches infinity, $F_i(\bm{w})$ converges to the global empirical risk $F(\bm{w})$, hence optimal solutions coincide.  
In the following, we introduce some works trying to address the heterogeneity challenge. 

One method to allay heterogeneity in FL is regularization. In the literature, regularization has been a popular method to reduce model complexity. As less complex models usually generalize better \citep{friedman2001elements,rademacher}, regularization attains better testing accuracy. In FL, regularization places penalties on a set of parameters in the objective function to encourage the model to converge to desired critical points. Researchers in FL have proposed several notable algorithms using regularization techniques to train global models with non-\textit{i.i.d.} data. Perhaps the most basic one is \texttt{FedProx} \citep{fedprox} which adds a quadratic regularizer term (a proximal term) to the client objective:
\[
\min_{\bm{w}^t_i} F_i(\bm{w}^t_i)+\frac{\mu}{2}\left\|\bm{w}^t_i-\bm{w}^t\right\|^2.
\]

The proximal term $\frac{\mu}{2}\left\|\bm{w}^t_i-\bm{w}^t\right\|^2$ in \texttt{FedProx} limits the impact of client-drift by penalizing local updates that move too far from the global model in each communication round. Parameter $\mu$ controls the degree of penalization.  It was also seen that \texttt{FedProx} allows each device to have a different number of local iterations $E_i$, which is especially useful when IoFT devices vary in reliability and communication/computation power. Experimental results show that \texttt{FedProx} can partially alleviate heterogeneity, while reducing communication cost due to the often faster convergence and ability of reliable clients to run more updates than others. Here it is important to note that despite reducing client-drift, \texttt{FedProx} is still based on in-exact minimization since it does not align local and global stationary solutions.


Besides \texttt{FedProx}, \cite{dane,fedsvrg,zhang2020fedpd,acarfederated} also develop a framework to tackle heterogeneity through regularization. Among this literature, \texttt{DANE} \citep{dane} was proposed for distributed optimization yet is readily amenable to FL settings. \texttt{DANE} uses a local objective:
\begin{align}
\label{eqn:daneobjective}
\min_{\bm{w}_i^{t}} F_i(\bm{w}^t_i)-&\left\langle\nabla F_i(\bm{w}^{t-1})-\nabla F(\bm{w}^{t-1}),\bm{w}^t_i-\bm{w}^{t-1}\right\rangle \notag \\&+\frac{\mu}{2}\left\|\bm{w}^t_i-\bm{w}^{t-1}\right\|^2 \, ,
\end{align}
where $\mu$ is also a parameter for weighting the regularization and $\bm{w}^{t-1}$ is the global update at the previous communication round. Compared with the \texttt{Fedprox} objective, \eqref{eqn:daneobjective} adds one term that linearly depends on $\bm{w}_i^t$. This term aligns the gradient of the local risk to that of the global risk. To see it, one can calculate the gradient of \eqref{eqn:daneobjective} as $\nabla F_i(\bm{w}^t_i)-\left(\nabla F_i(\bm{w}^{t-1})-\nabla F(\bm{w}^{t-1})\right)+\mu\left(\bm{w}^t_i-\bm{w}^{t-1}\right)$, where the term $\nabla F_i(\bm{w}^{t-1})-\nabla F(\bm{w}^{t-1})$ approximates the difference between the local and global gradient by its value at the last communication round. It is shown that objective \eqref{eqn:daneobjective} can be interpreted as mirror descent. Interestingly, if the local loss function is quadratic, optimizing \eqref{eqn:daneobjective} can approximate performing Newton updates.

The exact minimization in \eqref{eqn:daneobjective} is sometimes infeasible, as edge devices usually have limited computation resources. 
To resolve the issue, Stochastic Controlled Averaging algorithm (\texttt{SCAFFOLD}) \cite{karimireddy2020scaffold} replaces the exact minimization by several gradient descent steps on the local objective below,
\begin{equation}
\label{eqn::scaffoldobjective}
 \min_{\bm{w}_i^{t}} F_i(\bm{w}_i^{t})-\left\langle\bm{c}^t_i-\bm{c}^t,\bm{w}_i^{t} \right\rangle  \, .
\end{equation}
where control variables $\bm{c}_i^{t}$ and $\bm{c}$ are defined as $\bm{c}_i^t=\nabla F_i(\bm{w}_i^{t-1})$, i.e. the local gradient at the end of the last communication round, and $\bm{c}^t=\frac{1}{N}\sum_i\bm{c}^t_i$. Objective \eqref{eqn::scaffoldobjective} is akin to \eqref{eqn:daneobjective}, since $\left\langle\bm{c}^t_i-\bm{c}^t,\bm{w}_i^{t}\right\rangle$ also has the alignment effect, except that it does not have the proxy term $\frac{\mu}{2}\left\|\bm{w}^t_i-\bm{w}^{t-1}\right\|^2$. To show the update rule in communication round $t$, we use $\bm{w}_i^{t,e}$ to denote the weight at the $e$-th local iterate, and set $\bm{w}_i^{t,0}=\bm{w}^t$. In round $t$, the server samples a group of clients $\mathcal{S}$. For client $i$ in $\mathcal{S}$, the local update of \texttt{SCAFFOLD} is:
\begin{align}
\label{eq:update_sca}
    \bm{w}^{t,e+1}_i=\bm{w}^{t,e}_i-\eta(\nabla F_i(\bm{w}_i^{t,e})-\bm{c}^t_i+\bm{c}^t),
\end{align}
for $e\in \{0,\cdots,E-1\}$. After $E$ iterations, clients send weights $\bm{w}_i^t=\bm{w}_i^{t,E}$ and gradients $\bm{c}_i^{t+1}=\nabla F_i(\bm{w}_i^{t})$ to the server. The server takes the average of control variables $\bm{c}^{t+1}=\bm{c}^t+\frac{1}{N}\left(\sum_{i\in\mathcal{S}}\left[\bm{c}_i^{t+1}-\bm{c}^t\right]\right)$, and re-scales the updates for weights by $\eta_g$, $ \bm{w}^{t+1}=\bm{w}^{t}+\frac{\eta_g}{|\mathcal{S}|}\left(\sum_{i\in\mathcal{S}}[\bm{w}_i^{t}-\bm{w}^{t}]\right)$. Note here that $c^t$ is taken over all $N$ clients. For those that did not participate \texttt{SCAFFOLD} re-uses the previously computed gradients. 

The idea behind \texttt{SCAFFOLD} is very intuitive. To solve \eqref{eq:obj}, the ideal (centralized) update is that each client uses all client's data $\bm{w}^{t+1}_i=\bm{w}^t-\frac{\eta_g}{N}\sum_{i=1}^N\nabla F_i(\bm{w}^{t})$. However such update rule is not possible in IoFT due to the need to communicate the gradients $\nabla F_i(\bm{w})$ with the orchestrator  at every optimization iterate. To mimic the ideal update, \texttt{SCAFFOLD} uses $\bm{c}_i^t $ to approximates $\nabla F_i(\bm{w}_i^t)$ at using the last communication round, for all $i$. Then also $\bm{c}^t$ may approximate the gradient of the global risk, $\bm{c}^t=\frac{1}{N}\sum_{i=1}^N\bm{c}^t_i\approx \frac{1}{N}\sum_{i=1}^N\nabla F_i(\bm{w}_i^{t})$. If this approximation holds, the update of \texttt{SCAFFOLD} becomes similar to ideal (centralized) update. One caveat in such update scheme is that $\bm{c}^t$ may not always equal (or approximate) the ideal value  $\frac{1}{N}\sum_{i=1}^N\bm{c}^t_i$. Adding to that, $\bm{c}^t$  re-uses the previously computed gradients when clients do not participate. Therefore, when client participation rate is low, $\bm{c}^t$ can deviate far away from the ideal update leading to degraded optimization performance.

Empirically, \texttt{SCAFFOLD} requires fewer communication rounds to converge compared with \texttt{FedAvg}. A very similar algorithm is \texttt{Federated SVRG} \citep{fedsvrg}, which applies stochastic variance reduced gradient descent to approximately solve \eqref{eqn::scaffoldobjective}. The update rule is $\bm{w}^{t,e+1}_i=\bm{w}^{t,e}_i-\eta\left(S_i\left(\nabla F_i(\bm{w}^{t,e}_i)-\bm{c}^t_i\right)+\bm{c}^t\right)$, where $S_i$ is a diagonal matrix to rescale gradients. \texttt{Federated SVRG} reduces to \texttt{SCAFFOLD} when one sets $S_i$ to the identity matrix. 

As discussed, despite its efficiency on several FL tasks, \texttt{SCAFFOLD} does not work well in low client participation cases. To this end, \texttt{FedDyn} \citep{acarfederated}
uses a specially designed dynamic regularization to align gradients under partial participation. The objective on client $i$ is defined as:
\begin{equation}
\label{eqn::feddynobjective}
 \min_{\bm{w}^t_i} F_i(\bm{w}^t_i)-\left\langle\nabla F_i(\bm{w}^{t-1}_i),\bm{w}^t_i\right\rangle+\frac{\mu}{2}\left\|\bm{w}^t_i-\bm{w}^{t-1}\right\|^2.   
\end{equation}

Objective \eqref{eqn::feddynobjective} is also closely related to \eqref{eqn:daneobjective}. In \eqref{eqn:daneobjective}, when the weight $\bm{w}$ is near critical points of the global risk $F(\bm{w})$, $\nabla F(\bm{w})$ is close to $0$, thus \eqref{eqn:daneobjective} reduces to \eqref{eqn::feddynobjective}. As a simple fixed points analysis, when all models start from $\bm{w}_i^{t-1}=\bm{w}^{t-1}=\bm{w}^*$, i.e. a critical point of the global loss, the optimal solution of \eqref{eqn::feddynobjective} is still $\bm{w}^*$, thus local updates will stay at $\bm{w}^*$. \texttt{FedDyn} is proved to converge to critical points of the global objective with a constant stepsize. 
Also, to deal with partial client participation, \texttt{FedDyn} uses a SAG-style \cite{sag} averaging rule in \textit{server\_update}: instead of only averaging gradients from clients that participated in the training in one communication round, \texttt{FedDyn} estimates gradients on disconnected clients based on historic values and averages all gradients (or gradient estimates).  
In practice, \texttt{FedDyn} is shown to achieve similar test accuracy with much fewer communication rounds compared with \texttt{FedAvg} and \texttt{FedProx}, especially when client participation rate is low. 
 
A closely related algorithm to \texttt{FedDyn} is Federated primal-dual (\texttt{FedPD}) \citep{zhang2020fedpd}. \texttt{FedPD} and \texttt{FedDyn} have different formulations, but end up with the same update rule under some conditions. In \texttt{FedPD}, the optimization problem in \eqref{eq:obj} is reformulated to a constrained optimization problem
\begin{equation}
\label{eqn::constrainedoptimizationformulation}
 \min_{\bm{w}_0,\{\bm{w}_i\}}\frac{1}{N}\sum_{i=1}^NF_i(\bm{w}_i)\quad\text{subject to}\quad\bm{w}_1=...=\bm{w}_N=\bm{w}_0 \, .  
\end{equation}

To solve the constrained optimization problem, \texttt{FedPD} introduces dual variables $\lambda_1,\cdots,\lambda_N$, then defines the augmented Lagrangian (AL) for client $i$ to be $\mathcal{F}_i(\bm{w}_i, \lambda_i,\bm{w}_0)=F_i(\bm{w}_i)+\left\langle\lambda_i,\bm{w}_i-\bm{w}_0\right\rangle+\frac{\mu}{2}\left\|\bm{w}_i-\bm{w}_0\right\|^2$. 
\texttt{FedPD} uses alternative descent on primal and dual variables to optimize $\mathcal{F}_i$. More specifically, \texttt{FedPD} first randomly initializes $\bm{w}_{0}, \lambda_i, \bm{w}_i$ for all clients. At round $t$, the algorithm updates $\bm{w}^{t+1}_i$ by optimizing $\mathcal{F}_i$ and fixing $\lambda_i=\lambda_i^t$ and $\bm{w}_0 = \bm{w}_{0}^t$. It then updates dual variables by $\lambda_i^{t+1}=\lambda_i^{t}+\mu(\bm{w}_i^{t+1}-\bm{w}^{t}_{0,i})$ and also  $\bm{w}^{t+1}_{0,i}=\bm{w}_i^{t+1}+\frac{1}{\mu}\lambda_i^{t+1}$. After the local updates, \texttt{FedPD} makes a random choice with probability $1-p$, that all clients send updated $\bm{w}^{t}_{0,i}$ back to the orchestrator which updates $\bm{w}_0^{t+1}=\frac{1}{N}\sum_{i=1}^N\bm{w}^{t+1}_{0,i}$ and broadcasts updated $\bm{w}_0^{t+1}$. With probability $p$, all clients set $\bm{w}^0=\bm{w}^{t+1}_{0,i}$ and continue local training. 
Interestingly, by letting $\lambda_i=\nabla F_i(\bm{w}_i^{t-1})$, it was shown that \texttt{FedPD} is equivalent to \texttt{FedDyn} with full client participation on an algorithmic level \cite{equivalencefeddynandfedpd}. However, different from \texttt{FedDyn}, \texttt{FedPD} does not directly apply to partial participation settings.

Another algorithm that uses a constrained optimization formulation is \texttt{FedSplit} \citep{fedsplit}. \texttt{FedSplit} \citep{fedsplit} applies Peaceman-Rachford splitting \citep{prsplit,monotoneoperator}. More specifically, \texttt{FedSplit} concatenates $\bm{w}_1,\cdots,\bm{w}_N$ into one long vector $\bm{\mathcal{W}}=(\bm{w}_1^{\texttt{T}},...,\bm{w}_N^{\texttt{T}})^{\texttt{T}}$  and finds the optimal solution of $\frac{1}{N}\sum_{i=1}^NF_i(\bm{w}_i)$ on the subspace $\mathcal{E}=\{\bm{\mathcal{W}}|\bm{w}_1=\cdots=\bm{w}_N\}$. The problem is also known as consensus optimization \cite{consensusoptimization}. An important concept in consensus optimization is the normal cone defined as $\mathcal{N}_\mathcal{E}(\bm{\mathcal{W}})=\mathcal{E}^\perp$ for $\bm{\mathcal{W}}\in \mathcal{E}$ and empty otherwise. At the optimal solution, the gradient should be in the normal cone of $\mathcal{E}$: 
$$
0\in \nabla_{\bm{\mathcal{W}}} \sum_{i=1}^NF_i(\bm{w}_i)+\mathcal{N}_\mathcal{E}(\bm{\mathcal{W}}) \,.
$$

\texttt{FedSplit} treats gradient $\nabla_{\bm{\mathcal{W}}}$ and normal cone $\mathcal{N}_\mathcal{E}$ as two operators, and uses Peaceman-Rachford splitting \cite{monotoneoperator} to find a solution $\bm{\mathcal{W}}$ that satisfies the optimality condition. After some derivations, the authors propose the following update rules. At communication round $t$, clients update their local weights $\bm{w}_i^t$, send them to server and store a local copy. In the following round, client $i$ receives global update $\bm{w}^t$, and calculates:
\[
\left\{\begin{aligned}
&\bm{w}_i^{t+\frac{1}{2}}=\arg\min_{\bm{w}_i} F_{i}(\bm{w}_i)+\frac{\mu}{2}\left\|\bm{w}_i-\left(2\bm{w}^t-\bm{w}_i^t\right)\right\|^2\\
&\bm{w}_i^{t+1} = \bm{w}_i^{t} + 2\left(\bm{w}^t-\bm{w}_i^{t+\frac{1}{2}}\right) \, .
\end{aligned}\right. 
\]
The \textit{server\_update} simply averages $\bm{w}^{t+1}=\frac{1}{N}\sum_{i=1}^N\bm{w}_i^{t+1}$. Intuitively, operator splitting adds a regularization term centered at $2\bm{w}^t-\bm{w}_i^t$ to the local objective. The carefully designed update rule has two advantages. Firstly it help alleviate client-drift: \texttt{FedSplit} convergences linearly to critical points of the global loss on convex problems. Also, it accelerates convergence: the theoretical convergence rate is faster than that of \texttt{FedAvg} on strongly convex problems. 

In addition to algorithms applicable to general federated optimization problems, there are models designed specifically for neural networks to handle heterogeneity. For instance, researchers pointed out that re-permutation of neurons may cause declined performance in the aggregation step of FL. This re-permutation problem is due to the fact that different neural networks created by a weight permutation might represent the same function.

For example, consider a simple NN, $f_{\bm{w}}(x)=W_2\sigma\left(W_1x\right)$ where $\sigma$ is an activation function, $f_{\bm{w}}(x) \in \mathbb{R}^{d_y}$, $x \in \mathbb{R}^{d_x}$ and $W_1\in\mathbb{R}^{w}\times\mathbb{R}^{d_x}$ and $W_2\in\mathbb{R}^{d_y}\times\mathbb{R}^{w}$ are weight matrices. One can multiply a permutation matrix $\Pi \in \mathbb{R}^{w} \times \mathbb{R}^{w}$ to $W_1$ and $W_2$, $f_{\bm{w}}(x)=W_2\Pi^\texttt{T}\sigma\left(\Pi W_1 x\right)$, and the function remains the same. However updates on different clients may be attracted to networks with a different permutation matrix $\Pi$. This can cause averaging over weights to fail. To cope with this, \cite{yurochkin2019bayesian} propose a neuron matching algorithm called Probabilistic Federated Neural Matching (\texttt{PFNM}). \texttt{PFNM} assumes $\Pi_iW_1$ and $W_2\Pi_i^{\texttt{T}}$ are generated by a hierarchical probabilistic model whose hyper-parameters are determined by global weights. Then \texttt{PFNM} uses Bayesian inference to estimate the hyper-parameters, and reconstructs the global model from the inference.


However, \cite{wang2020federated} argue that \texttt{PFNM} can only work on simple fully connected neural networks. To solve the problem, they extend \texttt{PFNM} to Federated Matched Averaging (\texttt{FedMA}) algorithm. \texttt{FedMA} updates weights of a neural network layer by layer. Firstly, clients train local NNs and send the trained first layer weights $W^{(1)}_{i}$ to the orchestartor, where $W^{(1)}_{i}$ denotes the weight vector of layer $1$ from client $i$. The server uses matching algorithms such as the Hungarian algorithm in \texttt{PFNM} to estimate $\Pi^{(1)}_i$ that represents the permutation vector of first layer model weights for client $i$. Thus, $\Pi^{(1)}_iW^{(1)}_{i}$ become the matched weights after re-permutation. The server then averages the results $\overline{W}^{(1)}=\frac{1}{N} \sum_{i=1}^N\Pi^{(1)}_iW^{(1)}_{i}$, and broadcasts the averaged $\overline{W}^{(1)}$. After receiving $\overline{W}^{(1)}$, clients continue to train the remaining layers with the first layer fixed to $\overline{W}^{(1)}$. A similar match-then-average process repeats for remaining layers. For \texttt{FedMA}, the number of communication rounds equals the number of network layers. \texttt{FedMA} is reported to have strong performance on CIFAR-10 and a well known  language dataset called Shakespeare. Additionally, the performance of FedMA improves with the increase of local epochs $E$, while that of \texttt{FedAvg} and \texttt{FedProx} drops after a threshold of $E$ due to the discrepancy between local models (i.e. local weights wander away from each other). Thus \texttt{FedMA} enables clients to train more epochs between consecutive communications.

All approaches described above are of a frequentist nature. However, there has also been a recent push on improving global modeling through a Bayesian framework. The intuition is simple; rather than betting our results on one hypothesis ($\bm{w}$) obtained via optimizing the empirical risk, one may average over a set of possible $\bm{w}$ or integrate over all $\bm{w}$ weighted by their posterior probability $\mathbb{P}(\bm{w}|D=\{D_1,\cdots, D_N\})$. This is the underlying philosophy of marginalization compared to optimization, whereby in the frequentist approach predictions are obtained through substituting the posterior by $\mathbb{P}(\bm{w}|D)=\delta (\bm{w}=\hat{\bm{w}})$, where $\hat{\bm{w}}$ is the single optimized weight and $\delta$ is an indicator function. Indeed, this notion of Bayesian ensembling has seen a lot of empirical success in Bayesian deep learning \citep{maddox2019simple, izmailov2018averaging}. 

\begin{figure}[htbp]
\centering
\includegraphics[scale=0.35]{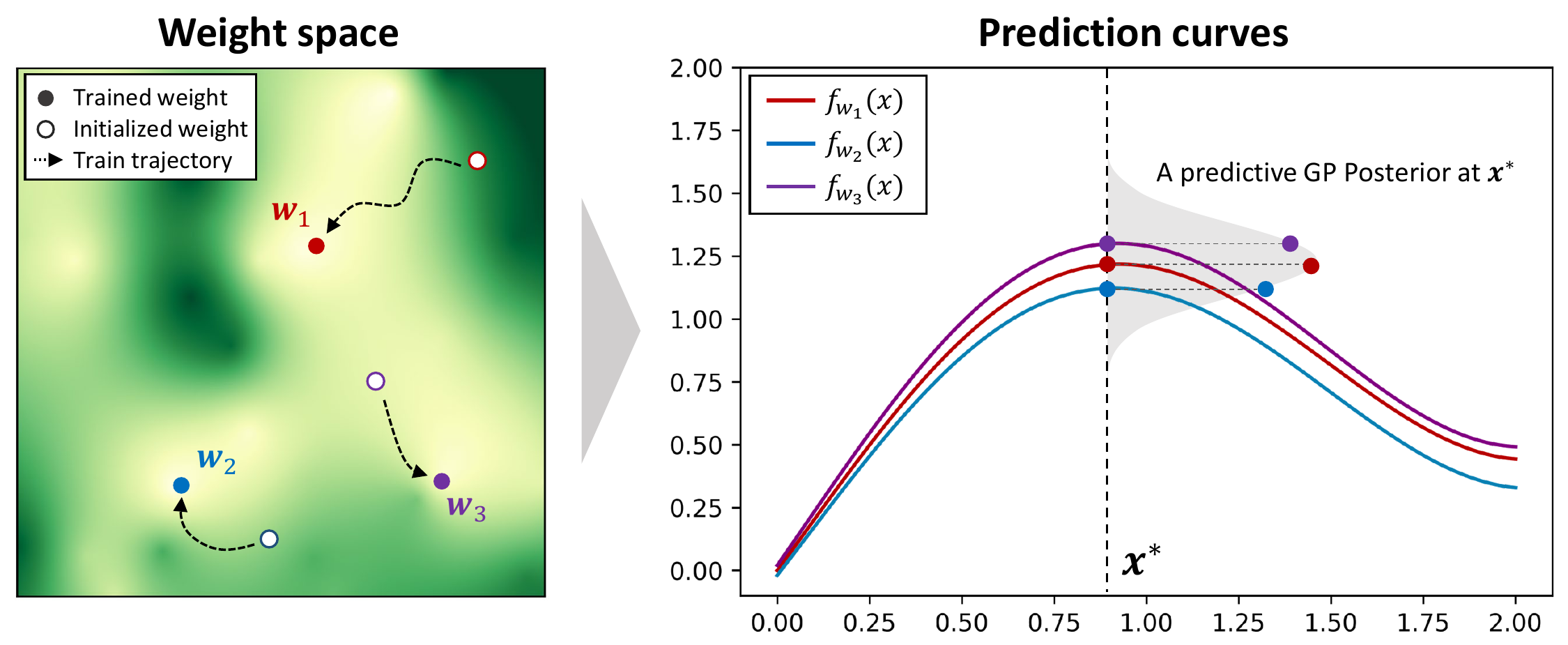}
\caption{An illustration of an ensemble of $K=3$ models. The three model weights on the left figure correspond to the three predictions on the right. Although the weights are well separated, the predictions admit the same limiting posterior distribution.}
\label{fig::ensemble}  
\end{figure}

One such approach is \texttt{Fed-ensemble} \citep{Fedensemble}.  \texttt{Fed-\allowbreak ensemble}, is a simple plug-in into any FL algorithm that aims to learn an ensemble of $K$-models without additional communication costs. To do so, \texttt{Fed-ensemble} follows a random permutation sampling scheme where at each communication round, every client trains one of the $K$ models and then aggregation happens for each model separately (using \texttt{FedAvg} or other FL approaches). This approach corresponds to a variational inference scheme \citep{blei2017variational, zhang2018advances} for estimating a Gaussian mixture variational distribution whose centers are randomly initialized at the beginning. Predictions on a new input $x^*$ are then obtained by taking an average over the predictions of the $K$ models
\begin{equation} \label{eq:ensemble}
    f(x^*)=\frac{1}{K}\sum_kf_{\bm{w}_k}(x^{\star}) \, .
\end{equation}
\texttt{Fed-ensemble} is also able to quantify predictive uncertainty. Using a neural tangent kernel argument, the authors show that all predictions from all $K$ models converge to samples from the same limiting Gaussian process in sufficiently overparameterized regimes (see Fig. \ref{fig::ensemble}) where each mode can behave like a model trained by centralized training. 

Another recent work taking insights from Bayesian inference is \texttt{FedBE} \citep{chen2020fedbe}. It performs statistical inference on the client-trained models and uses knowledge-distillation (KD) to update the global model. Intuitively, the goal of KD is to use high-quality base models from a global distribution $\mathbb{P}(\bm{w})$ to direct the global model update. More specifically, after receiving $\{\bm{w}_i\}_{i \in \mathcal{S}}$ from clients, the server fits them with a Gaussian or Dirichlet distribution and then samples from the estimated distribution to form an ensemble of $K$ models $\{\bm{w}_1,\cdots,\bm{w}_K\}$. Similar to \eqref{eq:ensemble}, the ensemble prediction on a new point $x^*$ is given by $y_{ensemble}=\frac{1}{K}\sum_{i=1}^{K}f_{\bm{w}_k}(x^*)$. In \textit{server\_update}, the global model $f_{\bm{w}}$ is trained to mimic the average prediction of models in the ensemble by minimizing the discrepancy between the two predictions evaluated on an additional unlabeled dataset $D^{\text{add}}$ on the server:
\[
\bm{w}^{t+1}=\arg\min_{\bm{w}} \mathbb{E}_{x\sim D^{\text{add}}}\left[\text{Div}(y_{ensemble}(x),f_{\bm{w}}(x))\right] \, ,
\]
where \text{Div} denotes a divergence measure, here cross-entropy. The updated $\bm{w}^{t+1}$ is then sent to all clients. The authors empirically show that the ensemble and knowledge distillation turns out to be more robust to non-\textit{i.i.d}. data than \texttt{FedAvg}. This approach however requires storing additional data on server which is not always feasible. 

\subsection{\emph{Efficient \& Effective optimization}} \label{sec:eff}

Several studies attempt to improve \texttt{FedAvg} by adopting adaptive optimization algorithms to the FL realm. They show theoretically or empirically that the improved algorithms can converge faster and accelerate global model training. In general, acceleration can be achieved by either improving the server aggregation step (\textit{server\_update}) or client updates (\textit{client\_update}). \texttt{FedAdam} and \texttt{FedYogi} \citep{fedadam} bring the well known Adam \citep{kingma2014adam} and Yogi \citep{reddi2018adaptive} algorithms to FL through augmenting the \textit{server\_update} function by adaptive stepsizes. More specifically, \texttt{FedAdam} and \texttt{FedYogi} use a second order moment estimate $v_t$ to adaptively adjust the learning rate. $v_{0}$ is initialized at the beginning. Upon receiving $\bm{w}^t_i$ from clients, server calculates $\Delta \bm{w}^t_i=\bm{w}^t_i-\bm{w}$, and averages them $\Delta_t=\sum_{i\in \mathcal{S}}\frac{1}{|\mathcal{S}|}\Delta\bm{w}^t_i$. \texttt{FedAdam} updates $v_t$ as:
\[
v_t=\zeta v_{t-1}+\left(1-\zeta\right)\Delta_t^2 \, ,
\]
and \texttt{FedYogi} as:
\[
v_t=v_{t-1}-\left(1-\zeta\right)\Delta_t^2\text{sign}(v_t-\Delta_t^2) \, ,
\]
where $\zeta$ is a parameter for exponential weighting. The update rule for both \texttt{FedAdam} and \texttt{FedYogi} is:
\[
\bm{w}^{t+1}=\bm{w}^t+\eta\frac{\Delta_t}{\sqrt{v_t}+\epsilon} \, ,
\]
where $\epsilon$ is a small constant for numerical stability. Though the proved theoretical convergence rates of \texttt{FedAdam} and \texttt{FedYogi} are only comparable to those of \texttt{FedAvg}, the adaptive methods show strong performance on several FL tasks.
Considering the success of adaptive stepsize methods in numerous important fields including language models \citep{attention}, GANs \citep{gantraining2019,gantraining2020}, amongst others, we believe their use in FL is promising. 
A related algorithm in this vein is federated averaging with server momentum (\texttt{FedAvgM}) \citep{liu2020accelerating}, which uses server momentum in the \textit{server\_update} step.

Besides modifying \textit{server\_update}, multitudes of algorithms redesign the \textit{client\_update} function. 
For instance, there are some attempts to expedite local training by combining accelerating techniques in optimization.



\texttt{FedAc} \citep{fedac} is a federated version of an accelerated SGD. Instead of updating a single variable $\bm{w}_i$ as \texttt{FedAvg} does, \texttt{FedAc} updates three sequences $\{\bm{w}_i,\left(\bm{w}_i\right)^{ag},\left(\bm{w}_i\right)^{md}\}$ iteratively on the client side by the following rules for several steps:
\[
\left\{\begin{aligned}
\left(\bm{w}_i^t\right)^{md}&\leftarrow\zeta_1\bm{w}_i^t+\left(1-\zeta_1\right)\left(\bm{w}_i^t\right)^{ag}\\
\bm{g}_i&\leftarrow\nabla F_i(\left(\bm{w}_i^t\right)^{md})\\
\left(\bm{w}_i^t\right)^{ag}&\leftarrow\left(\bm{w}_i^t\right)^{md}-\eta_1 \bm{g}_i\\
\bm{w}_i^t&\leftarrow\left(1-\zeta_2\right)\bm{w}_i^t+\zeta_2\left(\bm{w}_i^t\right)^{md}-\eta_2 \bm{g}_i \, .\\
\end{aligned}\right.
\]
$\zeta_1, \zeta_2, \eta_1, \eta_2$ are four hyper-parameters. Among them $\zeta_1, \zeta_2$ are exponential averaging parameters, and $\eta_1, \eta_2$ are stepsize parameters. The server averages $\bm{w}_i^t$ and $\left(\bm{w}_i^t\right)^{ag}$ from sampled clients, and broadcasts the averaged $\bm{w}^{t+1}$ and $\left(\bm{w}^{t+1}\right)^{ag}$, which clients will take as initialization of $\bm{w}_i^{t+1}$ and $\left(\bm{w}_i^{t+1}\right)^{ag}$ in the next communication rounds. The algorithm then proceeds till convergence. \cite{fedac} theoretically prove that \texttt{FedAc} can achieve a linear convergence rate faster than \texttt{FedAvg} when global risk $F(\bm{w})$ in \eqref{eq:obj} is strongly convex. 
Empirical results show that \texttt{FedAc} saves communication cost when there are many devices in the network. 



\texttt{LoAdaBoost} \citep{loadaboost} adaptively determines the training epochs of clients by monitoring the training loss on each client and adjusting the training schedule accordingly. More specifically, after one communication round, clients send training losses, in addition to updated weights to the server. The server estimates the median of the training loss $F_i$, $F_{median}\coloneqq median(\{F_i\}_{i=1,\cdots,N})$. In the next round, all clients train for a certain amount of epochs $\frac{E'}{2}$, where $E'$ is the average budget of epochs. If the training loss is lower than $F_{median}$, the local training is deemed to have reached its goal in this round, and the updated weight will be directly sent back to server. If the training loss is higher than $F_{median}$ on client $i$, then the model underfits client $i$. As a result, \texttt{LoAdaBoost} will train the model on client $i$ for extra epochs until the local training loss is lower than $L_{median}$ or the total epochs exceed $\frac{3}{2}E$, whichever comes faster. Such dynamic training schedules allow \texttt{LoAdaBoost} to take resources of clients into consideration, thus can better utilize computation power on edge devices and enable faster training.

\subsection{\emph{Sampling Clients}} 

Due to the often sheer size and unreliability of edge devices participating within IoFT, not all clients can participate in each communication round of the training process as shown in Algorithm \ref{alg::generalfl}. Therefore, choosing the appropriate subset $\mathcal{S}$ at each communication round between the orchestrator and client is of utmost importance in FL. Here we shed light on some existing schemes, other possible alternatives and their implications.

We first start by noting that an alternative approach to write the global objective in  \eqref{eq:obj} is through giving different weights to client risk function. This is given as 
\begin{equation}\label{eq:objweighted}
\min_{\bm{w}} F(\bm{w})\coloneqq \sum_{i=1}^Np_iF_i(\bm{w})=\mathbb{E}_i[F_i(\bm{w})] \, ,
\end{equation}
where $p_i$ is a weight such that $p_i\geq0$ and $\sum_{i=1}^N p_i=1$. In IoFT, it is common to have datasets of different sizes. Thus, a natural choice is to set $p_i=\frac{n_i}{n}$ where $n$ is the total data size across all clients $n=\sum_{i=1}^N n_i$. Clearly if all clients have the same dataset size $n_i$, objective \eqref{eq:objweighted} reduces to \eqref{eq:obj}.


 
Indeed, although most algorithms for FL use \eqref{eq:obj}, both \texttt{FedAvg} and \texttt{FedProx} (among the earliest methods) use \eqref{eq:objweighted} by adding weights $p_i$. \texttt{FedAvg} samples clients $\mathcal{S}\subseteq [N]$ uniformly with probability $\mathbb{P}_i=\frac{1}{N}$, and averages client models with weights proportional to their local dataset size $n_i$: $\bm{w}^{t+1} = \frac{1}{\sum_{i\in \mathcal{S}}n_i}\sum_{i\in \mathcal{S}}n_i\bm{w}_i^{t+1}$. On the other hand, \texttt{FedProx} samples clients with probability $\mathbb{P}_i=p_i=\frac{n_i}{n}$ and averages client models with equal weights: $\bm{w}^{t+1} = \frac{1}{|\mathcal{S}|}\sum_{i\in \mathcal{S}}\bm{w}_i^{t+1}$. These sampling probability and weights are chosen to make client updates unbiased estimates of global updates - i.e. unbiased estimates of  $\sum_{i=1}^Np_i\nabla F_i(\bm{w})$.

However, both sampling schemes may have some drawbacks. For example, uniform sampling may be inefficient since the orchestrator can often sample unreliable clients or clients with very small datasets. Dataset size-based sampling addresses this issue, but it may raise fairness concerns as some clients are rarely sampled and trained. This also makes the training procedure more prone to adversarial clients with large datasets that can directly impact the training process.



To form better sampling schemes and accelerate training, adaptive sampling techniques have also been proposed. These FL algorithms update the sampling probability $\mathbb{P}_i^t$ after each communication round from historical statistics \citep{clientsamplewithloss,clientsamplewiththeta,oort,clientsamplingwithgn}.  Such methods usually sample clients on which the model fits worse, more often. Intuitively, when a model incurs high training loss or large gradient norms on client $i$, client $i$ is not performing well under the current model and should be trained for more epochs.

There are a range of choices for measuring the performance of a model on the client. Among them, there is a set of literature that calculates the sampling probabilities adaptively using gradient norms of the clients \citep{convexadaptivesampling,nonconvexadaptivesampling,adambs}. Generally this can be given as

$$\mathbb{P}_i^t \propto \mbox{exp} \left(\gamma {\left\|\nabla F_i (\bm{w}^t_i)\right\|^2} \right) \, ,$$

where $\gamma$ is some constant. Other approaches sample clients based on their training loss  \citep{banditsurvey, clientsamplewithloss} where the gradient is substituted by the local loss $F_i (\bm{w}^t_i)$ at the end of each training round. In this set of literature,  exploration and exploitation schemes are also used to continuously update the sampling  probability. Client selection usually improves model performance and speeds up the training process. For example, \cite{oort} can achieve $1.2\times$ to $14.1\times$ speed up in terms of time-to-accuracy compared with vanilla \texttt{FedProx} or \texttt{FedYogi}.

Due to prevailing statistical and system heterogeneity among clients, we believe client sampling techniques will be of great significance when practitioners try to deploy FL frameworks in IoFT. An effective sampling scheme can efficiently exploit differences in client's resources while at the same time improving training speed and accuracy. Further, studying the connections between adaptive sampling and client re-weighting schemes (see Sec. \ref{sec:fairness}) used for fairness is an interesting topic worthy of investigation.


\subsection{\emph{Fairness across clients}} \label{sec:fairness}

In IoFT, it is crucial to ensure that all edge devices have good prediction performance. However, the key challenge is that devices with insufficient amounts of data, limited bandwidth, or unreliable internet connection are not favored by conventional FL algorithms. Such devices can potentially end with bad predictive ability. Besides this notion of individual fairness, group fairness also deserves attention in FL. As FL penetrates many practical applications, it is important to achieve fair performance across groups of clients characterized by their gender, ethnicity, etc. Before diving into the literature, we first start by formally defining the notion of fairness. Suppose there are $d$ groups (e.g., ethnicity) and each client can be assigned to one of those groups. Group fairness can be defined as follows.

\begin{definition}
\label{definition:1}
Denote by $\{a^i_{\bm{w}}\}_{1\leq i\leq d}$ the set of performance measures (e.g., testing accuracy) of a trained model $\bm{w}$. For trained models $\bm{\theta}$ and $\tilde{\bm{\theta}}$, we say $\bm{\theta}$ is more fair than $\tilde{\bm{\theta}}$ if $Var(\{a^i_{\bm{\theta}}\}_{1\leq i\leq d})<Var(\{a^i_{\tilde{\bm{\theta}}}\}_{1\leq i\leq d})$, where $Var$ denotes variance.
\end{definition}

When $d=N$, this definition is equivalent to the individual fairness. Definition \ref{definition:1} is widely adopted in most FL literature \citep{mohri2019agnostic, li2019fair, huang2020fairness,zhang2020fairfl, zeng2021improving}. This notion of fairness might be different from traditional definitions such as demographic disparity \citep{feldman2015certifying}, equal opportunity and equalized odds \citep{hardt2016equality} in centralized systems. The reason is that those conventional definitions cannot be extended to FL as there is no clear notion of an outcome which is optimal for an edge device \citep{kairouz2019advances}. Instead, fairness in FL can be defined as equal access to effective models (e.g., the accuracy disparity \citep{zafar2017fairness} or the representation disparity \citep{li2019fair}). Specifically, the goal is to train a global model that incurs a uniformly good performance across all devices or groups \citep{kairouz2019advances}. 

Despite the importance of fairness, unfortunately, very limited work exist along this line in FL. As will become clear shortly, the few works in this area mainly focus on a client re-weighting scheme through exploiting the weighted global objective in (\ref{eq:objweighted}) instead of (\ref{eq:obj}). 

\texttt{GIFAIR-FL} \citep{yue2021gifair} is the first algorithm that can handle both group and individual fairness in FL. Specifically, it achieves fairness by penalizing the spread in the loss among client groups. This can be translated to the following optimization problem:
\begin{align*}
    H(\bm{w})=\sum_{i=1}^Np_iF_i(\bm{w}) + \lambda\sum_{1\leq j<k\leq d}\left|L_j(\bm{w})-L_k(\bm{w})\right|,
\end{align*}
where $\lambda$ is a regularization parameter and
\begin{align*}
    L_j(\bm{w})=\frac{1}{|\mathcal{A}_j|}\sum_{i\in\mathcal{A}_j}F_i(\bm{w})
\end{align*}
is the average loss for group $j$ and $\mathcal{A}_j$ is the set of indices of devices who belong to group $i$. The original formulation of $H(\bm{w})$ can be further simplified as
\begin{align*}
    &H(\bm{w})=\sum_{i=1}^Np_i\left(1+\frac{\lambda}{p_i|\mathcal{A}_{s_i}|}r_i(\bm{w})\right) \\ &F_i(\bm{w})\coloneqq\sum_{i=1}^Np_iH_i(\bm{w})
\end{align*}
where
\begin{align*}
    r_i(\bm{w})=\sum_{1\leq j\neq s_i\leq d}\text{sign}(L_{s_i}(\bm{w})-L_j(\bm{w})),
\end{align*}
and $s_i\in[d]$ is the group index of device $i$. $r_i(\bm{w})$ can be viewed as a scalar that related to the statistical ordering of $L_{s_i}$ among client group losses. Therefore, to collaboratively minimize $H(\bm{w})$, each edge device $i$ is minimizing $H_i(\bm{w})$, a scaled version of the original local loss function $F_i(\bm{w})$. The central server will aggregate local parameters and update $\{r_i(\bm{w})\}_{i=1}^N$ at every communication round. From the expression of $r_i(\bm{w})$, one can see that a higher value of $r_i(\bm{w})$ will be assigned to the client that has higher group loss. Therefore, \texttt{GIFAIR-FL} will impose higher weights for clients with bad performances. Furthermore, those weights will be dynamically updated at every communication round to avoid possible model over-fitting. \citep{yue2021gifair} has shown that \texttt{GIFAIR-FL} will converge to an optimal solution or a stationary point even when heterogeneity exists.

Agnostic federated learning (\texttt{AFL}) \citep{mohri2019agnostic} is another algorithm that re-weights clients at each communication round. Specifically, it solves a robust optimization problem in the form of
\begin{align*}
    \min_{\bm{w}}\max_{\max{p_1,\ldots,p_N}}\sum_{i=1}^Np_iF_i(\bm{w}).
\end{align*}
\texttt{AFL} computes the worst-case combination of weights among edge devices. This approach is robust but may be conservative since it only focuses on the largest loss and thus causes pessimistic performance to other clients. \citet{du2021fairness} further refine the notation of \texttt{AFL} by linearly parametrizing weight parameters by some kernel functions. Upon that \citet{hu2020fedmgda+} combine minimax optimization with gradient normalization to formulate a new fair algorithm \texttt{FedMGDA+}. 

Inspired by fair resource allocation for wireless networks problems, \cite{li2019fair} propose the \texttt{q-FFL} algorithm for fairness. They slightly modify the loss function and add a power $q$ to each user
\begin{equation}
\label{q-FFL}
\min_{\bm{w}} \sum_{i=1}^N\frac{p_i}{q+1}F^{q+1}_i(\bm{w}).
\end{equation}
The intuition is that $q$ tunes the amount of fairness: the algorithm will incur a larger loss to the users with poor performance. Therefore, \texttt{q-FFL} can ensure uniform accuracy across all users. 

To the best of our knowledge, fairness is an under-investigated yet critical area in the FL setting. We hope this section can inspire the continued exploration of fair FL algorithms.




\section{Learning a Personalized Model}
\label{sec:personlized}

 \begin{figure}[!htb]
	\centering
	\includegraphics[width=0.4\textwidth]{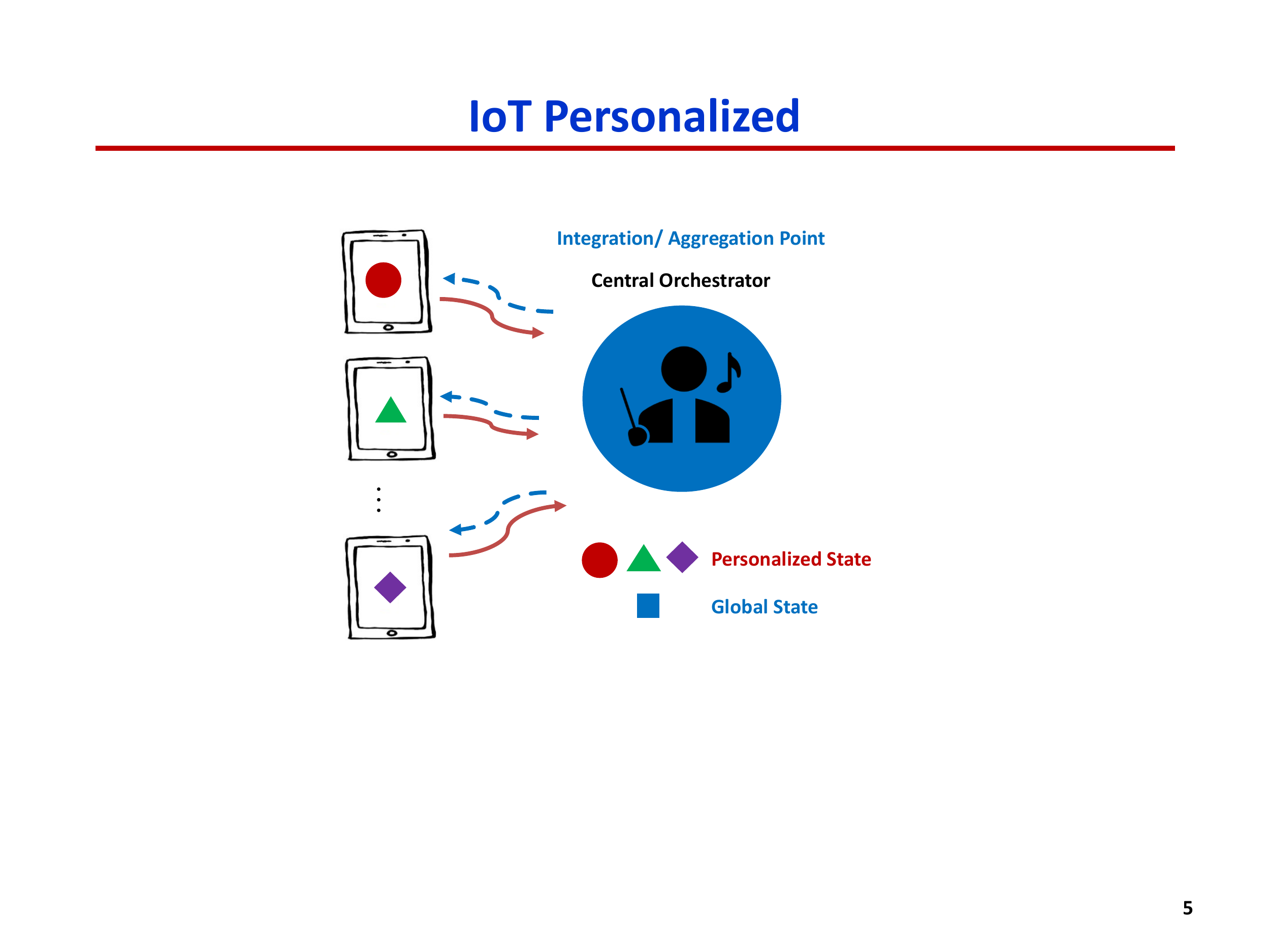}
	\caption{\label{fig:personalized} Personalized IoFT}
\end{figure}

As highlighted in previous sections, heterogeneity is a fundamental challenge for IoFT.  IoFT devices often exhibit highly heterogeneous trends and behaviors due to differences in operational, environmental, cultural, socio-economic and specification conditions \cite{kontar2017nonparametric,kontar2018nonparametric, yue2019variational}. For instance, in manufacturing, operational differences involve changes in the speed, load, or temperature a product experiences. As a result, data distribution across edge devices can be vastly heterogeneous, that one single global model cannot perform consistently well on all edge devices. This also has severe fairness implications as devices with limited data and unreliable connection will not be favored by many FL algorithms due to higher weights (recall \texttt{FedAvg} and its variants) given to devices with more data or those that can participate more often in the training process. Indeed, in the past few years, multiple papers have shown the wide gap in a global model's performance across different devices when  heterogeneity exists \cite{jiang2019improving,hard2018federated,wang2019federated,smith2017federated, kairouz2019advances}.

One straightforward solution to address the challenges above is through personalization. As shown in Fig. \ref{fig:personalized}, instead of using one global model for all edge devices, personalized FL fits tailor-made models for IoFT devices while levering information across all those devices. The rest of this section will discuss current personalization approaches, their drawbacks and potential alternatives. We divide the personalization techniques into fully personalized and semi-personalized. For fully personalized algorithms, each edge device retains its own individualized model, and for semi-personalized algorithms, models are tailor-made only to a group of clients. In Sec. \ref{sec:metalearning}, we will further discuss personalization from a meta-learning perspective.  

\subsection{\emph{Fully Personalized}}

From a statistical perspective, let $x_i \sim \mathbb{P}_x$, but let the conditional distributions of $y_i$ given $x_i$ vary across IoFT devices. One can write this as $y_i \sim \mathbb{P}^i_{y|x} \big(f_{\bm{\beta}_i}(x_i)\big)$ where clients share the same $f$ (a linear model, neural network) yet with different parameters $\bm{\beta}_i$. In this situation, the difference in the data distributions $\mathbb{P}^i_{x,y}=\mathbb{P}_x \mathbb{P}^i_{y|x}$ across clients can be explained by the difference in $\mathbb{P}^i_{y|x}$.  This is often referred to as a concept shift and implies a change in the input-output relationship across clients \citep{machinelearninginnonstationaryenvironments,mcmahan2021advances}. For example, in manufacturing the same design setting can have different effects on the manufactured product given external factors such as operational speed or load. Also, take the sequence prediction task on mobile phones as an example: for different users, the word following ``I live in ...'' should be different \citep{surveyonpersonalization}. This example corresponds to a concept shift: $x$ is assumed the given part of the sentence `I live in', and $y$ is the next word to predict. In this situation, $\mathbb{P}^i_{y|x}$ should be customized for different clients even if $x$ is the same.  

This section discusses current approaches to address a concept shift across clients, their drawbacks, and promising alternatives. Modeling for a shift in $\mathbb{P}_x$ is highlighted in our statistical perspective (Sec. \ref{sec:stat}).

To accommodate client specific concept shifts while leveraging global information one can extend the global FL model in (\ref{eq:obj}) to the following general objective for personalized FL: 
\begin{equation}
\label{eq:generalobjective}
\min_{\bm{w}, \, \bm{\beta}} F(\bm{w}, \bm{\beta})\coloneqq \frac{1}{N}\sum_{i=1}^NF_i(\bm{w}, \bm{\beta}_i) \, ,
\end{equation}

where $\bm{w}$ are shared global parameters while $\bm{\beta}=\{\bm{\beta}_i\}_{i=1}^N$ is a set of unique parameters for each client. 

The current literature aiming to address a concept shift can be broadly split into two categories: (i) weight sharing and (ii) regularization. \textbf{It will also become clear shortly that many current approaches follow a train-then-personalize philosophy which may be dangerous in some instances}.



\subsubsection{Weight Sharing}

The first set of literature solves (\ref{eq:generalobjective}) by using different layers of a neural network to represent $\bm{w}$ and $\bm{\beta}_i$ \cite{fedper,lg-fedavg}. The underlying idea is that base layers process the input to learn a shared feature representation across clients, and top layers learn task-dependent weights based on the features.

\texttt{FedPer} \citep{fedper} fits global base layers, and personalizes top layers. As an example, a fully connected multi-layer neural network can be expressed as $f_{\bm{w}}(x)=W_l\sigma\left(W_{l-1}\sigma\left(...\sigma\left(W_{1}x\right)\right)\right)$, where $l$ are the number of network layers. Recall from Sec. \ref{sec:heterogeniety}, $\sigma$ denotes an activation function and $W_j$'s are weight matrices. In this example, \texttt{FedPer} takes $W_1$ to $W_B$ as base layers that characterize $\bm{w}$, and $W_{B+1}$ to $W_n$ as personalized layers that characterize $\bm{\beta}_i$ in \eqref{eq:generalobjective}. In one communication round, client $i$ uses SGD to update $\bm{w}$ and $\bm{\beta}_i$ simultaneously. However, different from \texttt{FedAvg}, only $\bm{w}$ is transmitted to the server where it is then aggregated. \texttt{FedPer} is found to perform better than \texttt{FedAvg} on image classification tasks such as CIFAR-10 and CIFAR-100. On these datasets, the authors show that having the last one or two basic residual blocks of Resnet-34 personalized can yield the best testing performance. Similarly, \texttt{LG-FedAvg} \citep{lg-fedavg} takes top layers as a global weight $\bm{w}$ and base layers as personalized weights $\bm{\beta}_i$. The intuition is to learn customized representation layers for different clients, and to train a global model that operates on local representations. Additionally, by carefully designing the loss of representation learning, the generated local representation can confound protected attributes like gender, race, etc. 

\subsubsection{Regularization}
In contrast to splitting of global and local layers, other recent work treat neural networks holistically and learn personalized $\bm{\beta}_i$'s by exploiting regularization  \citep{mocha,multitask-fl,pfedme}.

Perhaps the most straightforward method to personalize via regularization is to follow a \textbf{train-then-personalize (TTP)} approach.  As the name suggests, this approach trains the global model on all clients then adapts it to individual devices. The simplest way for the adaptation is fine-tuning \citep{local-adaptation,finetune}, which is also widely employed in computer vision and natural language processing \citep{gpt2}. More specifically, in the TTP approach, we have a two step procedure. Step 1 - \textit{Train}: clients collaborate to train a global model $\bm{w}^*\overset{\Delta}{=}\bm{w}^T$ using \texttt{FedAvg} (or its variants) - recall $T$ is the last communication round. Step 2 - \textit{Personalize}: clients make small local adjustments based on their local data to personalize $\bm{w}^*$. Notice that for such methods, $\bm{\beta}_i$'s and $\bm{w}$ are in the same parameter space thus it's possible to perform addition or calculate the difference of these weight vectors. Weight regularizing methods thus usually allow all the weight vector to differ across clients, instead of forcing some coordinates of these weight vectors to be exactly the same.

A simple means for the personalization step is to start from $\bm{\beta}_i=\bm{w}^*$  and perform a few steps of SGD to minimize the local loss function $\min_{\bm{\beta}_i}F_i(\bm{\beta}_i)$. Indeed, this approach to fine-tuning  is shown to generalize better than fully local training or global modeling on next word prediction \cite{finetunebert} and image classification tasks (e.g. \cite{finetuneface,finetunecovid}). In this same essence, one may exploit regularization to encourage the weights of personalized models to stay in the vicinity of the global model parameters to balance each client's shared knowledge and unique characteristics. For instance, using ideas from \texttt{FedProx}, the personalization step can encourage $\bm{\beta}_i$ to remain within a vicinity of the global solution $\bm{w}^*$ as shown below.

\begin{equation}
\label{eqn::regularizedobjective}
\min_{\bm{\beta}_i} \left(F_i(\bm{\beta}_i)+\frac{\mu}{2}\left\|\bm{\beta}_i-\bm{w}^*\right\|^2\right) \, .
\end{equation}

Other forms of regularization can also be used. For instance, by employing the popular elastic weight consolidation model (\texttt{EWC}) \citep{kirkpatrick2017overcoming} that is often used in continual learning, we can control $\bm{\beta}_i$ as

\[
\min_{\bm{\beta}_i} \left(F_i(\bm{\beta}_i)+\frac{\mu}{2}\sum_j\mathcal{FI}_j\left\|\beta_j-w^*_j\right\|^2\right) \, ,
\]
where $\mathcal{FI}_j$ are diagonal elements of the Fisher information matrix. 


Some recent approaches \cite{ditto,pfedme} have exploited the ideas above but in an iterative manner, where local and global parameters in the train-then-personalize step are obtained by alternating optimization methods. Among them, \texttt{Ditto} \cite{ditto} simply proposed the following bi-level optimization problem for client $i$: 

\begin{equation} \label{eq:ditto}
\begin{aligned}
&\min_{\bm{\beta}_i}F_i(\bm{\beta}_i)+\frac{\mu}{2}\left\|\bm{\beta}_i-\bm{w}^*\right\|^2\\
&\text{s.t.} \, \bm{w}^*\in\arg\min_{\bm{w}}\frac{1}{N}\sum_{i=1}^NF_i(\bm{w}) \, .
\end{aligned}
\end{equation}

To solve this formulation, \texttt{Ditto} uses the following update rule.  In communication round $t$, clients  firstly receive a copy of global weight $\bm{w}^t$ which is updated to $\bm{w}^{t+1}_i$ using multiple (S)GD steps on the local risk function $F_i(\bm{w})$; much like \texttt{FedAvg}.  In the meantime, clients $i$ also obtains $\beta_i$ by multiple descent steps on the regularized loss (\ref{eq:ditto}):

\[
\bm{\beta}_i\leftarrow \bm{\beta}_i-\eta \nabla F_i(\bm{\beta}_i)-\eta\mu\left(\bm{\beta}_i-\bm{w}^t\right).
\]

At the end of the training round, client $i$ sends only global weight $\bm{w}^{t+1}_i$
back to server. Server simply averages received weights $\bm{w}^{t+1}=\frac{1}{N}\sum_{i=1}^N\bm{w}^{t+1}_i$. Empirically, \texttt{Ditto} has shown strong personalization accuracy on multiple commonly used FL datasets.

In \texttt{Ditto}, the global weight update is independent of personalized weights and follows the  \texttt{FedAvg} procedure. Hence global weights cannot learn from the performance of personalized weights. To integrate the update of $\bm{w}$ and $\bm{\beta}_i$, \cite{pfedme} proposes Moreau envelope FL (\texttt{pFedMe}) for personalization. \texttt{pFedMe} formulates the following bi-level optimization problem:
$$
\min_{\bm{w}}\frac{1}{N}\sum_{i=1}^N\min_{\bm{\beta}_i}\left[F_i(\bm{\beta}_i)+\frac{\mu}{2}\left\|\bm{w}-\bm{\beta}_i\right\|^2\right] \, .
$$

\sloppy
\texttt{pFedMe} gets its name because $\min_{\bm{\beta}_i}\left[F_i(\bm{\beta}_i)+\frac{\mu}{2}\left\|\bm{w}-\bm{\beta}_i\right\|^2\right]$ is the Moreau envelope of $F_i(\bm{w})$. In the inner level optimization, personalized weights $\bm{\beta}_i$ minimize the local risk function in the vicinity of reference point $\bm{w}$, and in the outer level minimization, $\bm{w}$ is minimized to produce a better reference point. This objective is closely related to model-agnostic meta-learning (MAML). 
Sec \ref{sec:metalearning} will cover more details about meta-learning algorithms. The optimal solution of \texttt{pFedMe} satisfies the relation $\bm{\beta}_i^*(\bm{w})=\bm{w}-\frac{1}{\mu}\nabla F_i(\bm{\beta}_i^*)$. Through some calculus, the gradient with respect to $\bm{w}$ is: $\nabla_{\bm{w}}[F_i(\bm{\beta}_i^*(\bm{w}))+\frac{\mu}{2}\left\|\bm{w}-\bm{\beta}_i^*(\bm{w})\right\|^2]=\mu(\bm{\beta}_i^*(\bm{w})-\bm{w})$. In practice, at one communication round, selected clients receive global weight $\bm{w}$ and find an approximate optimal solution $\tilde{\bm{\beta}_i^*}$ to the inner optimization problem $\min_{\bm{\beta}_i}\left[F_i(\bm{\beta}_i)+\frac{\mu}{2}\left\|\bm{w}-\bm{\beta}_i\right\|^2\right]$ via SGD or its variants. Client $i$ then multiplies the update by $\mu\eta_l$, $\Delta\bm{w}_i=\mu\eta_l(\tilde{\bm{\beta}_i^*}-\bm{w})$, and sends $\Delta \bm{w}_i$ to the server. $\mu$ is set to $15\sim30$ in the case studies. The server then averages received $\Delta \bm{w}_i$ to renew the global model as $\bm{w}\leftarrow \bm{w}+\eta_g \sum_i\Delta\bm{w}_i$. $\eta_g$ is another hyperparameter whose value is $1\sim 4$ in experiments. With proper hyper-parameter choices, the convergence rate of \texttt{pFedMe} is proved to be faster compared with \texttt{FedAvg}.

Local loopless GD (\texttt{L2GD}) \citep{lgd} simply sets $\bm{w}$ to be the average of $\bm{\beta}_i$, $\bm{w}=\overline{\bm{\beta}}=\frac{1}{N}\sum_{i=1}^N\bm{\beta}_i$. The objective is:

\[
\min_{\bm{\beta}}\frac{1}{N}\sum_{i=1}^N\left[F_i(\bm{\beta}_i)+\frac{\mu}{2}\left\|\bm{\beta}_i-\overline{\bm{\beta}}\right\|^2\right].
\]
To optimize the objective, \texttt{L2GD} chooses to update $\frac{1}{N}\sum_{i=1}^NF_i(\bm{\beta}_i)$ and $\frac{1}{N}\sum_{i=1}^N\frac{\mu}{2}\left\|\bm{\beta}_i-\overline{\bm{\beta}}\right\|^2$ by a random experiment. More specifically, at round $t$, with probability $1-p$, each client will perform one step of GD to minimize local training loss $\bm{\beta}_i^t$ as $\bm{\beta}_i^t\leftarrow \bm{\beta}_i^t-\frac{\eta}{N(1-p)}\nabla F_i(\bm{\beta})$. With probability $1-p$, clients will send updated models $\bm{\beta}_i^t$ back to the server, unless the model is not updated since the previous synchronisation. The \textit{server\_update} consists of two steps: first, the server takes the average of $\bm{\beta}_i^t$ to obtain $\overline{\bm{\beta}^t}=\frac{1}{N}\sum_{i=1}^N\bm{w}_i^t$; second, the server calculates the initialization of client $i$'s weight on the $t+1$-th communication round as $\bm{\beta}_i^{t+1}=\left(1-\frac{\alpha \mu}{Np}\right)\bm{\beta}_i^t+\frac{\alpha \mu}{Np}\overline{\bm{\beta}^t}$ and sends it to the corresponding client. Here $\alpha$ is a tunable hyperparameter. 

Finally we shed light on another formulation of the train then personalize approach provided by \cite{deng2020adaptive}. \cite{deng2020adaptive} learn the global parameter $\bm{w}^*$ using traditional global modeling approaches, yet, in the personalization step the  local objective is given as  

\begin{equation}
    \min_{\bm{\beta}_i} F_i(\zeta_i\bm{\beta}_i+(1-\zeta_i)\bm{w}^*),
\end{equation}

The tuning parameter $\zeta_i$ balances the importance of the local and global model. When users data are \textit{i.i.d.}, the authors argue that the global model will have better generalization and suggest choosing a smaller $\zeta_i$. On the other hand, when data are heterogeneous, they choose to use a larger $\zeta_i$ to encourage personalization.

\subsubsection{A counter-example}

The rationale behind regularization approaches is that the global model learns shared knowledge. Then, by encouraging local weights to stay close to the global model, every client can borrow strength from this shared knowledge. Unfortunately, this simple intuition does not apply to all problems. As a simple counterexample, assume client $i$'s ground truth is a sine function:

$$
f_{\theta_i}(x)=\sin(2\pi (x+\theta_i)) \, ,
$$

where $\theta_i$ is client-dependent. We assume that among clients, $\theta_i$ admits a uniform distribution on $[0,1]$. Sine functions are the basis of almost all periodic functions and a phase shift is common for vibration signals, which usually have strong similarity and large shifts. 

\begin{figure}[!htb]
	\centering
	\includegraphics[width=0.25\textwidth]{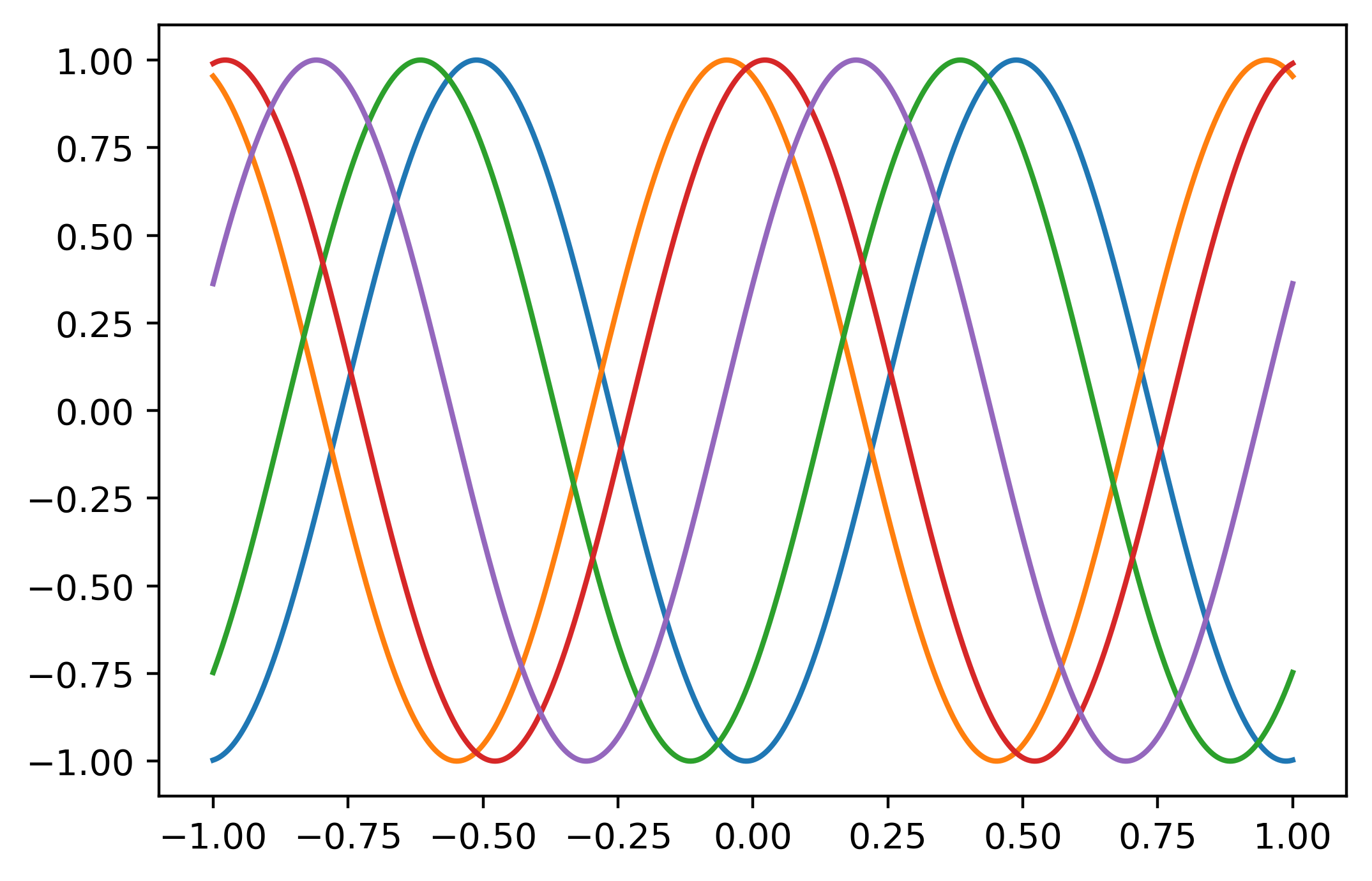}
	\caption{\label{fig:sin} Underlying true sine functions from five different clients. }
\end{figure}

Now, if we train a global model to minimize the population risk:
$$
\min_{\bm{w}}\mathbb{E}_i[||f_{\bm{w}}-f_{\theta_i}||_2^2] \, .
$$
where $f_{\bm{w}}$ is a global model parametrized by weight $\bm{w}$, and $||\cdot||_2$ is a functional on $[0,1]$ defined as:
$$
||f||_2^2=\int_{0}^{1}f(x)^2dx \, .
$$
Then $f_{\bm{w}}$ should minimize:
$$
\begin{aligned}
&\arg\min_{f_{\bm{w}}}\mathbb{E}_{\theta_i}\left[\int_{0}^{1}\left(f_{\bm{w}}(x)-\sin(2\pi x+2\pi\theta_i)\right)^2dx\right]\\
&=\arg\min_{f_{\bm{w}}}\mathbb{E}_{\theta_i}\left[\int_{0}^{1}f_{\bm{w}}(x)^2-2f_{\bm{w}}(x)\sin(2\pi x+2\pi\theta_i)dx\right]\\
&=\arg\min_{f_{\bm{w}}}\mathbb{E}_{\theta_i}\left[\int_{0}^{1}f_{\bm{w}}(x)^2dx\right].\\
\end{aligned}
$$
Therefore, the unique minimizer is $f_{\bm{w}}(x)=0$ for every $x$ in $[0,1]$. Clearly, a global model does not learn anything from such a dataset. Now, assume there exists such a weight vector $\bm{w}_{zero}$ such that $f_{\bm{w}_{zero}}=0$. Now examine the performance of personalized FL models. \texttt{Ditto}'s local objective in (\ref{eq:ditto}) becomes a local empirical risk plus a regularization: $$\int_0^1\left(f_{\bm{\beta}_i}(x)-\sin\left(2\pi x+2\pi\theta_i\right)\right)^2dx+\frac{\mu}{2} ||\bm{\beta}_i-\bm{w}_{zero}||^2.$$

Not only no useful information about common patterns in the sine function is shared among clients, the regularization will further exacerbate the problem by forcing $\bm{\beta}_i$ close to $\bm{w}_{zero}$ which is clearly a bad point as its predicts a zero everywhere on the function. Similar phenomena can be witnessed on other regularization based 
train-then-personalize approaches such as \texttt{L2GD}.

In the example above, data across clients have strong commonalities as they all share the same functional form with only one parameter $\theta$ explaining their variations. Nevertheless, a train-then-personalize approach or its iterative counterparts fail due to a faulty global model. So what are potential alternative approaches ?

\subsubsection{Alternative solutions}

One approach to circumvent the need for a global model in FL is through multi-task learning (MTL) which aims to leverage commonalities across different but related outputs to improve prediction and learning accuracy \citep{caruana1997multitask}. In MTL, a shared representation across all tasks is built to allow the inductive transfer of knowledge \citep{yuan2012visual, kontar2017nonparametric}. Here each task is a client/edge device.

\begin{figure}[!htb]
	\centering
	\includegraphics[width=0.48\textwidth]{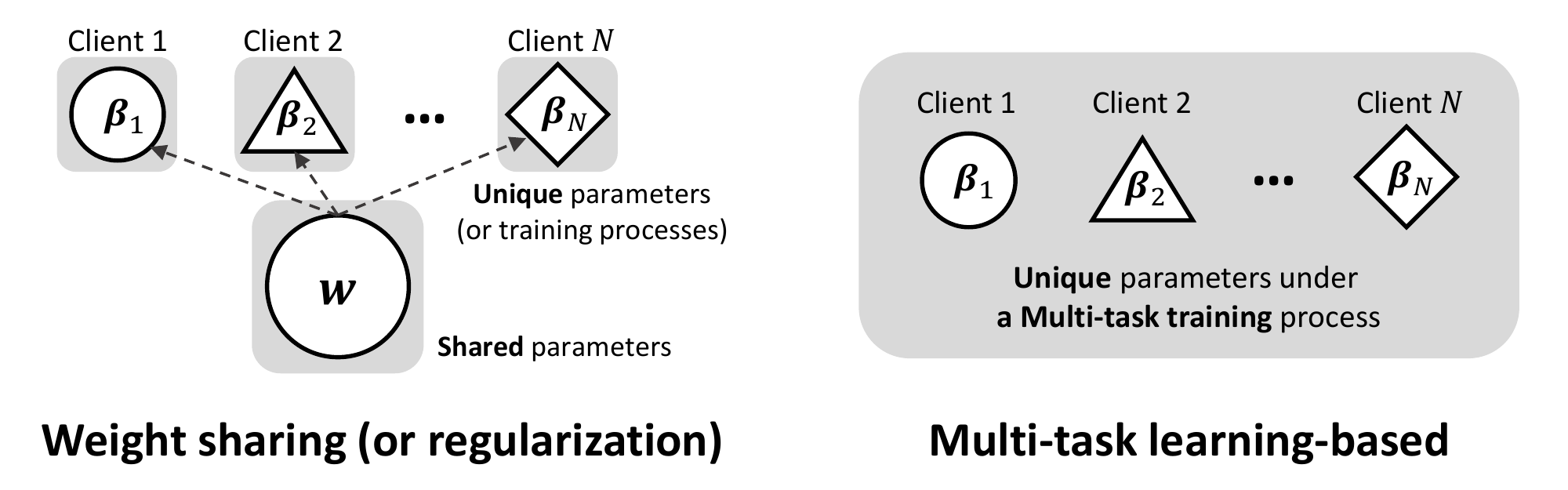}
	\caption{\label{fig:mtl} A comparison between train-then-personalize and multi-task learning approaches to personalization in IoFT.}
\end{figure}

An illustrative example of MTL vs train-then-personalize is shown in Fig. \ref{fig:mtl}. The main difference is that MTL directly estimates $\bm{\beta}_i$ without the need to learn a global model $\bm{w}$.

Often, the shared representation in MTL is induced via regularization to facilitate information transfer across tasks. This can be written as 
\begin{align*}
    \min_{\bm{\beta},\Omega}\left\{\frac{1}{N}\sum_{i=1}^NF_i(\bm{\beta}_i)+\mathcal{R}(\mathcal{B},\Omega)\right\} \, ,
\end{align*}

where $\bm{\beta}_i$ are the personalized weights, $B$ is a matrix whose $i$th row is $\bm{\beta}_i$, $\mathcal{R}$ is a regularization term and $\Omega$ is an $N\times N$ matrix that models relationships amongst clients.

The formulation above has been studied extensively in centralized regimes \citep{pontile2007convex, kumar2012learning, kang2011learning, zhang2012convex, gonccalves2016multi}. Yet literature on MTL in FL is still very limited. One of such approaches is  \texttt{MOCHA} \citep{mocha} which defines the objective below
\[
\min_{\bm{\beta},\Omega}\sum_{i=1}^N\ell(\bm{\beta}_i^\mathtt{T} x_i,y_i)+\mu_1\text{Tr}\left(\mathcal{B}^\texttt{T}\Omega\mathcal{B}\right)+\mu_2\left\|\mathcal{B}\right\|_F^2 \, ,
\]
where, the first term defines a loss function for linear models, $\text{Tr}\left(\mathcal{B}^\texttt{T}\Omega\mathcal{B}\right)$ induces knowledge transfer such that negative off-diagonal entries of $\Omega$ encourages the alignment of two clients' local weights and $\left\|\cdot\right\|_F$ represents a Frobenius norm for shrinkage. \citep{mocha} extends the primal-dual formulation in \citep{zhang2012convex} to distribute the update of $\mathcal{B}$ over clients while keeping $\Omega$ fixed. \texttt{MOCHA} is shown to outperform baseline algorithms on datasets including GLEAM, Human Activity Recognition, and Vehicle Sensors.

Although the idea of \texttt{MOCHA} is intriguing the objective is only confined to the problems whose losses are convex, since the authors use dual algorithms to solve it and strong duality is not guaranteed for general non-convex functions. Here, future work that exploit MTL to learn a graphical model through $\Omega$ may open interesting new research directions in understanding the underlying commonality structure across clients. 

It is also worth noting that other alternative routes can be taken instead of MTL. One such route is to first validate the performance of a the global model prior to personalization. Along this line, \texttt{Fed-ensemble} \cite{Fedensemble} proposes an alternative approach for personalization via the learned $K$ global models described in Sec. \ref{sec:heterogeniety}. After training a diverse group of $K$ models, \texttt{Fed-ensemble} evaluates the loss of model $k$ on a local validation dataset $D^{\mbox{val}}_i$ by $\hat{F}_{k,i}=\frac{1}{n_i}\sum_{j=1}^{n_i}\ell(f_{\bm{w}_k}(x_{i,j}),y_{i,j})$.  Then each model $\bm{w}_k$ is assigned a weight $\alpha_{k, i}=\exp (-\gamma \hat{F}_{k,i})/\sum_{k=1}^K\exp (-\gamma \hat{F}_{k,i})$ where $\gamma$ is some constant. The prediction on a new sample $x^*$ is given by a weighted ensemble $\sum_{k=1}^K\alpha_{k,i}f_{\bm{w}_k}(x^*)$. The intuition is to check which models perform well on a local dataset, then assigns higher weights $\alpha_{k, i}$ to better fitted models in the ensemble.






\subsection{\emph{Semi-personalized}} \label{sec:semi}
\begin{figure}[!htb]
	\centering
	\includegraphics[width=0.4\textwidth]{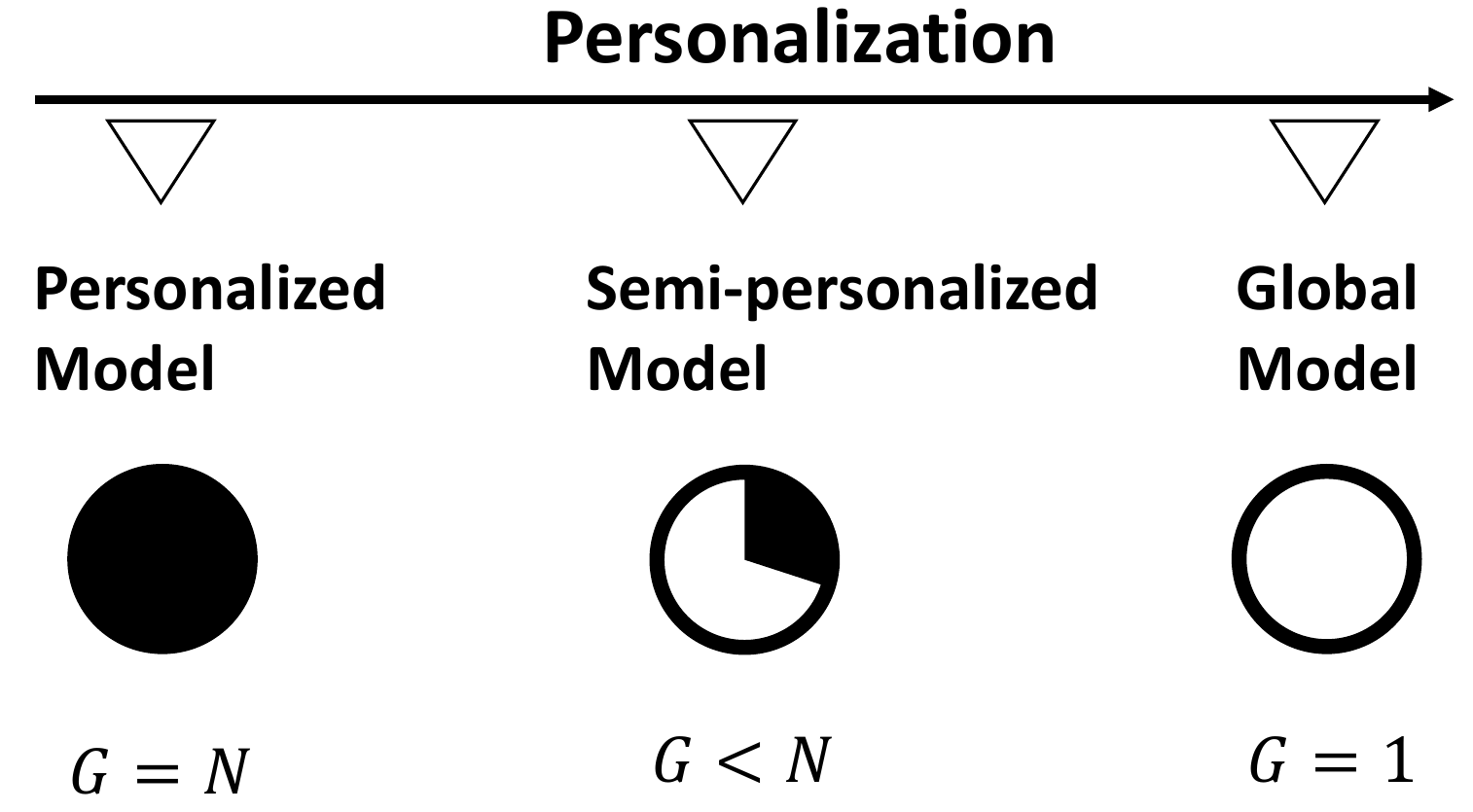}
	\caption{\label{fig:semipersonalized} Degrees of personalization }
\end{figure}

A possible alternative to global or fully personalized modeling is semi-personalized modeling. Semi-personalized FL fits a stylized model for a group of clients. This approach balances between the need for $N$ individualized models or one model that fits all. This is highlighted in Fig. \ref{fig:semipersonalized} . Usually these algorithms cluster clients into $G << N$ groups and assume the data distribution of clients inside one group is homogeneous.

\cite{threepersonalizationapp} proposed an intuitive user clustering method. The number of clusters $G<N$ is predetermined. Each cluster is represented by a cluster parameter in $\{\bm{w}_{1},...\bm{w}_{G}\}$. The objective is: 
\[
\min_{\{\bm{w}_{1},...\bm{w}_{G}\}}\sum_{i=1}^N\min_{j\in1,\cdots,G}F_i(\bm{w}_{j})
\]
The inner level of minimization taken over cluster index $j$ assigns one client $i$ to the cluster with the lowest training loss, and the outer level of minimization taken over $\bm{w}_j$'s optimizes cluster model weights. Authors propose a \texttt{HypCluster} algorithm to optimize the objective. 
In \texttt{HypCluster}, $G$ is predetermined. In each communication round, all cluster weights are broadcasted to all clients, and each client chooses one with the lowest loss. Afterwards, clients train the corresponding model by SGD on their own local dataset. On EMNIST, \texttt{HypCluster} with $G=2$ has higher test accuracy than \texttt{FedAvg} and \texttt{AFL} \citep{agnosticfl}.

Clustered FL (\texttt{CFL}) \citep{clusterfed} clusters clients dynamically. \texttt{CFL} measures the similarity between two clients in the following manner: suppose in communication round $t$, the update of clients $i$ and $j$ is $\Delta \bm{w}_i^t$ and $\Delta \bm{w}_j^t$, respectively. The cosine similarity is defined as $\alpha_{ij}=\frac{\left\langle\Delta \bm{w}_i^t,\Delta \bm{w}_j^t\right\rangle}{\left\|\Delta \bm{w}_i^t\right\|\left\|\Delta \bm{w}_j^t\right\|}$. For one communication round, server calculates the cosine similarity intra and across clusters. If clients in one cluster are homogeneous, the similarity of these clients should be large compared with similarity across clusters, then \texttt{CFL} simply performs \texttt{FedAvg} for this cluster. If otherwise, clients inside one cluster are heterogeneous, the similarity of these clients is low, and \texttt{CFL} will divide them into two subclusters. \texttt{CFL} then repeats the procedure on the two subclusters. As the algorithm proceeds, clients can automatically be divided into different subclusters. \texttt{CFL} ends when gradient norms on all clients are small and no further sub-dividing is needed. 


Semi-personalized modeling and client clustering in IoFT are important questions that still require much further investigation. In the statistical perspective (Sec. \ref{sec:statopt}) we pose some open questions that are critical along this research direction. 

\section{Meta-Learning} \label{sec:metalearning} 

Meta-learning is the science of observing how different learning algorithms perform on a wide range of tasks and then learning new tasks more efficiently based on prior experience \citep{vanschoren2018meta, chen2019closer, hospedales2020meta, munkhdalai2017meta, nichol2018first, ravi2017optimization}. Meta-learning tries to learn a global model that can quickly adapt to a new task with only a few training samples and optimization steps. This process is also known as “Learning to Learn Fast,'' where the goal of the model is not to perform well on all tasks in expectation, instead \textbf{to find a good initialization} that can directly adapt to a specific task. Therefore, meta-learning can be viewed as an approach to enable fast personalization and fine-tuning. 

Meta-learning opens a unique opportunity to resolve many challenges in IoFT, such as scalability, fast adaptability,  and improved generalization. With proper prior knowledge, a well-trained task can be rapidly generalized to new tasks with few samples. This few-shot property becomes especially crucial in IoFT where each device only has a small amount of data and does not have access to a centralized repository of all datasets.





\subsection{\emph{Frequentist Perspective}}
From a frequentist perspective, the seminal work of \citet{finn2017model} proposed one of the first model-agnostic meta-learning (MAML) algorithms. Given $N$ different tasks $\{\mathcal{T}_i\}_{i=1}^N$, MAML optimizes a meta-loss function in the form of
\begin{equation} \label{eq:meta}
    \min_{\bm{w}}\sum_{i=1}^N F_{\mathcal{T}_i}(\bm{w}-\eta\nabla F_{\mathcal{T}_i}(\bm{w})).
\end{equation}
In (\ref{eq:meta}), $\bm{w}$ can be viewed as an initialization used to perform one gradient update and obtain $\bm{w}'=\bm{w}-\eta\nabla F_{\mathcal{T}_i}(\bm{w})$. MAML minimizes the objective $\sum F_{\mathcal{T}_i}(\bm{w}')$ given an initial parameter $\bm{w}$. Therefore, the goal of MAML is to find an optimal initial parameter $\bm{w}^*$ such that one gradient step on a new task can incur maximally effective behavior on that task. This idea is similar to fine-tuning in which several steps of gradients are performed given a tuned parameter $\bm{w}^*$. 


The idea of meta-learning can be naturally extended to FL where each device is treated as a task and the goal is to learn an initialization for fast personalization on the edge devices. Indeed, recently researchers have tried to introduce the notion of meta-learning to FL to enable personalization and few-shot learning \citep{li2020federated, mcmahan2021advances}. Along this line, \cite{chen2018federated}, \cite{lin2020real} and \cite{fallah2020personalized} introduce MAML to FL. They reformulate the global objective of FL as \eqref{Per-FedAvg}. In the meta-learning literature, this global objective function is also known as meta-loss. In this section, we will use them interchangeably.
\begin{equation}
\label{Per-FedAvg}
\min_{\bm{w}} \frac{1}{N}\sum_{i=1}^NF_i(\bm{w}-\eta\sum_{e=1}^{E}\nabla F^{e}_i(\bm{w})).
\end{equation}
Here, $\nabla F^{e}_i(\bm{w})$ is the gradient after $e$ steps of SGD. Recall that $t$ refers to the communication round, while $e\in \{0,\cdots,E-1\}$ denotes the optimization iterates at the edge device. It is easy to see that \eqref{Per-FedAvg} is different from the conventional objective function \eqref{eq:obj}. The formulation \eqref{Per-FedAvg} allows each user to exploit $\bm{w}$ as an initial point and update it with respect to its local data (e.g., running $E$ steps of gradient descent). When $E=1$, the above objective function can be simplified as
\begin{equation}
\frac{1}{N}\sum_{i=1}^NF_i(\bm{w}-\eta\nabla F^{1}_i(\bm{w}))=\frac{1}{N}\sum_{i=1}^NF_i(\bm{w}-\eta\nabla F_i(\bm{w}))
\end{equation}
and the gradient on device $i$ can be computed as
\begin{equation}
\label{eq:second_order}
\begin{split}
    &\nabla F_i(\bm{w}-\eta\nabla F_i(\bm{w}))\\
    &=\big(I-\eta\nabla^2F_i(\bm{w})\big)\nabla F_i(\bm{w}-\eta\nabla F_i(\bm{w})).
\end{split}
\end{equation}
Based on this formulation, \cite{fallah2020personalized} propose the \texttt{Per-FedAvg} algorithm to efficiently optimize \eqref{Per-FedAvg}. Given an initial parameter $\bm{w}^t$, local user $i$ runs $E$ steps of SGD. For simplicity, we assume $E=1$. Therefore,
\begin{align}
\label{eq:local_up}
    \bm{w}_i^t = \bm{w}^t - \eta_1\nabla F_i(\bm{w}^t), 
\end{align}
where $\eta_1$ is the learning rate for each local user. Afterwards, each local user calculates the gradient evaluated at $\bm{w}_i^t$ and the Hessian evaluated at $\bm{w}^t$. The user then sends the gradient $\nabla F_i(\bm{w}_i^t)$ and the Hessian $\nabla^2 F_i(\bm{w}^t)$ back to the central server. The central server aggregates them as:
\begin{align*}
    \bm{w}^{t+1}=\bm{w}^t-\eta_2\frac{1}{N}\sum_{i=1}^N\big(I-\eta_1\nabla^2F_i(\bm{w}^t)\big)\nabla F_i(\bm{w}^t_i),
\end{align*}
where $\eta_2$ is a learning rate on the central server. This procedure is repeated several times till some exit conditions are met. One notable thing is that, in the central server, the gradient is evaluated at $\bm{w}_i^t$ rather than $\bm{w}^t$. This naturally arises from the meta-loss function \eqref{Per-FedAvg}. \cite{fallah2020personalized} demonstrate the advantages of \texttt{Per-FedAvg} on image classification tasks. Interestingly, \cite{jiang2019improving} interpret \texttt{FedAvg} as the linear combination of \texttt{FedSGD} and MAML when ignoring the second-order term $\eta\nabla^2F_i(\bm{w})$ in \eqref{eq:second_order}. Specifically, assume each client runs $E$ steps of local updates, then the gradient of \texttt{FedAvg} can be written as
\begin{align}
\label{linear_comb}
    &g_{FedAvg} \nonumber\\
    &\coloneqq \frac{1}{N}\sum_{i=1}^N\nabla F_i(\bm{w})=\frac{1}{N}\sum_{i=1}^N\sum_{t=1}^{E}\nabla F^{e}_i(\bm{w}) \nonumber \\
    &=\sum_{i=1}^N\frac{1}{N}\nabla F^{1}_i(\bm{w}) + \sum_{i=1}^N\frac{1}{N}\sum_{e=2}^{E}\nabla F^{e}_i(\bm{w}) \nonumber\\
    &=\sum_{i=1}^N\frac{1}{N}\nabla F^{1}_i(\bm{w}) + \sum_{i=1}^N\frac{1}{N}\sum_{e=2}^{E}\nabla F_i(\bm{w}-\eta\sum_{j=1}^{e-1}\nabla F^{j}_i(\bm{w}))
    \nonumber \\
    &= g_\texttt{FedSGD} + \sum_{e=2}^{E} g_{MAML}(e-1),
\end{align}
where $g_\texttt{FedSGD}$ is the gradient of \texttt{FedSGD} and $g_{MAML}(e)$ is the gradient of MAML with $e$ steps of local updates. Though the optimization of this linear combination in \eqref{linear_comb} is not strictly equivalent to \texttt{FedAvg} (due to the second-order term), this interpretation sheds light on the intrinsic connection between FL and meta-learning.

Based on this observation, \cite{jiang2019improving} slightly modify the \texttt{Per-FedAvg} algorithm as follows: first run \texttt{FedAvg} (or another conventional FL algorithm) at the early stage of training and then switch to MAML (in \eqref{Per-FedAvg}) or Reptile \citep{nichol2018reptile} to fine-tune the model. Through many empirical results, the authors argue that this combined strategy ensures fast and stable convergence compared to directly optimizing the federated MAML objective in \eqref{Per-FedAvg}. Besides, this paper also delivers an important message: no single FL is a panacea for all problems, instead, differnet dataset requires different inference strategies or a combination of them. 

There are also few other variants of meta-learning-algorithms that can be readily applied to FL. For instance, \texttt{MetaSGD} \citep{li2017meta, chen2018federated} specifies a coordinate-wise learning rate $\bm{\eta}$. Instead of constraining the learning rate to be a fixed positive scalar, they define $\bm{\eta}$ as a vector that is of the same size as $\bm{w}$ and each element in it can be positive, negative or zero. Different from the traditional definition of a learning rate, $\bm{\eta}$ encodes both the update direction and rate. To understand the intuition behind $\bm{\eta}$, it is very helpful to first see the \texttt{MetaSGD} procedure in detail. In \texttt{MetaSGD}, both $\bm{w}$ and $\bm{\eta}$ are treated as model parameters to optimize. Specifically, at communication round $t$, \texttt{MetaSGD} first samples $\mathcal{S}\subseteq [N]$ clients and divides the data for each client into a training set $D^{\text{train}}$ and validation set $D^{\text{val}}$. Here, \texttt{MetaSGD} then runs one step of SGD using $\bm{\eta}^t$ and $\bm{w}^t$ on each sampled client to obtain updated parameters $\{\bm{w}^t_i\}_{i=1}^N$:
\begin{align*}
    \bm{w}^t_i \leftarrow \bm{w}^t - \bm{\eta}^t\circ\nabla F_i(\bm{w}^t)
\end{align*}
where $\circ$ is a element-wise product operation and $F_i(\bm{w}^t)$ is the loss function evaluated on the training set at communication round $t$. Afterward, \texttt{MetaSGD} updates all model parameters as
\begin{align*}
    (\bm{w}^{t+1},\bm{\eta}^{t+1})\leftarrow(\bm{w}^t,\bm{\eta}^t)-\eta_{\texttt{MetaSGD}}\frac{1}{N}\sum_{i=1}^N\nabla F^\text{val}_{i}(\bm{w}^t_i)
\end{align*}
where $\eta_{\texttt{MetaSGD}}$ is a scalar learning rate and $\nabla F^\text{val}_{i}(\bm{w}^t_i)$ is the local gradient evaluated on the validation set $D^{\text{val}}$. Note that $\bm{w}^t_i$ is a function of both $\bm{w}^t$ and $\bm{\eta}^t$ since $\bm{w}^t_i \leftarrow \bm{w}^t - \bm{\eta}^t\circ\nabla F_i(\bm{w}^t)$. Therefore, the gradient of $F_{i}(\bm{w}^t_i)$ can be taken with respect to $\bm{w}^t$ and $\bm{\eta}^t$. From this algorithm, one can see that $\bm{\eta}$ is learned from all tasks to avoid possible model over-fitting on a specific task. The intuition is that the gradient $\nabla F_i(\bm{w}_i)$ on the training set can potentially lead to over-fitting, especially when the sample size is small. Therefore, $\bm{\eta}$ acts as a regularization role to control the sign and magnitude of $\nabla F_i(\bm{w}_i)$.

\subsection{\emph{Bayesian Perspective}}
\label{sec:5-2}
Besides the frequentist perspective on meta-learning, there are recent yet few efforts to explore Bayesian meta-learning. Below we discuss some recent advances. We note that the works discussed below focus on meta-learning and not FL. However as shown earlier those concepts are naturally related. We will also provide examples on how to extend the models to FL. 

Bayesian meta-learning simply defines $\bm{w}$ as a random variable and takes a Bayesian route to estimate its posterior. This posterior then serves as a ``prior'' for a new task. Such an approach would allow uncertainty quantification for predictions on both the central server and local clients \citep{wang2019bayesian}. Notably, \cite{yoon2018bayesian} propose a Bayesian counterpart of MAML (BMAML) for fast adaption and uncertainty quantification. 


Instead of finding a single parameter to minimize \eqref{Per-FedAvg}, BMAML aims at finding a posterior distribution of the parameter such that one can quantify uncertainties. To achieve so, the gradient method used in MAML is replaced by one of its Bayesian counterparts - Stein variational gradient descent (SVGD) \citep{liu2016stein}. The SVGD method combines the strengths of SGD and Markov chain Monte Carlo (MCMC) such that one can sample from a posterior distribution to quantify uncertainties. As described in \cite{liu2016stein}, SVGD maintains $M$ instances of model parameters, called particles. Those particles can be viewed as samples from the posterior distribution of the model parameter. 

Here we detail the BMAML algorithm. At the global optimization iterate $t$, which is equivalent to the communication round in the FL, BMAML starts with $M$ initial particles $\{\bm{w}^{t}_{m}\}_{m=1}^M$ that are sent to the clients. Each client then applies $E$ steps of SVGD to obtain updated parameters as shown below:
\begin{align*}
    \{\bm{w}^{t,E}_{i,m}\}_{m=1}^M=SVGD( \{\bm{w}^{t}_{m}\}_{m=1}^M;D_i,\eta) \, , 
\end{align*}

where $\eta$ and $SVGD(\{\bm{w}^{t}_{i,m}\}_{m=1}^M;D,\eta)$ is the SVGD algorithm that aims to collect samples from the posterior $\mathbb{P}(\bm{w}|D_i)$ during local training.

Afterwards, the updated task-dependent parameters $\{\bm{w}^{t,E}_{i,m}\}_{i\in[N],m\in[M]}$ are used to calculate the gradient of the meta-loss \sloppy $\sum_{i}\nabla F_i(\{\bm{w}^{t,E}_{i,m}\}_{m=1}^M)$ and perform a one step update
\begin{align*}
    \{\bm{w}^{t+1}_{m}\}_{m=1}^M \leftarrow  \{\bm{w}^{t}_{m}\}_{m=1}^M-\eta\frac{1}{N}\sum_{i}\nabla F_i(\{\bm{w}^{t,E}_{i,m}\}_{m=1}^M).
\end{align*}
Recall that the gradient of the meta-loss is evaluated at the updated particles $\{\bm{w}^{t,E}_{i,m}\}_{m=1}^M$ and the global optimization is performed over $\{\bm{w}^{t}_{m}\}_{m=1}^M$. The overall idea of BMAML is very similar to MAML (Eq. \eqref{Per-FedAvg}): the goal is to train a model that can quickly adapt to a new task.

Besides BMAML and its variants, there are also many other studies that formulate meta-learning from a Bayesian perspective \cite{edwards2017towards, bauer2017discriminative,lacoste2017deep, lacoste2018uncertainty, grant2018recasting, yoon2018bayesian, finn2018probabilistic,gordon2018meta,ravi2019amortized, tossou2019adaptive, rusu2019meta, patacchiola2020bayesian, nguyen2020pac, zou2020gradient}. Most notably, \cite{tossou2019adaptive, patacchiola2019deep} consider applied deep kernel methods to learn complex task distributions in a few-shot learning setting. \cite{lacoste2018uncertainty, wu2018meta} introduce deep parameter generators that capture a wide range of parameter distributions. However, this method is not amenable to first-order stochastic optimization methods and therefore may scale poorly. 

The works discussed above focus on meta-learning and not FL. However as shown earlier those concepts are naturally related. To give an example, in the Bayesian MAML, each local user $i$ can be treated as a task $i$. During training, each local user performs SVGD to obtain $M$ updated weight parameters $\{\bm{w}^{t,E}_{i,m}\}_{m=1}^M$. The local user $i$ then sends the gradient of local loss evaluated at $\{\bm{w}^{t,E}_{i,m}\}_{m=1}^M$ to the central server. The central server collects information from all users and updates the meta-loss function to obtain $\{\bm{w}^{t+1}_{m}\}_{m=1}^M$. At the end of each communication round, the central server broadcasts $\{\bm{w}^{t+1}_{m}\}_{m=1}^M$ to all devices. This cycle is repeated until some exit conditions are met.

\begin{figure}[htbp]
\centering
\includegraphics[width=0.5\textwidth]{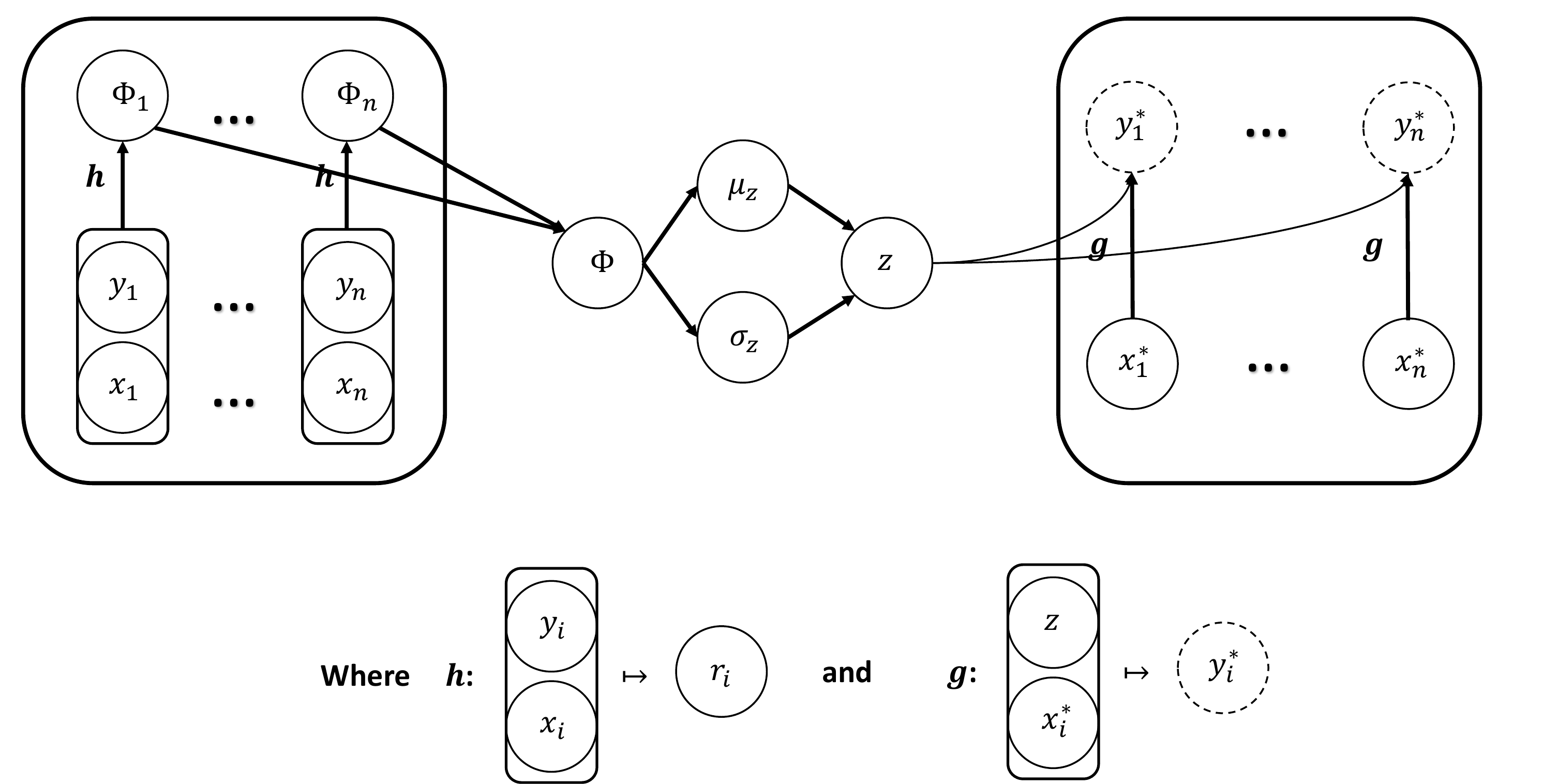}
\caption{The diagram of neural processes.}
\label{fig::neural_process}
\end{figure}

Interestingly, there is also a trend that formulates meta-learning from a stochastic process perspective \citep{garnelo2018conditional, kim2019attentive, ma2019variational,louizos2019functional}. Most notably, \cite{garnelo2018neural} introduce neural processes (NPs) that combine advantages of neural networks and Gaussian processes (GPs). Fig. \ref{fig::neural_process} illustrates the idea behind NPs. Given a point $(x_i,y_i)$, the algorithm first defines a representation function $\Phi_i=h((x_i,y_i))$ that maps inputs into a feature space. Here $\Phi$ is a NN. Then it defines a latent distribution over the feature representation. Specifically, $z\sim\mathcal{N}(\mu_z(\Phi),I\sigma_z(\Phi))$ where $\Phi=a(\{\Phi_i\}_{i=1}^N)=\frac{1}{n}\sum_i \Phi_i$ and $a$ is called the aggregator. In turn this latent distribution mimics a GP as it defines a prior over $\Phi$ \citep{williams2006gaussian}. Now, given this latent distribution a conditional decoder $g$, learned through variational inference, takes inputs sampled from $z$ and testing data $x^*$ to make predictions $y^*$.  One key feature of NP is that it encodes input data into a single order-invariant global representation. This representation captures the global uncertainty which allows sampling at a global level. The similarity to meta-learning in NPs is the fact that we are learning a prior over the latent representation. This prior then acts as an starting point for predictions at a new point.




To extend NP to a federated framework, one can develop some aggregation strategies that allow $N$ edge devices to collaboratively learn the global feature $z$. For example, all edges devices can use \texttt{FedAvg} to train a global encoder $h$. This shared encoder is then used to calculate the global latent distribution $z$.


As shown above, various meta-learning approaches may be readily extended to FL.  We hope this review will help inspire continued exploration along this direction as we envision that meta-learning will have a great impact in IoFT. Further, exploring and contrasting of  meta-learning models in FL may help guide practitioners chose amongst existing methods given the data properties and features.

\section{Statistical and Optimization Perspective} \label{sec:statopt}

\subsection{\emph{Statistical Perspective}} \label{sec:stat}

Much of the current work on FL within IoFT has focused on algorithm development, and thus far, there has been little in the context of developing statistical models. In this subsection, we discuss challenges and interesting open directions for FL from a statistical perspective. In particular, several statistical challenges such as heterogeneity, dependence, and sample bias under privacy and communication constraints need to be addressed. Part of the challenge is developing a suitable modeling framework that allows the assessment and validation of methods handling the challenges above.

Below, we discuss areas where statistics can make a significant contribution. This is by no means an exhaustive list as IoFT is still in its infancy phase, and new challenges will arise as IoFT infiltrates new applications.

\subsubsection{Dependence (beyond empirical risk minimization)}

Statistical dependence in IoFT is a common challenge since clients may be dependent (due to geographic or spatial dependence, common features, e.t.c.) or data within each client may be correlated. Current FL algorithms operate under an empirical risk minimization (ERM) framework and assume independence both within and across clients. In correlated settings, even classic SGD will lead to a biased estimator of the gradients \citep{chen2019closer} as the loss function cannot be simply summed over all data points. Adding to that, correlation needs to be learned with additional challenges posed by communication and privacy constraints in IoFT. 

Yet here lies a significant opportunity: if learned effectively, a dependence structure across clients can be exploited to improve prediction, update sampling schemes and better allocate resources. More specifically, statistical approaches for dealing with dependence typically involve learning a suitable dependence structure (e.g. graphical model) amongst the clients of interest and then exploiting the structure (e.g., \cite{Koller2009,Pearl2000, RasWil06}) for inference and prediction. Following this line of thinking, the challenges are to (i) develop an FL approach to learn a suitable dependence structure amongst the clients; (ii) exploit the learned dependence structure for improved inference and prediction within FL; (iii) develop an FL approach to deal with correlated data points within each client. While there are numerous techniques to address (i)-(iii) in the standard centralized setting (e.g., \cite{LiuLafWas09,park2015learning,Ravetal11}), their applications to IoFT are yet to be explored.



{\bf Network learning:}
Learning networks (especially through graphical modeling) is a problem that has received significant attention and research in the statistics and machine learning literature \citep{LiuLafWas09, Ravetal11, YuaLin07}. In IoFT, when there is dependence amongst the clients, learning a graphical model/network structure potentially improves overall performance. However, there remains the open challenge of adapting and implementing these graphical modeling algorithms that learn pairwise sufficient statistics to respect communication and differential privacy constraints. At the heart of the challenge, if our goal is to learn a network structure amongst nodes, second-order statistics are required in the computation. From a privacy and communication perspective, this requires communication between all pairs of nodes in order to compute these second-order sufficient statistics. One possible solution is to carry over ideas from differential privacy to network learning problems. Also, statistical ideas such as sketching or randomization \citep{DrinMah05, DrinMah09, DrinMuthuMahSarlos11,RaskuttiMahoney}) may improve privacy whilst still learning sufficient statistics.

{\bf Correlation:}
Handling correlated data within each client is also an essential challenge as it goes beyond ERM. Here stochastic processes, such as Lévy processes and auto-regressive models, can play an important role. One example is  multivariate Gaussian process (GPs) models, where a covariance matrix encodes dependence both within and across clients. Yet again, privacy and communication considerations in IoFT yield decentralized estimation of a covariance matrix, a challenging task. Here, one may assume that the covariance matrix/kernel is parameterized by a small set of parameters (e.g. Mattern, RBF). Very recent work has shown that despite biased gradients, SGD can still converge in correlated settings, specifically GPs \citep{chen2020stochastic}. A natural question to think about is whether a natural extension exists for FL.

{\bf Validating dependence models:}
The final challenge associated with network learning for IoFT is to address whether the learned network improves predictive power, and if so, which approaches are best. For example, would performing network learning improve predictive performance compared to clustering nodes and doing personalization? Since the ultimate goal is predictive power (although the network may contain helpful information), this presents a natural validation metric for network learning methods. Due to the modeling framework, predictive power can be validated theoretically using refined bounds that incorporate dependence, simulation studies and real data examples.

\subsubsection{Uncertainty Quantification \& Bayesian Methods}
As seen in Secs. \ref{sec:global}-\ref{sec:metalearning}, very few approaches are able to quantify uncertainty. Besides, $\texttt{Fed-ensemble}$ and $\texttt{Fed-BE}$, FL methods have mainly focused on point predictions. Yet a model should acknowledge the confidence in its prediction. Therefore, further exploration into Bayesian methods is important for the application of IoFT within different domains. 

One possible route is to place a prior over $\bm{w}$ (and possibly personalized weights $\beta_i$) $\mathbb{P}(\bm{w}, \bm{\beta}_i)$ and try to estimate the posterior $\mathbb{P}(\bm{w}, \bm{\beta}_i|D=\{D_1, \cdots, D_N\})$. Clearly, if $f_{\bm{w}}(x)$ is a complex function such as a neural network,  $\mathbb{P}(\bm{w}, \bm{\beta}_i|D)$ is usually intractable, yet one may hope to extract some samples from it. Here recent advances in  approximate posterior sampling such as Stochastic Gradient Langevin Dynamics (SGLD) \cite{welling2011bayesian} or Stein Variational Gradient Sescent (SVGD) \citep{liu2016stein} may be possible solution techniques. Besides the approximate sampling schemes above, hierarchical Bayes may be worth investigating, as IoFT has an inherent hierarchy between the orchestrator and clients. The challenge to be addressed here is how to estimate a hierarchical Bayes model in a decentralized fashion that preserves edge privacy and minimizes communication. 

As an alternative to a prior on model parameters, one can take a functional route by directly specifying a prior on the functional space $f_{\bm{w}}(x)$; commonly done using Gaussian processes (GP). Indeed, GPs have a long history of success in engineering applications \citep{wei2018reliability, plumlee2019computer, joseph2019space, gramacy2020surrogates} and their success in IoFT may pave the way for many new applications. However, the challenge is that GPs are based on correlations and do not conform to the ERM paradigm FL is currently based on. \textit{Interestingly, learning a prior on the functional space may also help in personalization and  meta-learning. A learned GP innately acts as a prior through which predictions are obtained by conditioning on new data}.      

\subsubsection{Statistical heterogeneity \& Personalization} In IoFT, statistical heterogeneity is a central challenge as individual devices may have different data patterns and potentially collect different amounts and types of data. As highlighted in Secs. \ref{sec:personlized} and \ref{sec:metalearning}, personalization (and meta-learning) are one way to overcome the heterogeneity challenge by allowing clients to retain their individualized models while still borrowing strength from each other. However, personalization poses many exciting challenges and open questions. Below we list a few.   

\begin{enumerate}[i.]
    \item It is essential to investigate when a personalized model is needed and understand the trade-off between a global model and device-specific features. Intuitively, when there is significant heterogeneity among clients, fitting personalized models would be better than using a global model. Also, as pointed in \cite{kontar2020minimizing}, when data is highly heterogeneous, negative transfer of knowledge may occur where each client can generate better models using their data in isolation compared to sharing knowledge with other clients. \cite{lg-fedavg} conducted some analysis on additive models that matched this intuition. Yet, literature on approaches that characterize the heterogeneity of data and decide on whether personalization is needed is largely missing.

    \item Following the idea above, one can also decide on how many personalized models to build. As described in Sec. \ref{sec:semi}, this entails inferring the number of clients clusters where data within each cluster is homogeneous. A key question remains: how to cluster clients in IoFT? Clients usually send back a set of weights or gradients, however, are these summary statistics sufficient to recover true client clusters? If not, what sufficient statistics can achieve such a goal, and do they guarantee privacy? Here one may pose a question: what client statistics are needed such that clustering (say using K-means) using all data ($D$) in a centralized regime and clustering using the client statistics in IoFT will yield similar outcomes. 

    \item Personalization may come at the price of privacy. \cite{gradientleakage} shows that input images can be reconstructed from unperturbed gradient signals, which opens the possibility of gradient attacks. Precautions like adding noise or quantization can somewhat reduce the risk. However, some experiments \citep{differentialprivacyreview} have shown that there exists trade-offs between privacy and performance. Therefore, it is essential to understand this trade-off better and propose methods that can improve both simultaneously.  Differential privacy may be a valuable tool along this line \citep{differentialprivacyreview}.

    \item A probabilistic approach to personalization is still needed. Ideas such as random effects have built the statistical foundation for incorporating unit-to-unit heterogeneity. They may be of great value if extended to the decentralized IoFT settings. 
\end{enumerate}


\subsubsection{Validation and hypothesis testing} To this point, there has been little work in FL on suitable statistical validation and hypothesis testing procedures. While there may be a general reluctance to impose statistical models since they are never truly correct and there are advantages to being model-agnostic; imposing statistical models provides several potential benefits. First and foremost, modeling provides a way to develop and assess a hypothesis testing approach. 

Most, if not all, prior work on FL has focused on deep learning algorithms due to their predictive power. If we are interested in questions associated with statistical estimation and inference, it makes sense to extend FL to incorporate models that are interpretable in addition to having good predictive capabilities. Algorithms that come to mind include kernel methods (see e.g.~\cite{Bach08a, KolYua10Journal,RasWaiYu12,RasWaiYu14}), Gaussian processes (see e.g.~\cite{Gor85, KontarRaskutti,RasWil06, yue2019renyi}) and other approaches. Like deep neural networks, all of these approaches exhibit non-linearities and function complexity while being more amenable to statistical inference.

One example of a statistical model was given in Sec. \ref{sec:personlized} where one can model the conditional distribution as: 
$$y_i \sim \mathbb{P}^i_{y|x} \big(f_{\bm{w}_i}(x_i)\big) \, , $$
Here, clients share the same $f$ (a linear model, kernel, Gaussian process, neural network) yet with different parameters $\bm{w}_i$ which allows for personalization. This represents $N$ semi-parametric models \citep{Bickelbic00} and a key question is what structure to impose on the $\bm{w}_i$'s which incorporate the degree of statistical heterogeneity and dependence? Assumptions such as graphical models, low-rank models, sparse models and many others may be incorporated (e.g., \citep{HastieTibshiraniWainwrightBook,BuhlmannVDGBook}). Graphical models naturally lend themselves to network learning while low-rank models naturally lend themselves to clustering of nodes.

\subsubsection{Other Open Questions}

{\bf Domain Adaptation:}
The joint distribution of data pairs for client $i$ is given as $\mathbb{P}^i_{x,y}=\mathbb{P}_x \mathbb{P}^i_{y|x}$. Current models assume that $x \sim \mathbb{P}_x$ across clients yet there is a change in the conditional distribution $\mathbb{P}^i_{y|x}$, i.e. input-output relationship across clients. What happens if there is a covariate shift, $\mathbb{P}^i_x$, across clients. Indeed, this is not uncommon in IoFT as different clients may observe a wide range of unseen input on other clients (e.g. unique defect types of failure modes). Here domain adaptation may play a key role where the input is first mapped to a shared feature space, and then inference is made on the feature embeddings. A simple example is: 

$$f_{\bm{\beta}_i, \bm{w}}(x)=g_{\bm{w}}\circ\Phi_{\bm{\beta}_i}(x) \, ,$$

where $\Phi$ is a personalized encoder that projects to the feature space and $g$ is a global decoder with shared parameters $\bm{w}$ across all clients. 

{\bf Decentralized \& Collaborative Design of Experiments (DOE):}
DOE is critical within many domains \citep{wu2011experiments,chen2017sequential,castillo2019bayesian,kang2011design,xiong2009optimizing}. In the realm of IoFT, a key question is how can DOE be achieved? For instance, for expensive experiments or computer models, DOE often uses a sequential strategy to find the next-to-sample design points that best help in estimating an unknown response surface \citep{hung2015analysis} or providing statistical inference. In IoFT, such an expensive computer experiment may reside on the central server or perhaps each client has its own computer model. To this end, how can decentralized DOE be achieved? How can sequential designs learn a global computer model, given that clients may be of different fidelities \citep{imani2019mfbo}? How can computer models borrow strength from each other for better calibration \citep{plumlee2017bayesian} while preserving privacy? How can we distribute the experimental design process across clients given resource limitations of each client? Such questions will be of key importance in IoFT and open a new area of exploration for DOE. These same questions extend to Bayesian optimization which also aims at optimal sequential sampling, yet with the goal of finding optima of an unknown response surface.   

{\bf Vertical IoFT:}
Current literature is mainly focused on horizontal IoFT where the edge denotes different clients. Yet more exploration should be oriented towards vertical IoFT,  where we have the same client  (ex: patient), yet information is stored across different system components (ex: hospitals). Here, feature extraction shall play a critical role, where features from different components are extracted and then jointly trained for a holistic model. Techniques such as sketching or random embeddings may be of use to preserve data privacy.

\subsection{\emph{Optimization Perspective}}
Several optimization algorithms have been proposed in recent years to learn a global or personalized model collaboratively within IoFT; see Secs.~\ref{sec:global}-\ref{sec:metalearning} for details. However, significant theoretical and computational hurdles associated with solving such problems remain unresolved. In this section, we discuss FL from an optimization perspective. More specifically, we categorize the main existing streams of work and provide insights on potential open directions. 

Several algorithms have been recently proposed to mitigate the issue of heterogeneity. A class of these algorithms add a local regularization term to the client's objectives. Popular examples of such algorithms include \texttt{FedProx} \cite{fedprox}, \texttt{FedDyn} \cite{acarfederated}, \texttt{DANE} \citep{dane} and its federated counterpart \texttt{FedDANE} \cite{li2019feddane}, \texttt{SCAFFOLD} \cite{karimireddy2020scaffold}, \texttt{FedPD} \cite{zhang2020fedpd}. While existing methods tackle heterogeneity by adding a regularization term, it is worth exploring algorithms that add adaptive ball constraints when minimizing local objectives. Such methods can control the variability of local parameters and can align the stationary solutions of local and global objectives when the radius of the ball constraints converges to zero in the limit.

The convergence guarantees and complexity rates of many of these algorithms were established under a variety of assumptions; see \cite{kairouz2019advances, karimireddy2020scaffold, zhang2020fedpd}. When the clients are identical, \texttt{FedAvg} coincides
with \texttt{FedSGD} \citep{parallelsgd,fedsgd} (also known as local SGD) and both converge asymptotically. In the (strongly)-convex case with \textit{i.i.d} data, \texttt{FedAvg} converges to a global solution with optimal complexity order. Despite its success in many applications, the former algorithm suffers when applied to FL settings with non-\textit{i.i.d} data. More specifically, in the presence of heterogeneity, the discrepancy between the average local client optimum and the global optimum results in a drift in local updates, often called client drift. This drift can significantly affect the convergence guarantees and complexity rate of the algorithm. In particular, \cite{zhang2020fedpd} provides a problem instance for which \texttt{FedAvg} with constant step-size diverges to infinity. \cite{karimireddy2020scaffold} show that the client-drift effect in unavoidable even if we use full batch gradients and all clients participate in each communication round. 
 
To mitigate the issue of heterogeneity, \texttt{FedProx} uses a proximal term that suppresses the variance among local client solutions. The method is shown to converge without any boundedness assumptions on the local gradients. Despite more stable convergence, the method is still based on inexact minimization since it does not align local and global stationary solutions. When all clients have the same optima, and full batch gradients are used, \texttt{FedAvg}  and \texttt{FedProx} have the same complexity rate. Other algorithms that directly tackle client-drift in FL appear in the work of \cite{karimireddy2020scaffold, zhang2020fedpd, acarfederated}. These algorithms are shown to converge without any bounded client gradient assumptions. Among them, \texttt{FedDyn} requires fewer communication rounds between the central orchestrator and clients compared to \texttt{SCAFFLOD}. 

Another FL method class uses an adaptive choice of step size for the client and server optimizer steps. This approach is motivated by the practical benefits that repeatedly appeared in adaptive optimization when training machine learning models. Examples of FL-versions of such algorithms are \texttt{FedAdam} and \texttt{FedYogi} \citep{fedadam}. We refer the readers to Sec. \ref{sec:eff} for a more breadth and depth discussion on the algorithms presented above.

\subsubsection{Open Directions}
Despite the focus on algorithm development and their corresponding optimization mechanisms, there are potential optimization questions  still to be explored for FL. 

{\bf Choice of the number of local steps $E$:}
One popular example of algorithms proposed for FL is \texttt{FedSGD}, a distributed version of SGD. While utilizing parallel computations yields efficient training for large data-sets, such methods incur high communication cost since they require passing the gradient vector of clients to the server at every iteration. To remedy the high communication cost, \texttt{FedAvg} was proposed. As detailed in previous sections, this method applies multiple local SGD steps in each communication round before updating the global parameter at the server. Despite being widely used in practice, several recent results have shown degrading performance in the presence of heterogeneity \cite{zhao2018federated}. One particular issue is that client heterogeneity can introduce a wide discrepancy between local and global objectives, resulting in a drift in local updates. While a higher number of local updates $E$ reduces the communication cost, it can magnify this client drift. Similarly, a low $E$ directly implies a high communication cost. Hence, the choice of $E$ presents a trade-off between convergence stability and communication cost. Some interesting open questions worth investigating are i) How should we choose the number of local steps $E$? ii) In the presence of heterogeneity, can we choose a different $E_i$ across clients? iii) Can we utilize an underlying client heterogeneity structure to decide on $E_i$?


{\bf Higher order algorithms:}
Almost all existing FL algorithms belong to the class of first-order methods. Designing second-order methods tailored for FL remains an exciting area to explore. This class of algorithms can potentially perform better than first-order algorithms when the objective is highly non-linear or ill-conditioned. Such methods aim at effectively using the curvature information for faster convergence. To overcome the computational drawback of computing the Hessian, estimating the curvature using first-order information is well-studied in quasi-newton methods, as well as their stochastic variants \citep{SQN}. Motivated by their potential superior performance in several applications and under special structural properties of the objective, a natural potential research direction is investigating FL variants of second-order methods.

{\bf Zeroth-order algorithms:}
These methods utilize a heuristic derivative-free approach for updating the sequence of optimization iterates. Such methods can be helpful in problems with access to only noisy evaluations of the objective function \citep{conn2009introduction, rios2013derivative}. Several recently arising machine learning applications \citep{ZO-RL, ZOO-ANN} have brought significant attention to zeroth-order algorithms. Studying FL variants of zeroth-order algorithms is a potential research direction that can pave the path for interesting FL applications.

{\bf Min-max optimization:}
Developing new methods for solving min-max optimization problems in FL settings is still premature and is worth investigating. Min-max optimization problems have recently appeared in a wide range of applications, including adversarial training of Neural Network, fair inference~\cite{Renyi, agnosticfl, li2019fair}, training GANS~\cite{goodfellow2014generative}, and many others. A wide-variety of algorithms have been proposed to solve these problems in non-FL settings. Most commonly, stochastic gradient descent-ascent (SGDA) that applies an ascent step followed by a descent step at every iteration is used in practice. Applying SGDA and its variations is undesirable in FL since it requires communication at each iteration. A natural research direction is to investigate FL variants of such methods. One potential direction that can be explored when the maximization problem is (strongly)-concave is using duality theory to minimize the model parameters and the dual variables jointly. 

{\bf Resource allocation \& constrained optimization: }
As mentioned in Sec. \ref{sec:iot}, we do not cover approaches for resource allocation in IoFT since literature in this area is scarce. However, in IoFT, clients themselves are heterogeneous in their capabilities in computation, memory, processing power, connectivity, amongst others. Modeling approaches should account for edge resources while at the same time ensuring uniform or comparable model performance across clients regardless of their capabilities. Resource allocation for optimal trade-offs between convergence rates, accuracy, energy consumption, latency, and communications are of high future relevance.

Along this line, client resources may be posed as constraints in the local and global objectives. However, constrained optimization is still to be investigated in IoFT. While deep learning models usually do not place constraints on the parameters, we envision that constrained optimization will arise within domain-specific applications. Similar and unique constraints across clients pose interesting questions on redefining local and global updates and understanding theoretical guarantees in such settings.    

{\bf Full decentralization: }
At the current stage, most developed methods for IoFT rely on a central orchestrator. Full decentralization is a step forward, where clients collaborate directly with each other without the guidance of an orchestrator (see Fig. \ref{fig:peer}). This may add an additional layer of privacy as it is difficult to observe the system's full state. With the increased penetration of blockchain applications, IoFT may become fully personalized. Yet, many of the FL techniques described in this paper will need to be re-thought to make this possible. Also, besides statistical and optimization challenges, there lies fundamental challenges of trust and privacy as malicious clients may corrupt the network and violate privacy without a central authority taking corrective action. Consequently, a level of trust in a central authority in a peer-to-peer network can benefit in regulating the network's protocols.

\section{Applications} \label{sec:applications}
In the previous sections, we have discussed defining features of IoFT and data-driven modeling approaches for decentralized inference. Yet, IoFT both shapes and is shaped by the application it encompasses. This boils down to a crucial question: how will IoFT shape different industries, and what domain-specific challenges it faces to become the standard practice? Through the lens of domain experts, we shed the light on the following sectors: \textbf{manufacturing}, \textbf{transportation}, \textbf{energy}, \textbf{quality \& reliability}, \textbf{computing}, \textbf{healthcare}, and \textbf{business}. To keep notation simple, this section slightly exploits earlier notation.

\subsection{\emph{Manufacturing}}  \label{sec:manufacturing}

The fourth industrial revolution (Industry 4.0), which is undergirded by smart technologies like IoT, has brought disruptive impacts on the manufacturing industry  \citep{behrendt2017achieve,leurent2018next}. In the United States alone, 86 percent of manufacturers believe that smart factories built on Industry 4.0 will be the primary driver of competition by 2025. Furthermore, 83 percent believe that smart factories will transform the way products are made \citep{wellener2019deloitte}. However, only five percent of US manufacturers surveyed in a recent study reported the full conversion of at least one factory to “smart” status, with another 30 percent reporting they are currently implementing initiatives related to smart factories \citep{laaper2020deloitte}. This means that nearly two out of three (65 percent) manufacturers surveyed report no progress on initiatives that they overwhelmingly point to as their primary driver of near-term competitiveness in five years \citep{laaper2020deloitte}.

Distrust is listed as one of the dominant factors inhibiting the spread of Industry 4.0 \citep{nagy2018role}. The current paradigm of IoT where data is agglomerated in a central server does not foster trust. Instead, it breeds concerns about privacy and security \citep{buckholtz2015cloud}. Also, several time-sensitive applications could be advanced by Industry 4.0 but are inhibited by the current IoT paradigm. For example, through cloud computing, manufacturing machines could benefit from advanced control algorithms that significantly improve their performance \citep{okwudire2018low}. However, with an IoT infrastructure reliant on the exchange of data with a centralized server, internet latency becomes a significant challenge \citep{okwudire2020three}. Moving large amounts of data to and from a central server also demands high internet bandwidth. Another major challenge of the current IoT landscape is that it is poised to benefit large enterprises at the expense of small and medium-sized enterprises. Given the concerns around privacy and security, companies are inclined to use private rather than public cloud infrastructures \citep{horn2016feasibility}. Therefore, smaller companies are unlikely to have the capital to set up and maintain their own private cloud infrastructure. Even if they can set one up, they are unlikely to generate sufficient data volumes for meaningful big data analytics.  

IoFT could help overcome the aforementioned challenges and create lots of new opportunities in present-day manufacturing. For example, it could enable vertical integration of IoT across a manufacturing ecosystem, which is key to capturing value from Industry 4.0 \citep{nagy2018role}. The ability for entities to keep their data private while collaborating on a shared model could allow original equipment manufacturers (OEMs), for instance, to integrate their data analytics with those of their suppliers to help improve quality across their entire supply chain. This benefits the OEMs as well as their suppliers. Similarly, developing a shared model without compromising privacy could help level the playing field between large and small enterprises. Small companies who cannot afford a private cloud infrastructure can benefit from public cloud infrastructures without sharing their data.
Moreover, even if they do not have large enough datasets for analytics, they can benefit from the data of other entities through a shared model. Furthermore, data analytics for time-sensitive applications can be run at the edge \citep{mubeen2017delay}, closer to the device or machine, to reduce latency while also benefiting from a shared cloud-based model across several machines \citep{madhyastha2020remotely}. The same benefit extends to bandwidth-intensive applications.

\begin{figure}[htb!]
\centering
\includegraphics[width=0.5\textwidth]{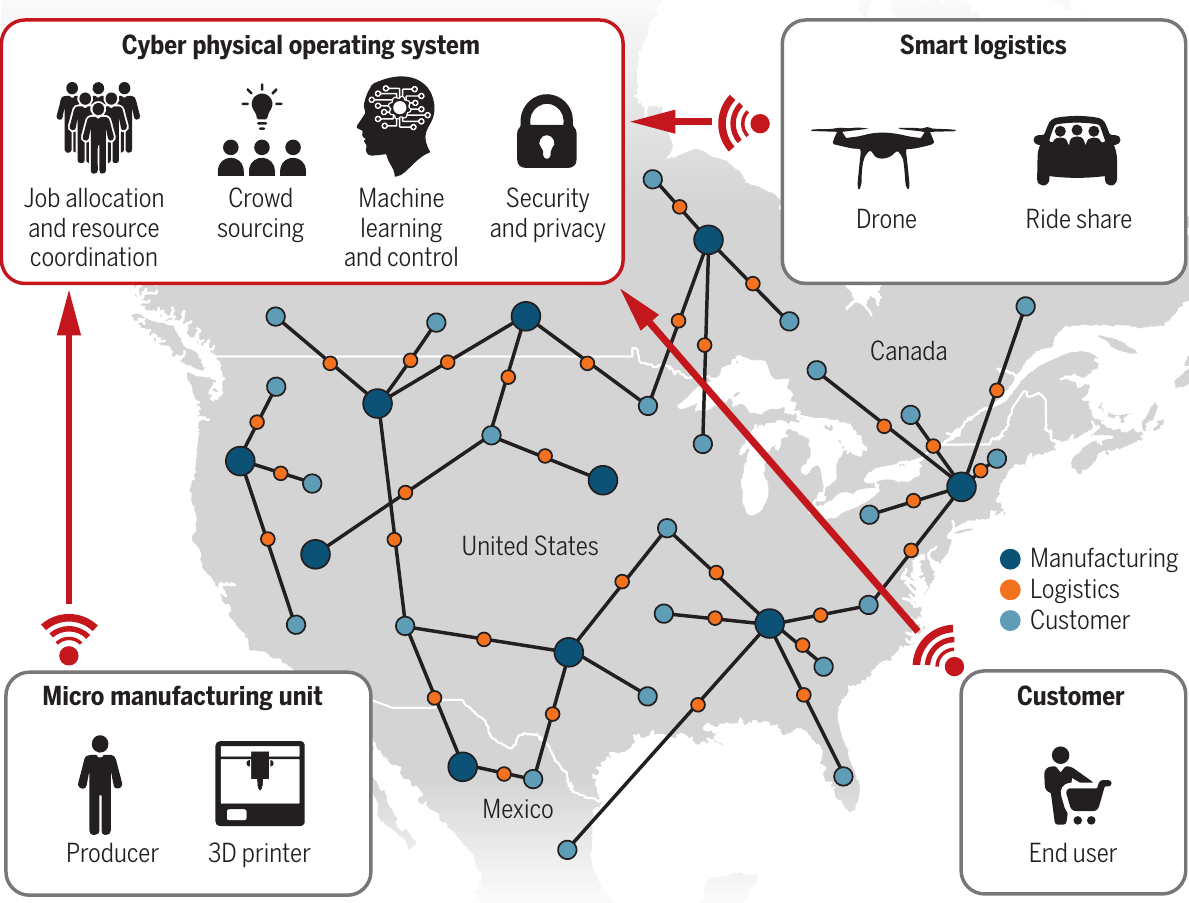}
\caption{Cyber-physical operating system connecting and coordinating customers, micro-manufacturing units and smart logistics to enable massively distributed manufacturing. Reprinted from C Okwudire and H Madhyastha, Science 372, 341 (2021); Graphic: C. Bickel.}
\label{fig::manufacturing}
\end{figure}

IoFT will also be a key enabler of futuristic paradigms like massively distributed manufacturing (MDM), briefly described in Sec. 1. MDM involves the manufacture of products by a large, diverse, and geographically dispersed but coordinated network of individuals and organizations with agility and flexibility, but at near-mass-production quality, productivity, and cost-effectiveness \citep{MDM}. A cyber-physical operating system (CPOS), which intelligently, efficiently, and securely coordinates large networks of cloud-connected, autonomous, and geographically-dispersed manufacturing resources will be needed to support MDM. The importance of operating systems to support present-day distributed manufacturing has been highlighted in recent works, together with ideas on how to realize them \citep{correa2018new,garcia2019sustainable}. However, in the context of MDM, some distinguishing features of CPOS are that: it will optimally allocate manufacturing jobs to the resources connected to it and leverage distributed and democratized delivery systems, like Uber/Lyft and drones, for logistics. It will apply machine learning to the data gathered from sensors to help assure and improve quality and optimize operations.
Furthermore, it will leverage the ingenuity of humans via crowdsourcing of ideas to improve manufacturing operations across networks of manufacturers. It will leverage cybersecurity measures to protect the intellectual property and privacy of participants. CPOS will thus allow the collaboration of large, autonomous, heterogeneous, and geographically dispersed networks of manufacturers to rapidly respond to production demands and disruptions with agility and flexibility while ensuring high quality, productivity, and cost-effectiveness \citep{MDM}.  

IoFT will allow CPOS to maintain the autonomy, privacy, and security of all the participants in MDM while enabling them to develop shared models that improve quality (and other performance metrics) across the entire system. MDM, enabled by CPOS and IoFT, promises to improve the responsiveness and resilience of manufacturing to urgent production demands (e.g., in emergencies like pandemics); promote mass customization and cost-effective low-volume production; gainfully employ lots of ordinary citizens in manufacturing (e.g., through the gig economy); and reduce the environmental footprint of manufacturing, by producing items closer to their points of use.  

In a nutshell, Industry 4.0 is poised to transform the manufacturing industry, but it would need a transition from traditional IoT to IoFT for its promise to fully materialize. IoFT will help alleviate issues around privacy, security, cost, data scarcity, communication latency, and bandwidths that are slowing down the adoption of IoT solutions in the manufacturing industry. It will also help catalyze new paradigms of manufacturing, for example, massively distributed manufacturing. To facilitate the transition of the manufacturing industry from traditional IoT to IoFT, the challenges discussed in Sec.\ref{sec:iot} have to be addressed in the context of manufacturing.

\subsection{\emph{Transportation}} \label{sec:transportation}

The prevalence of smart personal devices and the emergence of connected vehicle technology provide a plethora of opportunities for vehicles, travelers, and the transportation infrastructure to be in constant communication. This connectivity promises a safer and more sustainable transportation system with enhanced levels of mobility and accessibility. Connectivity allows subsystems that were previously modeled and optimized separately to be modeled as a single system, thereby capturing their interactions. This comprehensive modeling approach allows for increasing system throughput, which results in many benefits for travelers (e.g., lower prices, less congestion, more reliable travel times, lower levels of greenhouse gas emissions) and the transportation system (less pressure on the infrastructure). Take the example of traffic signal control systems. Traffic signal controllers are traditionally optimized locally, either per intersection or set intersections within an arterial. This optimization is based on local information: as vehicles approach an intersection, they activate loop detectors deployed in the pavement, sending a signal to road-side controllers. The controller then optimizes the traffic signal to minimize total delay. In an arterial setting, the optimization of the controllers at downstream intersections could be further informed by the state of the upstream intersections. Connected vehicle (CV) technology provides two unique opportunities for traffic signal control: (1) controllers can be optimized proactively before vehicles arrive at intersections using the messages received from connected vehicles; and (2) arterial-level optimization can be advanced to network-level optimization by customizing basic safety messages (BSMs) transmitted by vehicles to include route-choice information or estimating this information based on standard BSMs (\cite{Intersection_1, Intersection_2, Intersection_3}). 

 \begin{figure}[!htb]
	\centering
	\includegraphics[width=0.5\textwidth]{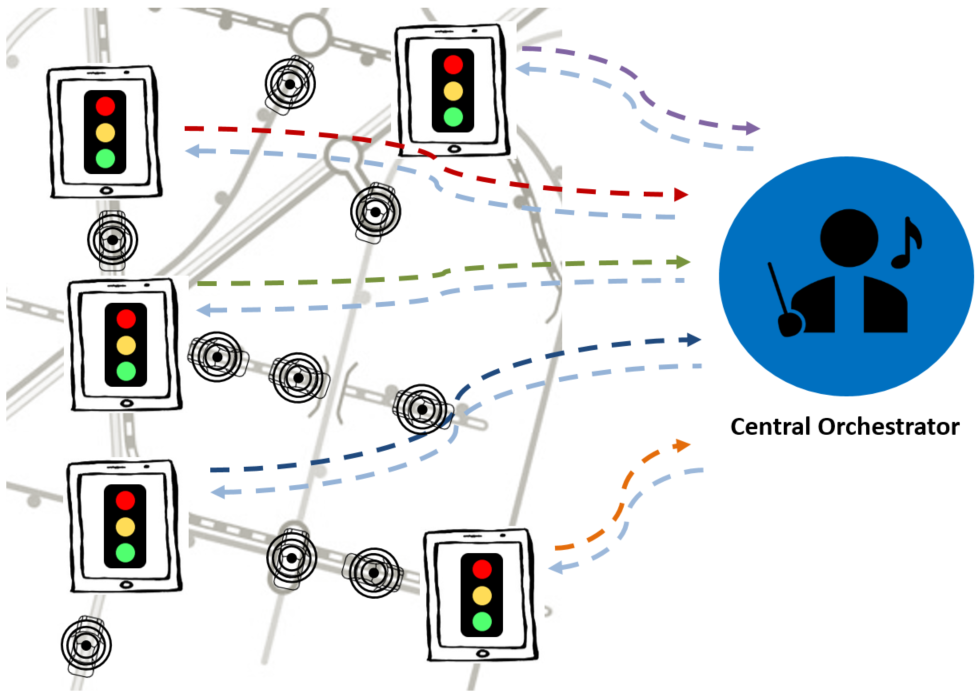}
	\caption{\label{fig:transport} Network-level intersection control}
\end{figure}

Despite the benefits that connectivity can offer, the existing methodologies are generally not scalable to allow for fast decision-making in connected systems. This lack of scalability creates a critical bottleneck in leveraging connectivity in transportation applications, especially since the state of the environment in transportation systems changes dynamically. Yet here lies a critical opportunity for transportation systems: with more compute power on edge devices (ex: AI chips in autonomous cars), we may can exploit these resources to decentralize model training.

Take the example of a shared mobility system, such as ride-sourcing, ridesharing, or micro-mobility service. Although the principle of sharing resources has been used in transportation systems for several decades (e.g., transit or carpooling), the advent of smart and connected personal devices led to the unprecedented growth in shared mobility systems. Consider the well-known ride-sourcing company, Uber.  
From an operational point of view, Uber can be considered a fleet operator. However, traditional optimization-based fleet dispatching schemes are not scalable for Uber as it scales up its operation to entire cities, states, and countries. Uber can fulfill ride requests in densely-populated urban regions using myopic solution methodologies (e.g., dispatching the closest vehicle to a request's pick-up location) with short wait times, providing a high level of service. However, as the demand level diminishes in suburban areas, using myopic matching schemes leads to degradation of level of service, prompting Uber not to offer its services when demand follows below a threshold. 
The need for using a decentralized solution methodology for solving centralized matching problems in shared mobility systems has been acknowledged in the literature. The proposed solutions typically falls under one of the two categories of decomposition methods (\cite{decomposition_1, decomposition_2, decomposition_3, decomposition_4}) or partitioning and clustering approaches (\cite{partitioning_1, partitioning_2}). Both families of solutions attempt to solve a large-scale optimization problem by means of solving smaller sub-problems, typically by adopting an iterative procedure that allows for asymptotically approaching the optimal solution.

Despite successfully striking a balance between solution quality and time, there are still practical challenges that limit the applicability of these methods. These challenges include the lack of a guarantee in finding a feasible solution within a specified period and the high set-up cost of the problem in a dynamic environment that is fast evolving. These challenges can be addressed through IoFT by exploiting edge resources to achieve massive model parrallelization where the computational burden is divided between local devices. Besides that, IoFT reduces communication and storage needs and can continuously update the model in real-time.


Besides that, due to the high computational complexity of optimization-based approaches, there has been a recent surge in interest to leverage deep learning in transportation applications (e.g., \cite{Transportation_deep, Ridesharing_deep}). The benefit of deep learning models is that once trained, their evaluation typically takes a fraction of a second, rendering them effective for real-time applications. However, training high-performing models requires immense amounts of data. In turn, FL can provide an elegant solution through efficiently training a global model by incorporating focused updates from several local (possibly heterogeneous) datasets, thereby enabling training generalizable deep learning models. 
Also, recent advances in FL can account for data heterogeneity through personalization, where each client retains a fine-tuned model based on its local data. These advances would facilitate using high-performing deep learning models to make operational decisions in dynamic systems. For example, adopting FL could allow Uber to learn a global matching policy customized for regional operations with minimal additional training to capture local idiosyncrasies. Such regional models are likely to outperform myopic algorithms that use only spatiotemporally local data.

The application of IoFT in transportation systems is not limited to connected intersections and shared mobility systems. Other existing applications that rely on spatiotemporal gathering of peers, such as vehicle platooning (\cite{larson2014distributed}) and P2P wireless power transfer (\cite{abdolmaleki2019vehicle}), can be enabled by IoFT. To effectively operate such systems, fast decision-making is necessary. FL can bridge the inherent trade-off between solution accuracy and computational complexity of finding a solution in such applications. Additionally, it is anticipated that the CAV technology will give rise to new applications that leverage connectivity and, therefore, require fast decision-making. 

In addition to improving system throughput, IoFT can be used in transportation systems for privacy preservation purposes. In the age of autonomous vehicles, training models that can predict the motion of different traffic agents, e.g., vehicles, pedestrians, cyclists, etc., is of utmost importance. Typically, roadside sensors, such as cameras, can be used to obtain historical trajectories based on which trajectory prediction models can be trained. However, transmitting camera recordings or other identifiable data to a central server may create privacy concerns. IoFT can address these concerns as it allows for processing the data in the edge device, and only sending focused updates (such as gradients) needed for updating the global model to the server. Similarly, using FL, other models that rely on sensitive traveler data, such as mode choice, destination choice, and route choice models, can be effectively trained.

\subsection{\emph{Energy }} \label{sec:energy}

\begin{figure}[htbp]
\centering
\includegraphics[width=0.5\textwidth]{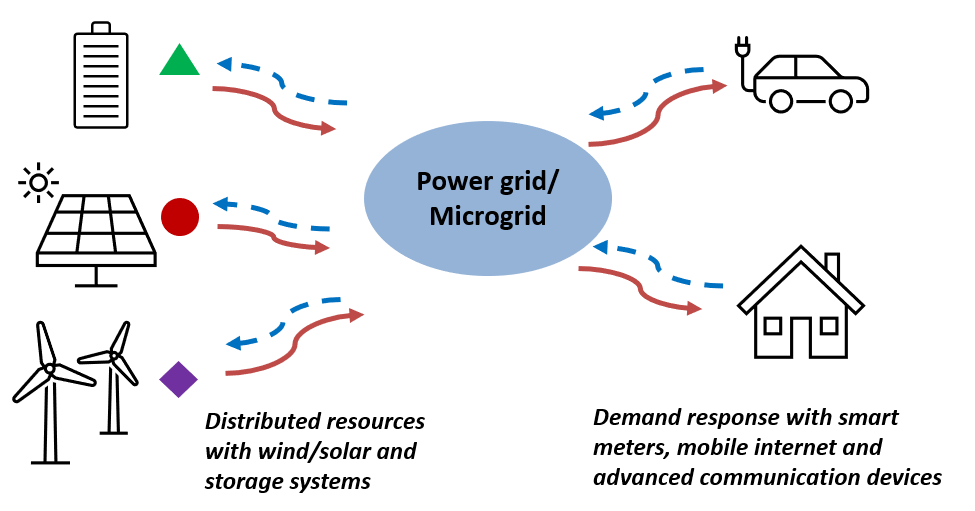}
\caption{Recent developments in energy infrastructure and technology}
\label{fig::energy}
\end{figure}

Modern society increasingly depends on complicated electric power systems. The US end-use of electricity reached 3.99 trillion kilowatt-hours (kWh) in 2019, and the total demand is expected to increase in the next decades \citep{consumption,energyfacts,mai2012renewable}.  Rapid developments in energy infrastructure and technology provide numerous opportunities for implementing new energy applications and services to meet demand, as depicted in Fig. \ref{fig::energy}. In particular, the market share of variable renewable energy sources, such as wind and solar, which provide local and distributed energy, grew to 19\% in 2019 \citep{energyfacts,mai2012renewable}. It is expected that the electricity generation from renewables will double over the next 30 years \citep{todayenergy}. Advanced communication capabilities, smart meter installations, mobile internet, and other smart technologies are enabling grid-responsive demand response and management services, such as shedding, shifting, and modulating load in peak and off-peak periods while minimizing occupant discomforts  \citep{neukomm2019grid,demandresponse,jang2020long,PALLONETTO2020,LI2020}. Additionally, increased use of battery storage technology and the growing penetration of electric vehicles will also change electricity supply and usage patterns \citep{LIU2019, wang2021look}.

Facing the massive transformation, IoT can provide system-wide, integrated approaches to managing modern power systems. The IoT infrastructure’s ability to capture and analyze data-intensive systems like the energy system can play a key role in managing renewables, demand response programs, electric vehicles, and other elements. Data collection and the use of intelligent algorithms can monitor and control the energy supply chain, including production, delivery, and consumption, so that suitable, cost-effective decisions can be made to balance supply and demand with minimal disruption to system operations. From the perspective of energy supply, since energy generation is an asset-intensive industry, data analytics can improve the efficiency of power production \citep{hossein2020internet,zhou2016energy}. On the demand side, buildings equipped with smart monitoring and communication devices can analyze end users’ energy consumption, identify their needs, and transform consumers into prosumers, adjusting their demand in response to system conditions \citep{hossein2020internet,zhou2016energy,NAN2018}. 

While IoT can empower data-intensive analytics by providing an integrated platform to collectively gather and process data, applying data science methods to analyze complex energy systems in the centralized IoT platform is not practical due to many untraceable complexities, making the IoFT paradigm more suitable.

First, massive data generated from sensors, actuators, and other devices in energy systems require real-time data analysis in high-dimensional regimes.  For instance, condition monitoring sensors, such as the vibration sensors in wind turbine gearboxes, produce frequent high-dimensional observations, and smart meters in residential, commercial, and industrial buildings produce massive amounts of end-user data in high frequency. In the standard cloud-based IoT framework, where centralized cloud/data centers collect and process data, the energy consumption for big data processing is substantial, thus possibly negating the benefits of IoT for the energy industry. 

Second, transmitting all the energy data to the central cloud can cause communication latency. Considering that the electric power end-use demand needs to be satisfied in real-time, such latency poses a severe risk to energy system operations. In power grid operations, the ancillary service allows the grid operator to maintain a balance between supply and demand at all times~\citep{Zou2016}. The ancillary service ranges in duration, ramping requirements, and magnitude \citep{ryan2014variable,zhang2018enabling}. These ancillary services become more important as renewable energy sources rapidly replace fossil fuel generation. Wind and solar, the fastest-growing renewables, are characterized by significant variability, often with limited predictability \citep{byon2015adaptive,alshelahi2021integrative}. Smart and grid-interactive buildings can provide such grid flexibility by adjusting their end-use patterns~\citep{BIEGEL2014,MCPHERSON2020}. Storages are also an attractive option for providing ancillary services. Communication latency causes ineffective coordination of these ancillary service resources and negatively affects grid reliability. As such, fast local updates for enhanced electricity supply and demand predictions are essential for successful ancillary service implementation.

Lastly, the modern power system is characterized by distributed energy due to the growing penetration of renewables, storage, and demand response. In these distributed systems, individual stakeholders such as utilities and consumers can perform their own decision-making \citep{Carli2019,Hosseini2021,Scarabaggio2021}. For example, utilities that manage their own renewable facilities may not want to share information with others to maximize their profit. Those who participate in the demand response programs may wish to adjust their end-use demand upon grid request without revealing their energy use patterns to others due to privacy issues. The decentralized IoFT framework provides the right platform for such decentralized and distributed decision-making.


While IoFT can remedy the limitations of the centralized IoT by building individual models locally at each end node, there are several challenging issues. First, end nodes often have limited computing power to train data science/machine learning models. As discussed above, sensors, actuators, and smart meters produce massive data. Inefficient computation at end nodes can delay predictive decision-making, fault diagnosis/condition monitoring, and change point detection, among others \citep{byon2015adaptive,choe2016change,yampikulsakul2014condition}. New data science methods are needed to optimally guide the model learning process for achieving computational efficiency with theoretical and practical implications in the IoFT paradigm.

Next, unlike traditional power supply with fuel-based generators and end-users who passively consume energy, modern power systems consist of highly heterogeneous units with distinct supply/demand characteristics.   On the demand side, technologies such as smart devices, demand management programs, and electric vehicles affect end-use patterns 24/7. On the supply side, energy units become more diverse and heterogeneous. Unlike traditional fuel-based sources, each renewable facility has distinctive power generation characteristics (e.g., facility layout, turbine type, each wind farm \citep{you2017wind}). While the personalized learning discussed in Sec. \ref{sec:personlized} can address the heterogeneity to some extent,  managing a large number of heterogeneous supply/demand units with distinct energy characteristics is challenging. Hence, personalized prediction needs to be translated into effective collaborative management.

Finally, energy consumption is significantly affected by ambient environmental and other localized conditions~\citep{Ning2017}. Peak demand predictions vary 1.5–2\% for every 0.5°C difference in predicted temperature \citep{bhandari2012evaluation,ocean}. Electricity needs vary due to many spatially localized characteristics, such as densely populated areas experiencing urban heat island effects in summer that increase electricity demand compared to suburban and rural areas \citep{jahn2019projecting,malings2017surface}, and EV charging stations exhibiting different charging patterns depending on localized characteristics. Renewable generations are also directly influenced by local weather and geographical conditions \citep{jang2020probabilistic}. Expanding the use of IoFT for environmental modeling will require modeling the spatially and temporarily correlated environmental conditions while incorporating local heterogeneous characteristics.  

In summary, modern power system faces technical challenges including computational scalability, efficiency, heterogeneity, localized characteristics, and distributed management. IoFT has the potential to address these challenges, and the successful development and implementation of IoFT will make the energy system (and its end users) ``smarter'' in terms of efficiency, flexibility, and economic competitiveness.

\subsection{\emph{Healthcare}} \label{sec:healthcare}

Healthcare stands to benefit significantly from IoFT because several unique contextual factors suggest that the status quo has failed when it comes to deploying machine learning (ML): (i) many existing models fail to generalize; (ii) legal and ethical implications limit the appetite to share data; (iii) the vendors who administer the electronic health records (EHRs) that contain patients’ data have an outsized influence on model deployment; (iv) national network-based research efforts started to adopt decentralized methods. This section will describe how these factors impact healthcare differently from other sectors and illustrate areas where IoFT is likely to thrive in this domain.

ML models are commonly used in early warning systems, diagnostic systems for radiology and pathology, and interpreting medical device output, such as in electrocardiography. While the medical literature contains numerous examples of apparently high-performing models, many of these studies suffer from poor generalization, which can either be demonstrated through an independent examination of its methods (by applying the PROBAST tool \citep{wolff2019probast}) or through external validation of the findings. This was particularly evident early in the COVID-19 pandemic, where nearly all studied models in an extensive systematic review were considered poorly generalizable. Although the pandemic has affected millions of people in the U.S. alone, most health systems did not have a sufficient number of patients, or adequate diversity, to ensure model generalizability. The lack of generalizability is not merely theoretical; it has also been systematically demonstrated. In a recent study examining over a thousand cardiovascular clinical prediction models, 81\% of validation studies found worse performance than was reported originally \citep{wessler2021external}.

Generalizability improves when models are developed using pooled data from multiple health systems. Under the Health Insurance Portability and Accountability Act (HIPAA) Privacy Rule, healthcare data may be shared for the purposes of research if identifiers have been removed, or under certain circumstances, if patients have authorized the use of their data for research after approval by an institutional review board \citep{ocr}. In most instances, data sharing between health centers also require legal agreements known as ``data use agreements'' or ``business associates agreements''. Even with such agreements in place, sharing of data between health systems may go against the expectations of the general public \citep{platt2018public}. Thus, pooling data from multiple health systems while enabling better ML models may potentially damage public trust. Indeed, when a 2019 partnership between Ascension and Google was reported on by the Wall Street Journal, it resulted in public outcry \citep{wallst}.

The difficulty faced by health centers in combining data with other health systems has led to a vacuum that EHR vendors have largely filled. Indeed, two of the most widely used healthcare ML models in the U.S. include the Epic Deterioration Index (owned by Epic Systems in Verona, Wisconsin) and the APACHE-IV scores (owned by Cerner Corporation, Kansas City, Missouri) \citep{singh2020evaluating,zimmeman2006acute}. Both models were developed using data from the EHR systems of multiple hospitals. Because EHR vendors have direct access to patient data (on behalf of their clients), they are well-positioned not only to combine data for analysis (with permission) but also to deploy the resulting models within their EHRs.

\begin{figure}[!htb]
	\centering
	\includegraphics[width=0.4\textwidth]{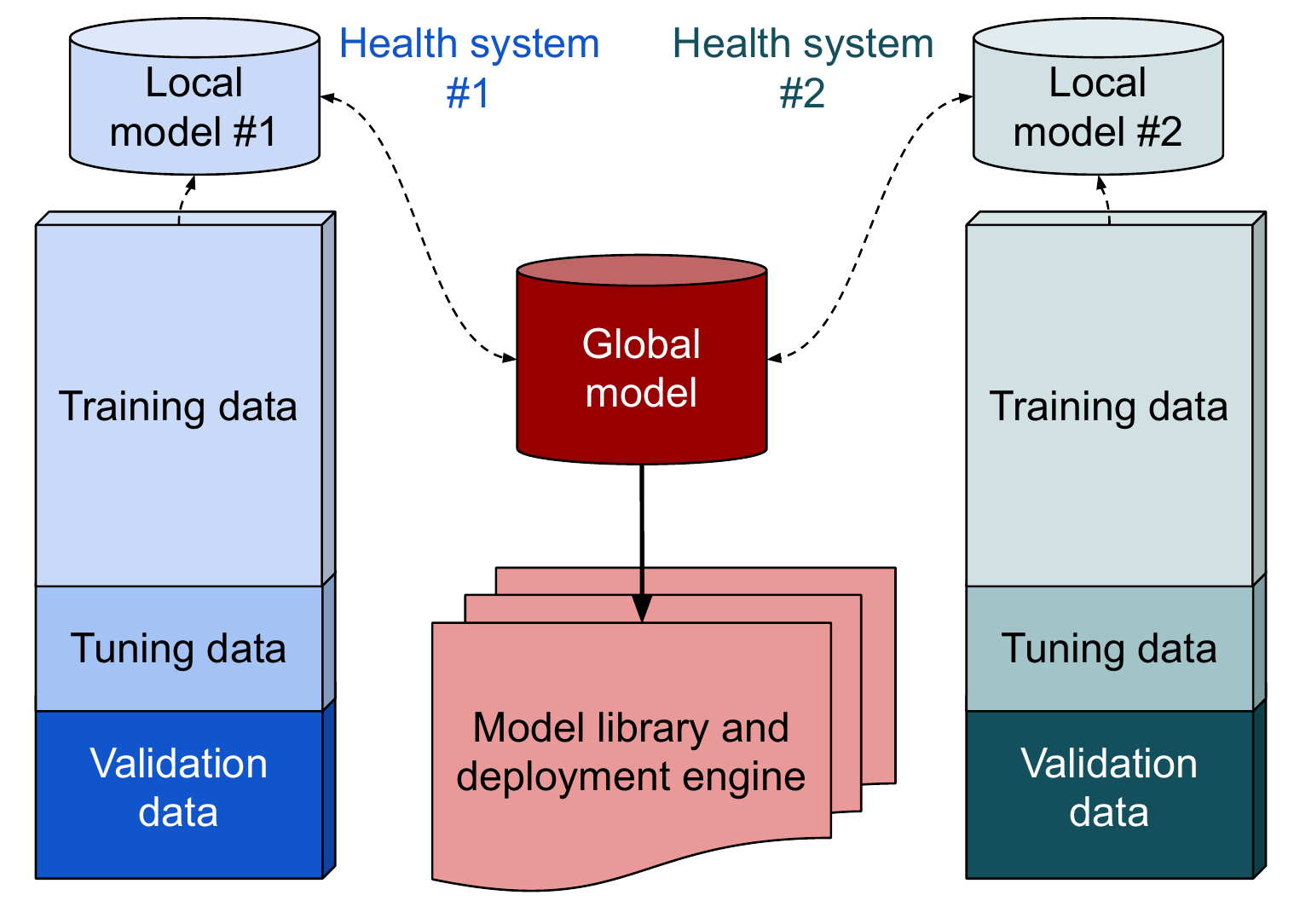}
	\caption{\label{fig:healthcare} IoFT in Healthcare }
\end{figure}

This siloing of data, while in the best interests of patients, has led to significant challenges in the development of high-quality, non-proprietary, freely available models. The healthcare informatics community has responded to this challenge, but there is still a long road ahead. In 2009, the development of the Shared Health Research Information Network (SHRINE) enabled federated querying of clinical data repositories \citep{weber2009shared}. Conceptually, federated querying of multi-hospital data allows researchers to identify optimal sites for ML model development based on rapid multi-system sample size determinations \citep{weber2015federated}. A federated querying system, known as the 4CE Consortium, was rapidly deployed to support COVID-19 research \citep{brat2020international}. Along this line, IoFT has been applied to the development of multi-hospital models in healthcare \citep{sarma2021federated,sheller2020federated,keane2021ai}. In these setting FL was used to allow health systems to share access to a model library and deployment engine without directly sharing data, as depicted in Fig. \ref{fig:healthcare}.

IoFT will be important in healthcare because it enables the creation of high-quality ML models without privacy risks. However, IoFT will have to contend with Food and Drug Administration (FDA) regulations that treat ML algorithms as a type of software-as-medical-device (SaMD). Initial applications of IoFT in health will likely be limited to class I and II medical devices, including ML models used primarily to inform care decisions. Class III-IV medical devices (e.g., cardiac pacemakers) require extensive review and premarket approval by the FDA. Class I-II devices only require premarket notification demonstrating equivalence with a legally marketed device. As a result, IoFT will likely be applied in supporting early warning systems in hospitals, automating order entry, and smart scheduling of patient visits. In each of these scenarios, global patterns exist but differ locally to the extent that the combination of global and local models will likely achieve superior results without sacrificing patient privacy.

\subsection{\emph{Business}} \label{sec:business}
Capturing and maintaining relationships between businesses raise many challenges for IoFT, and new methods to meet them are needed. To get an insight into these challenges, consider a business (the principal) that has the following decision to make: shall it build a facility to supply a particular item, or shall it `outsource’ it to another business (the agent)? In the former option, all the risks are carried by the principal, while in the latter these are shared with the agent. In the economy of today dominated by fast technological developments, outsourcing is often preferred. Despite the advantage of risk-sharing, outsourcing comes at a cost. The agent has more information about its operations and can use this information in its favor and at the expense of the principal, which is generally termed as ‘moral hazard’ in the economics literature \citep{laffont2009theory, spear1987repeated}. 

The example above highlights several key aspects of this relationship: 

\noindent \textbf{First:} Businesses are often intrinsically federated and are reluctant to share their proprietary business secrets. This federation forces decentralization of decisions. Thus, all the advantages of IoFT like localized computing, data privatization, security, and information privacy can be realized, along with the benefits of risk-sharing.

\noindent \textbf{Second:} A key challenge is that in business applications, agents may have different and often competing objectives. This becomes a new challenge for FL. Indeed, the models in Secs. \ref{sec:global} - \ref{sec:metalearning} are focused on maximizing a common objective. But if such a formulation does exist and somehow includes agents’ conflicting objectives, its successful implementation is contingent on true reporting by the agents. Otherwise, the principal has to monitor the agents’ operations continuously. Depending on the situation, monitoring may be impossible or too expensive and thus may nullify the advantage of decentralization of operations. 

\noindent \textbf{Third:} The principal can make arrangements with many independent agents, and each one of these may have an arrangement with other independent businesses supplying its services. This creates an organizational structure of relationships, forming a hierarchical structure of agents (i.e., a rooted tree with agents at various distances (levels) from the root). Here agents at intermediate levels (excluding the root agent and the agents at the end) play two roles: a principal to its subordinate nodes and an agent to the lower level node. This adds an additional level of complication for the use of FL in such business setups.

Though businesses are usually organized in a federated nature and are ideally suited for FL, the full potential of IoFT can only be realized if the above-stated problems are effectively solved. There are some recent encouraging developments towards this end. Below we highlight one possible solution. 

\begin{wrapfigure}{R}{0.29\textwidth} \label{fig:buss2}
    \centering
	\includegraphics[width=0.29\textwidth]{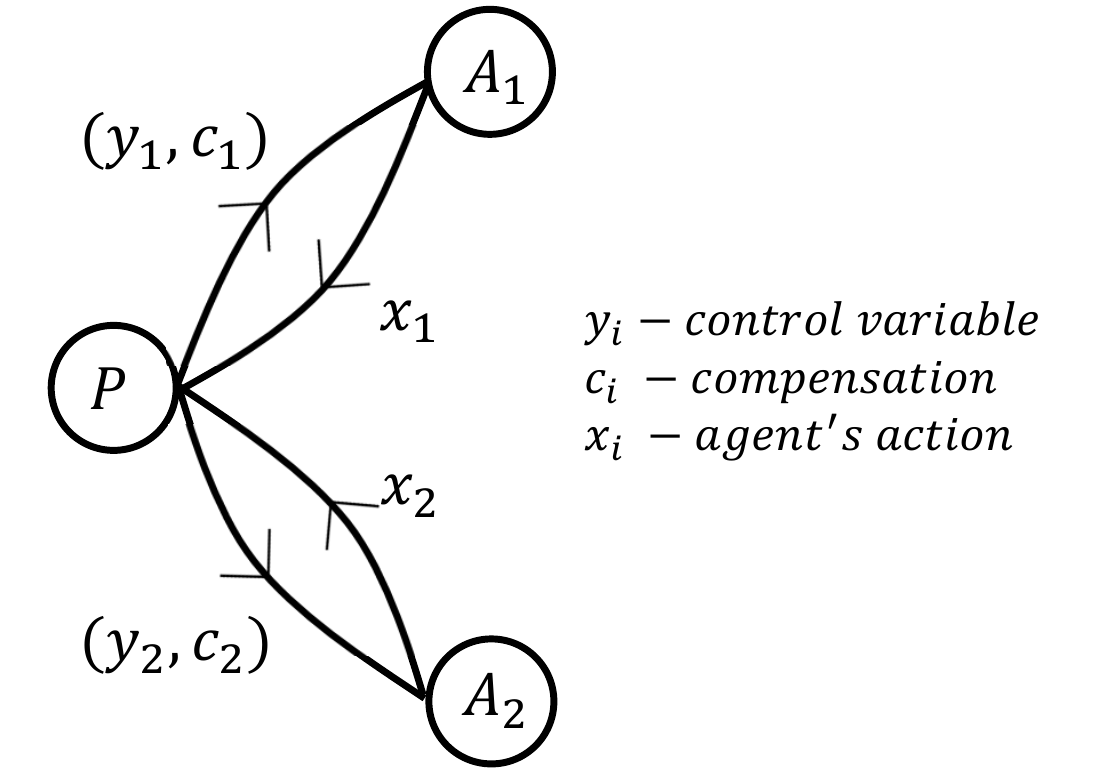}
    \caption{Example of decentralized decisions}
\end{wrapfigure}

In the single-agent case, an answer to this problem has been proposed by \cite{sannikov2008continuous} by the design of a mechanism that mitigates moral hazard and effectively decentralizes decision making. In a continuous dynamic setup, at the epoch $t$ the principal observes a noisy signal, $x(t)$ about the `effort’ of the agent and compensates $c(t)$ to the agent based on this signal.  It is shown that in case the signal noise is generated by a Brownian motion (a Gaussian process), there exists a `control’ variable $y(t)$ that can affect the noise of the signal to a level at which the principal can make a good decision about the compensation for the agent.  In a more realistic setting, the principal may create such relationships with several agents, each with private information, data, and objectives. There may also be interactions between agents' decisions, i.e., decisions by an agent may affect the outcomes of other agents' actions, and they may have conflicting objectives with each other and with the principal’s, making the moral hazard problem harder to mitigate. This case has been studied by \cite{qisaigal2021multiagent} who integrated the notion of Nash equilibrium into this model. Thus, if all the agents’ decisions form a Nash equilibrium, no agent can gain by falsifying information when all the other agents do not. This mitigates the moral hazard, and the decisions arrived at can be implemented. Fig. \ref{fig:buss2} shows a representation of decentralized decision-making with two agents. 

A brief overview of the solution methodology for a dynamic optimization problem in continuous time over a finite or an infinite horizon is as follows: the optimization problem faced by the principal is to find a policy which maximizes its expected discounted profit over the horizon, i.e., for the infinite horizon case:\[
\arg\min_{\{y(s), \, c(s) \, : \, s\geq 0\}} \mathbb{E} \int_0^{\infty}e^{-r_P\times s} f_P\Big(x(s),y(s),c(s)|\alpha (s)\Big) ds
\]

where $r_P$, $f_P$ and $\alpha_P$ are respectively the discount factor, profit function and data of the principal;  under the individual rational constraint  that each agent’s expected discounted profit over the horizon exceeds some predetermined minimum amount. When Brownian motions drive all randomness in the formulation, it can be seen that when the optimal policy is followed, the expected discounted profits of the principal and the agents are martingales. Using the principle of Bellman, this formulation is decomposed into federated optimization problems that are independently solved by each agent, while the principal solves a constrained Hamilton-Jacobi-Bellman (HJB) optimal control problem. The HJB solves for the continuation value (`value’ function of dynamic programming) of the principal as a function of the state variables, which include the continuation values of each agent. It is obtained by using Ito’s formula on the function and the fact that the expected profit of the principal is a Martingale when an optimal policy is adapted.  In the case of a hierarchical system, agents at the intermediate levels independently solve both an optimization and a constrained HJB problem, thus achieving a key goal of IoFT. 

In conclusion, the above-described decomposition mechanism between the principal and possibly several agents facilitates the use of federated learning. Each participant can effectively use the data collected from its operations and determine the profits, given the compensation it receives from the principal. The method described above is but one possible solution. The use of data-driven reinforcement learning is another viable option. Through this example, we hope to encourage researchers to explore decentralized decision-making within IoFT further.       

\subsection{\emph{Quality Engineering }} \label{sec:quality}
IoT as an enabling technology for real-time data sharing has stimulated a new paradigm in quality engineering, which expands quality control from the design and manufacturing stages to the whole product life cycle. For example, GE Prognostic Health Management Plus (PHM+) system uses its onboard sensors to collect engines’ operating data during flight. These data are communicated through its secure network and analyzed by the central server to provide proactive maintenance services. Similarly, the automotive industry uses vehicles’ onboard sensors to monitor vehicles’ real-time driving performance, allowing for early warnings of potential problems. Additionally, they can deploy integrated vehicle-based safety systems (IVBSS) \citep{feng2018estimation} to improve customers’ driving experience and safety. Throughout the product life cycle, quality assurance is highly demanded for customer satisfaction, which is especially imperative for expensive or safety-sensitive products. Further, these in-field operational data can provide quick feedback for continuous quality improvement. 

\begin{figure}[!htb]
\centering
\includegraphics[width=0.5\textwidth]{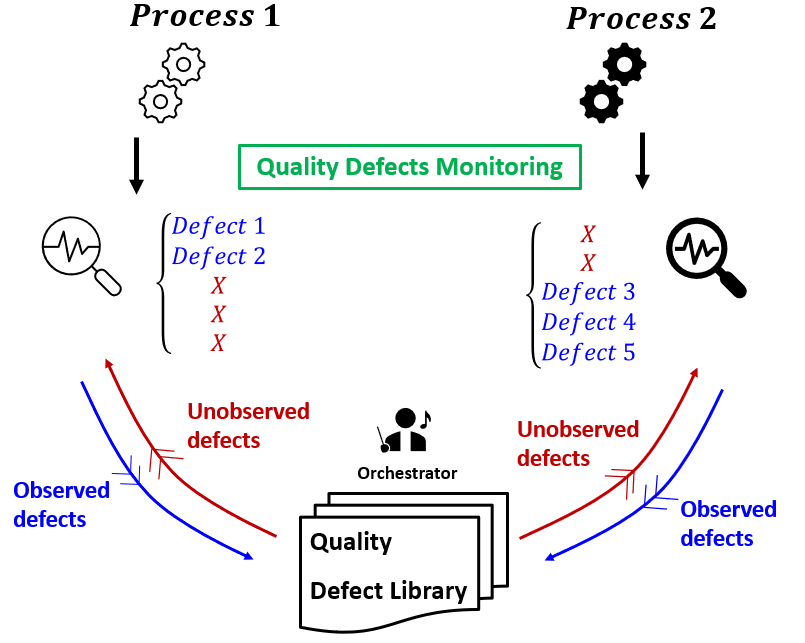}
\caption{\label{fig:quality} Quality control example in IoFT with missing anomalies}
\end{figure}

Online monitoring and fault diagnosis, which uses onboard sensors on products or in-situ process sensing signals during manufacturing, is one of the most critical research issues in quality engineering. Currently, most collected process sensing signals or quality inspection are multistream waveform signals; an image or video signals with very high frequencies \citep{woodall2014overview,qiu2020big,wang2021distribution}. Those require massive bandwidth and time to transmit the original data between local devices and the central orchestrator. More importantly, quality control requires fast decisions and real-time detection of anomalies. Therefore, decision-making ought to be on the edge and not on a central system. The shift towards IoFT, will allow tackling both of these challenges. In IoFT, compute resources at the edge are used so that only summary statistics and low dimensional information is transmitted to the central server (or perhaps a peer in a peer-to-peer network). Also, models reside on the edge and can be deployed immediately. Therefore, the IoFT platform shows distinctive advantages in quality control for reducing the communication load and making real-time decisions.

That being said, there are some unique research issues to advance quality control methodologies under the IoFT platform. Below we iterate a few:

\noindent \textbf{Insufficient data:} Quality control models, such as anomaly detection or fault diagnosis models \citep{liu2013adaptive,liu2013diagnosing}, are poised to greatly benefit from IoFT. In statistical process control (SPC), many edge devices or clients may lack sufficient data to build a normal operating baseline for abnormal change detection. For instance, (i) clients may not have observed the full set of possible anomalies as shown in Fig. \ref{fig:quality} (ii) new products/processes possess few data, so does small-scale clients (such as the 3D printing citizens in the Sec. \ref{sec:preamble}) and low volume manufacturing of rare and expensive products (ex: planes). IoFT as an emerging technology offers a medium to borrow strength across different clients for better SPC models while preserving copyrights and privacy. For instance, through meta-learning within IoFT, clients may directly adapt to new products/processes. Also, through domain adaptation, clients can learn across defect types observed only on some clients.  

\noindent \textbf{Continual learning:} IoFT allows knowledge to be readily shared. As a result, quality control models (e.g., anomaly detection) may be continuously updated to register new defects or improve detection and diagnosis accuracy for old ones \citep{delange2021continual}. Continuous process improvement requires updates of quality control models over the entire life cycle of a product. 

\noindent \textbf{Human feedback and expert knowledge:}  Upon the detection of an anomaly, most operators will do a post-inspection about the diagnosis results (i.e., false positive or false negative). Improving models upon such expert feedback will be of importance to IoFT. Indeed, much like data in IoFT, human knowledge is often decentralized, with different entities having expert knowledge on different elements of a system. Therefore, modeling approaches that combine expert knowledge, human feedback and data-driven models are needed. Such models have evolved recently under the notion of expert or physics-guided data-driven modeling. However, they are yet to be explored under the IoFT paradigm.

\noindent \textbf{Quality Control:} As described in Sec. \ref{sec:manufacturing}, quality control will immensely benefit from a shared library of knowledge, be it a library of in-control behaviors, common anomalies, root causes, etc. Many companies are reluctant to collaborate in building such a library due to privacy constraints. IoFT may bring this end goal to fruition. 

In conclusion, quality control is set to greatly benefit from IoFT. Yet many challenges still need to be tackled to realize its potential.

\subsection{\emph{Computing}} \label{sec:computing}
Intending to gain insights without exposing raw data, large technology companies such as Google, Apple, and Firefox started to deploy FL for computer vision and natural language processing tasks across user devices \citep{apple-dp, ggkeyboard, keyboard-acl, firefox}; others, including NVIDIA, apply FL to create medical imaging AI \cite{nvidia-fl}; smart cities perform in-situ image training and testing on AI cameras to avoid expensive data migration \citep{focus, camera-cluster, fed-oj}; and video streaming solutions use FL to interpret and react to network conditions \citep{puffer}.
However, we believe that these applications of FL are only scratching the surface, given that the applications of ML in computing are even broader, many of which can be deployed more widely and improved by leveraging FL.
In the following, we present an incomplete overview of FL's many existing and potential applications in computing.
The common theme across many of them is enabling information sharing between multiple administrative domains without sharing raw private data.

{\bf Databases:}
Indexes play a critical role in speeding up query processing in database management systems (DBMS).
In recent years, learned indexes are gaining popularity, whereby an ML model replaces traditional index structures including B-Tree, Hash-Table, Bitmap, and so on. 
These learned indexes can be classified into two broad categories: static, read-only indexes \citep{learned-index} focus more on read-heavy workloads, while updatable indexes \citep{alex} can handle lookups as well as inserts and deletes common in write-heavy workloads.
Nonetheless, all of these works focus on applying ML to a single administrative domain, which restricts the use of learned indexes to scenarios that have already been observed within the domain and leaves them with potentially weaker performance on previously unseen workloads. 
Applying FL in this context will help collaborative training among multiple competing domains without sharing raw data.
Indexes are only a part of the many research challenges faced in the database literature, and ML and FL can have possible applications in, among others, transaction processing, lock management, query planning/optimization, and cardinality estimation.

{\bf Networking:}
Networks are inherently distributed, and networking protocols are no exception. 
As a result, FL is a natural fit for many networking problems where ML can be applied and has already been applied in limited scope (e.g., not being able to copy all data to a centralized location). 
Over the past decade, many networking problems have relied on ML techniques; for example, to infer datacenter topology \citep{orchestra}, to determine hyperparameters for congestion control algorithms \citep{remy}, for Internet-scale congestion control using deep reinforcement learning \citep{drl-cc}, for leveraging single and multiple paths in an adaptive manner \cite{pcc, mpcc}, and for routing \citep{learning-routes}.
They primarily relied on a single trust domain (e.g., a data center, an Internet AS, etc.) where everything is controlled by and cooperates with a single entity within which data can be shared. 
FL can expand the scope of many of these algorithms to be applied at a broader scale via privacy-preserving learning that may incentivize multiple domains to collaborate to learn a global model and then personalize to their own needs. 

{\bf Cloud Computing:}
To cope with the increasing number of Internet users as well as IoT and edge devices, large organizations leverage tens to hundreds of data centers and edge sites. 
Collecting data related to end-user sessions, monitoring logs, and performance counters, and thereafter analyzing and personalizing this data can significantly improve the overall user experience.
Traditional approaches to ML require collecting all these data to a centralized cloud data center, which is often impossible due to bandwidth constraints and data privacy regulations. 
IoFT is the natural choice in this context to address both of these concerns \citep{sol, iridium}.

{\bf Video Analytics:}
Cameras deployed for traffic control and surveillance continuously record and analyze large volumes of recorded videos using video analytics \citep{focus, chameleon, awstream}, which has been made possible by recent advances in computer vision.
A key challenge in this context is training large models, typically in datacenters, before they are deployed in the wild. 
Traditional centralized training is expensive and narrow; the latter follows from the fact that the models are trained on relatively small training datasets.
With the advent of smart cameras in the edge, i.e., cameras with onboard or nearby computing capabilities, we can leverage FL to train models with much bigger training datasets, which can significantly improve the accuracy of the models and keep them continuously updated.

{\bf Video Streaming:}
Videos constitute the bulk of the Internet traffic today, and live video streaming is a major contributor to this category. 
Client-side video players typically employ a variety of adaptive bitrate (ABR) algorithms to optimize users' quality of experience.
Recent advances in ABR algorithms include using reinforcement learning to generate context-specific ABR algorithms \citep{pensieve} to more recently demonstrating that generating many such algorithms does not necessarily perform better than using FL to generate one model that works in conjunction with classic video streaming techniques \citep{puffer}.
A key research direction here would be adopting federated reinforcement learning to leverage the best of both approaches, and find a balance between the global and personalized ABR algorithms.

\subsection{\emph{Reliability}} \label{sec:reliability}

Reliability engineering is concerned with the failure behavior of a system under stated conditions. A failure can be catastrophic, meaning a complete, sudden, and often unexpected system breakdown, leading to significant or even total loss of system performance. It can also be a degradation-induced soft failure (e.g., the capacity drop of a lithium-ion battery). There are several ways to evaluate the reliability of a product, though generally, evaluation based on reliability data is most common. Reliability data are usually in the format of lifetime data or degradation data. However, in these datasets, failure data is scarce, given that most products are highly reliable and do not fail often. Nevertheless, IoFT, with its privacy-protecting protocols, provides a unique opportunity to overcome the challenges of scarce data in reliability engineering.  

Throughout the ages, reliability data have been classified as sensitive information by companies. With millions of products released in the marketplace by manufacturers, it is no secret that reliability data is both extensive and comprehensive. Yet, its sensitivity hinders its usability. IoFT provides a unique opportunity to enable knowledge sharing from available datasets without compromising its privacy in such a scenario. For instance, edge compute resources can be exploited to replace existing reliability databases (e.g., product lifetime) with summary statistics (or prior) distribution of modeling parameters for each product/component.
Further, scenarios exist where products have only a few lead manufacturers (e.g., the smartphones industry). In here, the designs from different manufacturers are distinct, implying a certain degree of heterogeneity \citep{ye2013heterogeneities}. For a smartphone manufacturer, the reliability information from previous generations of smartphone products can be more beneficial than information from other manufacturers. Such a setting poses another unique challenge for IoFT. In the following, we will discuss the potential applications of IoFT in three different settings: (i) among a group of manufacturers producing a similar product, (ii) within a manufacturer, and (iii) within a reliability organization.

We first start with IoFT among a group of manufacturers. Consider reliability testing for evaluating a product's reliability, where reliability data is in the format of lifetime data subject to right censoring. Generally, the Weibull distribution with reliability function (scale $\alpha$ and shape $\beta$) and the lognormal distribution (location $\mu$ and scale $\sigma$) are two of the most commonly used distributions for describing a product's lifetime data \citep{meeker1998}. Moving forward, we will focus on the Weibull distribution, though similar logic applies to different distributions. The Weibull shape parameter $\beta$ is commonly believed to depend on the product type (i.e., failure mode due to the material used: e.g., corrosion of semiconductor material) or the failure mode due to customer usage (e.g., the user breaks their cellphone). Such parameter can be regarded as insensitive information to the product's reliability. On the other hand, the Weibull scale parameter $\alpha$ (also known as the characteristic life) is usually dependent on the effort of reliability investment from the manufacturers \citep{meeker1998}. Suppose a manufacturer uses local data to evaluate product reliability. In that case, both parameters in the lifetime distribution have to be estimated, and the uncertainties in both parameters will affect the precision of the final reliability evaluation. Then, it is reasonable to advocate sharing information on $\beta$ to decrease uncertainty in $\beta$, which eventually helps all manufacturers achieve a more accurate evaluation of product reliability. Additionally, since the information on $\alpha$ is unshared, a manufacturer cannot infer the product reliability of other manufacturers. 

\begin{figure}[!htb] \label{fig:reliability}
	\centering
	\includegraphics[width=0.4\textwidth]{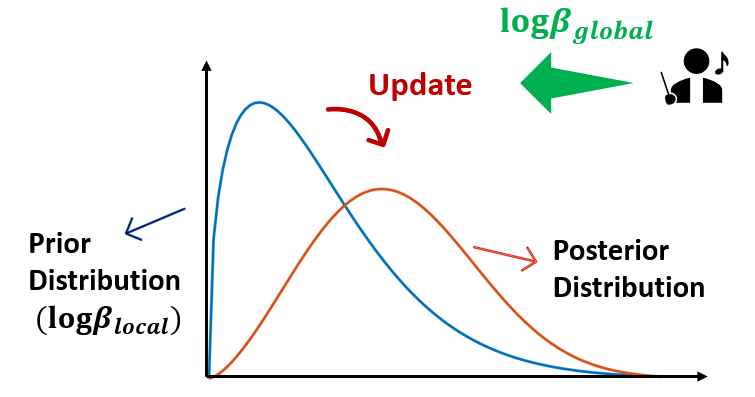}
    \caption{A schematic of FL in reliability testing by using Weibull as the global model.}
\end{figure}

Operationally, we can use a Bayesian approach. Let us consider the Weibull distribution for demonstration and provide a rough sketch of the parameter updating process in an IoFT system. First, in large sample sizes, the posterior distribution of $\log\beta$ can be well approximated by a normal distribution ($\log\beta$ ensures the positiveness of $\beta$). Afterward, when a manufacturer has recently conducted a life test and requests an update, or when the central server randomly chooses a manufacturer and mandates an update, the manufacturer will first get a broadcast of the current posterior distribution of $\log\beta$. The manufacturer can then use this posterior distribution of $\beta$ and the manufacturer's local posterior distribution on $\alpha$, which might be obtained from previous product testing, as a prior distribution for the newly collected reliability data. A routine Bayesian update gives the new posterior of $\alpha$ and $\log\beta$.
Then, the manufacturer can compute the mean and variance of the new posterior of $\log\beta$, and return the updated posterior distribution to the central server. Finally, the central server can then check the discrepancy between the broadcasted and the updated distributions to safeguard against data corruption during transmission or malicious attacks. If acceleration is used in the life test, then parameters in the acceleration model (as the activation energy in the Arrhenius model) contain no sensitive reliability information. Thus, such parameters can also be federated together with the Weibull shape. The same idea can be extended to accelerated degradation testing, where FL can be applied to the shape parameter of the mean degradation paths and the acceleration parameters. Fig. \ref{fig:reliability} provides a schematic view of the discussed protocol. 

Next, we explore the application of IoFT within a manufacturer. The underlying idea is that when a certain product is sold to customers, the collection of user data for early prediction of product failure must comply with some privacy terms, thus being restricted. Given the computational and communication capabilities of the product, IoFT provides a unique advantage in the presence of privacy constraints. Consider as an example, lithium-ion batteries which are widely used in electric vehicles. It is well-known that the usage pattern has a significant impact on the state-of-charge (short-term) and the remaining useful life (long-term) of lithium batteries \citep{peng2019bayesian}. However, it is almost impossible to associate the usage pattern to these two important performance characteristics because of difficulties replicating the heterogeneity in users' behavior. FL provides an opportunity to train an accurate model for each characteristic. To do so, we need a global statistical model that associates the customer usage pattern, the charge-discharge pattern, and the ambient environments to the performance characteristics. The global statistical model can be a random-effects model that allows for heterogeneity among customers. Then the approaches introduced in Secs. \ref{sec:global} and \ref{sec:personlized} can be used to learn a global model or a personalized model. Further, due to the different ambient environments of the users, there will be covariate shifts among users. Methods reviewed in Sec. \ref{sec:personlized} can be perfectly adopted to solve these problems.

Third, we discuss IoFT implementation on reliability organizations. The major of reliability organizations collect field failure data on a large variety of components from various sources. The ultimate goal is to estimate the reliability of any new system based on the component reliability estimated from collected databases. Some large databases can be found in \citet{oreda2009}, \citet{nprd2011} and \citet{denson2014}. Since there are millions of components, the data reported in these databases are aggregated in such a way that only a few summary statistics are provided for each component. This aggregation is based on the assumptions of exponential distributions, and it makes the fitting of a Weibull distribution extremely difficult \citep{chen2016random, chen2017,chen2020parametric}. However, FL provides a better solution to build such a database. Instead of recording these summary statistics, we can first agree upon a distribution for a component and then maintain a posterior distribution for the parameters. For example, the inverse Gaussian and the Birnbaum-Saunders distributions are commonly used for mechanical components, and Weibull is the most popular distribution in reliability. A conjugate distribution for the parameters, or a normal distribution for the transformed parameters (to ensure positivity), can be adopted for ease of use. Every time a partner of the database has new data to update, the database (which serves as the central server) can broadcast the current posterior of parameters for the component. The partner can then use this as prior and update the posterior with local data. The above rough idea can be materialized with the framework discussed in Sec. \ref{sec:global}.

In a nutshell, reliability of a manufactured product is usually shrouded with privacy concerns. Thus, implementing IoFT within a manufacturer is promising in solving the issues of data transmission and user privacy. On the other hand, IoFT across manufacturers is much more difficult. Nevertheless, with proper design of the information-sharing mechanisms, IoFT can tremendously help manufacturers increase the accuracy of reliability estimation and prediction without sacrificing confidentiality.

\bibliographystyle{plainnat}
\bibliography{references}

\begin{thebibliography}{368}
\providecommand{\natexlab}[1]{#1}
\providecommand{\url}[1]{\texttt{#1}}
\expandafter\ifx\csname urlstyle\endcsname\relax
  \providecommand{\doi}[1]{doi: #1}\else
  \providecommand{\doi}{doi: \begingroup \urlstyle{rm}\Url}\fi

\bibitem[syn()]{sync}
Ford sync.
\newblock \url{https://www.ford.com/technology/sync/}.
\newblock Accessed: 2020-07-18.

\bibitem[Abdolmaleki et~al.(2019)Abdolmaleki, Masoud, and
  Yin]{abdolmaleki2019vehicle}
Mojtaba Abdolmaleki, Neda Masoud, and Yafeng Yin.
\newblock Vehicle-to-vehicle wireless power transfer: Paving the way toward an
  electrified transportation system.
\newblock \emph{Transportation Research Part C: Emerging Technologies},
  103:\penalty0 261--280, 2019.

\bibitem[Acar et~al.(2019)Acar, Zhao, Navarro, Mattina, Whatmough, and
  Saligrama]{acarfederated}
Durmus Alp~Emre Acar, Yue Zhao, Ramon~Matas Navarro, Matthew Mattina, Paul~N
  Whatmough, and Venkatesh Saligrama.
\newblock Federated learning based on dynamic regularization.
\newblock In \emph{International Conference on Learning Representation}, 2019.

\bibitem[Al~Islam et~al.(2020)Al~Islam, Hajbabaie, and Aziz]{Intersection_3}
SMA~Bin Al~Islam, Ali Hajbabaie, and HM~Abdul Aziz.
\newblock A real-time network-level traffic signal control methodology with
  partial connected vehicle information.
\newblock \emph{Transportation Research Part C: Emerging Technologies},
  121:\penalty0 102830, 2020.

\bibitem[Aledhari et~al.(2020)Aledhari, Razzak, Parizi, and
  Saeed]{aledhari2020federated}
Mohammed Aledhari, Rehma Razzak, Reza~M Parizi, and Fahad Saeed.
\newblock Federated learning: A survey on enabling technologies, protocols, and
  applications.
\newblock \emph{IEEE Access}, 8:\penalty0 140699--140725, 2020.

\bibitem[Alshelahi et~al.(2021)Alshelahi, Wang, Yu, Byon, and
  Saigal]{alshelahi2021integrative}
Abdullah Alshelahi, Jingxing Wang, Mingdi Yu, Eunshin Byon, and Romesh Saigal.
\newblock Integrative density forecast and uncertainty quantification of wind
  power generation.
\newblock \emph{to appear in IEEE Transactions on Sustainable Energy}, 2021.

\bibitem[Apple(2019)]{appleprivacy}
Apple.
\newblock Designing for privacy.
\newblock \url{https://developer.apple.com/videos/play/wwdc2019/708}, 2019.
\newblock Accessed: 2021-04-21.

\bibitem[Arivazhagan et~al.(2019)Arivazhagan, Aggarwal, Singh, and
  Choudhary]{finetune}
Manoj~Ghuhan Arivazhagan, Vinay Aggarwal, Aaditya~Kumar Singh, and Sunav
  Choudhary.
\newblock Federated evaluation of on-device personalization.
\newblock \emph{arXiv preprint arXiv:1912.00818}, 2019.

\bibitem[Atzori et~al.(2010)Atzori, Iera, and Morabito]{atzori2010internet}
Luigi Atzori, Antonio Iera, and Giacomo Morabito.
\newblock The internet of things: A survey.
\newblock \emph{Computer networks}, 54\penalty0 (15):\penalty0 2787--2805,
  2010.

\bibitem[AWS(2019)]{amazonaws}
AWS.
\newblock What is aws?
\newblock \url{https://www.youtube.com/watch?v=a9__D53WsUs}, 2019.
\newblock Accessed: 2020-07-18.

\bibitem[AWS(2021)]{awsamazon}
AWS.
\newblock Amazon web services (aws).
\newblock \url{https://aws.amazon.com/}, 2021.
\newblock Accessed: 2020-07-18.

\bibitem[Azure(2018)]{azurevideo}
Azure.
\newblock How does microsoft azure work?
\newblock \url{https://www.youtube.com/watch?v=KXkBZCe699A}, 2018.
\newblock Accessed: 2020-07-18.

\bibitem[Bach(2008)]{Bach08a}
F.~Bach.
\newblock Consistency of the group {L}asso and multiple kernel learning.
\newblock \emph{Journal of Machine Learning Research}, 9:\penalty0 1179--1225,
  2008.

\bibitem[Bagdasaryan et~al.(2020)Bagdasaryan, Veit, Hua, Estrin, and
  Shmatikov]{bagdasaryan2020backdoor}
Eugene Bagdasaryan, Andreas Veit, Yiqing Hua, Deborah Estrin, and Vitaly
  Shmatikov.
\newblock How to backdoor federated learning.
\newblock In \emph{International Conference on Artificial Intelligence and
  Statistics}, pages 2938--2948, 2020.

\bibitem[Baharlouei et~al.(2020)Baharlouei, Nouiehed, Beirami, and
  Razaviyayn]{Renyi}
Sina Baharlouei, Maher Nouiehed, Ahmad Beirami, and Meisam Razaviyayn.
\newblock R\'enyi fair inference.
\newblock 2020.

\bibitem[Bartlett and Mendelson(2002)]{rademacher}
Peter~L. Bartlett and Shahar Mendelson.
\newblock Rademacher and gaussian complexities: Risk bounds and structural
  results.
\newblock \emph{Journal of Machine Learning Research}, 2002.

\bibitem[Bauer et~al.(2017)Bauer, Rojas-Carulla, {\'S}wi{k{a}}tkowski,
  Sch{\"o}lkopf, and Turner]{bauer2017discriminative}
Matthias Bauer, Mateo Rojas-Carulla, Jakub~Bart{l}omiej {\'S}wi{k{a}}tkowski,
  Bernhard Sch{\"o}lkopf, and Richard~E Turner.
\newblock Discriminative k-shot learning using probabilistic models.
\newblock \emph{arXiv preprint arXiv:1706.00326}, 2017.

\bibitem[Baumann and Roller(2017)]{baumann2017additive}
Felix~W Baumann and Dieter Roller.
\newblock Additive manufacturing, cloud-based 3d printing and associated
  services—overview.
\newblock \emph{Journal of Manufacturing and Materials Processing}, 1\penalty0
  (2):\penalty0 15, 2017.

\bibitem[Behrendt et~al.(2017)Behrendt, Kadocsa, Kelly, and
  Schirmers]{behrendt2017achieve}
A~Behrendt, A~Kadocsa, R~Kelly, and L~Schirmers.
\newblock How to achieve and sustain the impact of digital manufacturing at
  scale.
\newblock \emph{McKinsey\&Company Quartley}, 2017.

\bibitem[Beimel et~al.(2019)Beimel, Korolova, Nissim, Sheffet, and
  Stemmer]{beimel2019power}
Amos Beimel, Aleksandra Korolova, Kobbi Nissim, Or~Sheffet, and Uri Stemmer.
\newblock The power of synergy in differential privacy: Combining a small
  curator with local randomizers.
\newblock \emph{arXiv preprint arXiv:1912.08951}, 2019.

\bibitem[Bellet et~al.(2018)Bellet, Guerraoui, Taziki, and
  Tommasi]{bellet2018personalized}
Aur{\'e}lien Bellet, Rachid Guerraoui, Mahsa Taziki, and Marc Tommasi.
\newblock Personalized and private peer-to-peer machine learning.
\newblock In \emph{International Conference on Artificial Intelligence and
  Statistics}, pages 473--481. PMLR, 2018.

\bibitem[Bhandari et~al.(2012)Bhandari, Shrestha, and
  New]{bhandari2012evaluation}
Mahabir Bhandari, Som Shrestha, and Joshua New.
\newblock Evaluation of weather datasets for building energy simulation.
\newblock \emph{Energy and Buildings}, 49:\penalty0 109--118, 2012.

\bibitem[Bhowmick et~al.(2018)Bhowmick, Duchi, Freudiger, Kapoor, and
  Rogers]{bhowmick2018protection}
Abhishek Bhowmick, John Duchi, Julien Freudiger, Gaurav Kapoor, and Ryan
  Rogers.
\newblock Protection against reconstruction and its applications in private
  federated learning.
\newblock \emph{arXiv preprint arXiv:1812.00984}, 2018.

\bibitem[Bickel and Ritov(2000)]{Bickelbic00}
P.~J. Bickel and Y.~Ritov.
\newblock Non- and semiparametric statistics: compared and contrasted.
\newblock \emph{J. Stat. Plan. Infer.}, 91:\penalty0 209--228, 2000.

\bibitem[Biegel et~al.(2014)Biegel, Westenholz, Hansen, Stoustrup, Andersen,
  and Harbo]{BIEGEL2014}
Benjamin Biegel, Mikkel Westenholz, Lars~Henrik Hansen, Jakob Stoustrup, Palle
  Andersen, and Silas Harbo.
\newblock Integration of flexible consumers in the ancillary service markets.
\newblock \emph{Energy}, 67:\penalty0 479--489, 2014.

\bibitem[Bittau et~al.(2017)Bittau, Erlingsson, Maniatis, Mironov, Raghunathan,
  Lie, Rudominer, Kode, Tinnes, and Seefeld]{bittau2017prochlo}
Andrea Bittau, {\'U}lfar Erlingsson, Petros Maniatis, Ilya Mironov, Ananth
  Raghunathan, David Lie, Mitch Rudominer, Ushasree Kode, Julien Tinnes, and
  Bernhard Seefeld.
\newblock Prochlo: Strong privacy for analytics in the crowd.
\newblock In \emph{Proceedings of the 26th Symposium on Operating Systems
  Principles}, pages 441--459, 2017.

\bibitem[Blei et~al.(2017)Blei, Kucukelbir, and McAuliffe]{blei2017variational}
David~M Blei, Alp Kucukelbir, and Jon~D McAuliffe.
\newblock Variational inference: A review for statisticians.
\newblock \emph{Journal of the American statistical Association}, 112\penalty0
  (518):\penalty0 859--877, 2017.

\bibitem[Brat et~al.(2020)Brat, Weber, Gehlenborg, Avillach, Palmer, Chiovato,
  Cimino, Waitman, Omenn, Malovini, et~al.]{brat2020international}
Gabriel~A Brat, Griffin~M Weber, Nils Gehlenborg, Paul Avillach, Nathan~P
  Palmer, Luca Chiovato, James Cimino, Lemuel~R Waitman, Gilbert~S Omenn,
  Alberto Malovini, et~al.
\newblock International electronic health record-derived covid-19 clinical
  course profiles: the 4ce consortium.
\newblock \emph{Npj Digital Medicine}, 3\penalty0 (1):\penalty0 1--9, 2020.

\bibitem[Brisimi et~al.(2018)Brisimi, Chen, Mela, Olshevsky, Paschalidis, and
  Shi]{brisimi2018federated}
Theodora~S Brisimi, Ruidi Chen, Theofanie Mela, Alex Olshevsky, Ioannis~Ch
  Paschalidis, and Wei Shi.
\newblock Federated learning of predictive models from federated electronic
  health records.
\newblock \emph{International journal of medical informatics}, 112:\penalty0
  59--67, 2018.

\bibitem[Bubeck and Cesa-Bianchi(2012)]{banditsurvey}
Sebastien Bubeck and Nicolo Cesa-Bianchi.
\newblock Regret analysis of stochastic and nonstochastic multi-armed bandit
  problems.
\newblock \emph{Foundations and Trends in Machine Learning}, 5\penalty0
  (1):\penalty0 1--122, 2012.

\bibitem[Buckholtz et~al.(2015)Buckholtz, Ragai, and Wang]{buckholtz2015cloud}
Ben Buckholtz, Ihab Ragai, and Lihui Wang.
\newblock Cloud manufacturing: current trends and future implementations.
\newblock \emph{Journal of Manufacturing Science and Engineering}, 137\penalty0
  (4), 2015.

\bibitem[Buhlmann and van~de Geer(2011)]{BuhlmannVDGBook}
P.~Buhlmann and S.~van~de Geer.
\newblock \emph{Statistical for High-Dimensional Data}.
\newblock Springer Series in Statistics. Springer, New York, 2011.

\bibitem[Byon et~al.(2015)Byon, Choe, and Yampikulsakul]{byon2015adaptive}
Eunshin Byon, Youngjun Choe, and Nattavut Yampikulsakul.
\newblock Adaptive learning in time-variant processes with application to wind
  power systems.
\newblock \emph{IEEE Transactions on Automation Science and Engineering},
  13\penalty0 (2):\penalty0 997--1007, 2015.

\bibitem[Byrd et~al.(2016)Byrd, Hansen, Nocedal, and Singer]{SQN}
Richard~H Byrd, Samantha~L Hansen, Jorge Nocedal, and Yoram Singer.
\newblock A stochastic quasi-newton method for large-scale optimization.
\newblock \emph{SIAM Journal on Optimization}, 26\penalty0 (2):\penalty0
  1008--1031, 2016.

\bibitem[Carli and Dotoli(2019)]{Carli2019}
Raffaele Carli and Mariagrazia Dotoli.
\newblock Decentralized control for residential energy management of a smart
  users' microgrid with renewable energy exchange.
\newblock \emph{IEEE/CAA Journal of Automatica Sinica}, 6\penalty0
  (3):\penalty0 641--656, 2019.

\bibitem[Caruana(1997)]{caruana1997multitask}
Rich Caruana.
\newblock Multitask learning.
\newblock \emph{Machine learning}, 28\penalty0 (1):\penalty0 41--75, 1997.

\bibitem[Castillo et~al.(2019)Castillo, Joseph, and
  Kalidindi]{castillo2019bayesian}
Andrew~R Castillo, V~Roshan Joseph, and Surya~R Kalidindi.
\newblock Bayesian sequential design of experiments for extraction of
  single-crystal material properties from spherical indentation measurements on
  polycrystalline samples.
\newblock \emph{JOM}, 71\penalty0 (8):\penalty0 2671--2679, 2019.

\bibitem[Chang et~al.(2018)Chang, Balachandar, Lam, Yi, Brown, Beers, Rosen,
  Rubin, and Kalpathy-Cramer]{chang2018distributed}
Ken Chang, Niranjan Balachandar, Carson Lam, Darvin Yi, James Brown, Andrew
  Beers, Bruce Rosen, Daniel~L Rubin, and Jayashree Kalpathy-Cramer.
\newblock Distributed deep learning networks among institutions for medical
  imaging.
\newblock \emph{Journal of the American Medical Informatics Association},
  25\penalty0 (8):\penalty0 945--954, 2018.

\bibitem[Chen et~al.(2018)Chen, Luo, Dong, Li, and He]{chen2018federated}
Fei Chen, Mi~Luo, Zhenhua Dong, Zhenguo Li, and Xiuqiang He.
\newblock Federated meta-learning with fast convergence and efficient
  communication.
\newblock \emph{arXiv preprint arXiv:1802.07876}, 2018.

\bibitem[Chen et~al.(2015)Chen, Deng, Wan, Zhang, Vasilakos, and
  Rong]{chen2015data}
Feng Chen, Pan Deng, Jiafu Wan, Daqiang Zhang, Athanasios~V Vasilakos, and
  Xiaohui Rong.
\newblock Data mining for the internet of things: literature review and
  challenges.
\newblock \emph{International Journal of Distributed Sensor Networks},
  11\penalty0 (8):\penalty0 431047, 2015.

\bibitem[Chen et~al.(2020{\natexlab{a}})Chen, Zheng, Al~Kontar, and
  Raskutti]{chen2020stochastic}
Hao Chen, Lili Zheng, Raed Al~Kontar, and Garvesh Raskutti.
\newblock Stochastic gradient descent in correlated settings: A study on
  gaussian processes.
\newblock \emph{Advances in neural information processing systems},
  2020{\natexlab{a}}.

\bibitem[Chen and Chao(2021)]{chen2020fedbe}
Hong-You Chen and Wei-Lun Chao.
\newblock Fedbe: Making bayesian model ensemble applicable to federated
  learning.
\newblock \emph{International Conference on Learning Representation}, 2021.

\bibitem[Chen et~al.(2019{\natexlab{a}})Chen, Mathews, Ouyang, and
  Beaufays]{chen2019federated}
Mingqing Chen, Rajiv Mathews, Tom Ouyang, and Fran{\c{c}}oise Beaufays.
\newblock Federated learning of out-of-vocabulary words.
\newblock \emph{arXiv preprint arXiv:1903.10635}, 2019{\natexlab{a}}.

\bibitem[Chen et~al.(2019{\natexlab{b}})Chen, Suresh, Mathews, Wong, Allauzen,
  Beaufays, and Riley]{keyboard-acl}
Mingqing Chen, Ananda~Theertha Suresh, Rajiv Mathews, Adeline Wong, Cyril
  Allauzen, Francoise Beaufays, and Michael Riley.
\newblock Federated learning of n-gram language models.
\newblock In \emph{ACL}, 2019{\natexlab{b}}.

\bibitem[Chen and Ye(2016)]{chen2016random}
Piao Chen and Zhi-Sheng Ye.
\newblock Random effects models for aggregate lifetime data.
\newblock \emph{IEEE Transactions on Reliability}, 66\penalty0 (1):\penalty0
  76--83, 2016.

\bibitem[Chen and Ye(2017)]{chen2017}
Piao Chen and Zhi-Sheng Ye.
\newblock Estimation of field reliability based on aggregate lifetime data.
\newblock \emph{Technometrics}, 59\penalty0 (1):\penalty0 115--125, 2017.
\newblock \doi{10.1080/00401706.2015.1096827}.
\newblock URL \url{http://dx.doi.org/10.1080/00401706.2015.1096827}.

\bibitem[Chen et~al.(2020{\natexlab{b}})Chen, Ye, and Zhai]{chen2020parametric}
Piao Chen, Zhi-Sheng Ye, and Qingqing Zhai.
\newblock Parametric analysis of time-censored aggregate lifetime data.
\newblock \emph{IISE Transactions}, 52\penalty0 (5):\penalty0 516--527,
  2020{\natexlab{b}}.

\bibitem[Chen et~al.(2017{\natexlab{a}})Chen, Zhang, Sharma, Yi, and
  Hsieh]{ZOO-ANN}
Pin-Yu Chen, Huan Zhang, Yash Sharma, Jinfeng Yi, and Cho-Jui Hsieh.
\newblock Zoo: Zeroth order optimization based black-box attacks to deep neural
  networks without training substitute models.
\newblock In \emph{Proceedings of the 10th ACM workshop on artificial
  intelligence and security}, pages 15--26, 2017{\natexlab{a}}.

\bibitem[Chen et~al.(2017{\natexlab{b}})Chen, Wang, and Wu]{chen2017sequential}
Ray-Bing Chen, Weichung Wang, and CF~Jeff Wu.
\newblock Sequential designs based on bayesian uncertainty quantification in
  sparse representation surrogate modeling.
\newblock \emph{Technometrics}, 59\penalty0 (2):\penalty0 139--152,
  2017{\natexlab{b}}.

\bibitem[Chen et~al.(2019{\natexlab{c}})Chen, Liu, Kira, Wang, and
  Huang]{chen2019closer}
Wei-Yu Chen, Yen-Cheng Liu, Zsolt Kira, Yu-Chiang~Frank Wang, and Jia-Bin
  Huang.
\newblock A closer look at few-shot classification.
\newblock \emph{arXiv preprint arXiv:1904.04232}, 2019{\natexlab{c}}.

\bibitem[Chen et~al.(2020{\natexlab{c}})Chen, Horvath, and
  Richtarik]{clientsamplingwithgn}
Wenlin Chen, Samuel Horvath, and Peter Richtarik.
\newblock Optimal client sampling for federated learning.
\newblock \emph{arXiv preprint arXiv:2010.13723}, 2020{\natexlab{c}}.

\bibitem[Chen et~al.(2017{\natexlab{c}})Chen, Liu, Li, Lu, and
  Song]{chen2017targeted}
Xinyun Chen, Chang Liu, Bo~Li, Kimberly Lu, and Dawn Song.
\newblock Targeted backdoor attacks on deep learning systems using data
  poisoning.
\newblock \emph{arXiv preprint arXiv:1712.05526}, 2017{\natexlab{c}}.

\bibitem[Cho et~al.(2020)Cho, Wang, and Joshi]{clientsamplewithloss}
Yae~Jee Cho, Jianyu Wang, and Gauri Joshi.
\newblock Client selection in federated learning: Convergence analysis and
  power-of-choice selection strategies.
\newblock \emph{arXiv preprint arXiv:2010.01243}, 2020.

\bibitem[Choe et~al.(2016)Choe, Guo, Byon, Jin, and Li]{choe2016change}
Youngjun Choe, Weihong Guo, Eunshin Byon, Jionghua Jin, and Jingjing Li.
\newblock Change-point detection on solar panel performance using thresholded
  lasso.
\newblock \emph{Quality and Reliability Engineering International}, 32\penalty0
  (8):\penalty0 2653--2665, 2016.

\bibitem[Chowdhury et~al.(2011)Chowdhury, Zaharia, Ma, Jordan, and
  Stoica]{orchestra}
Mosharaf Chowdhury, Matei Zaharia, Justin Ma, Michael~I Jordan, and Ion Stoica.
\newblock Managing data transfers in computer clusters with {Orchestra}.
\newblock In \emph{ACM SIGCOMM}, 2011.

\bibitem[CNBC(2020)]{fordcorona}
CNBC.
\newblock Ford temporarily closes two plants after three workers test positive
  for coronavirus.
\newblock
  \url{https://www.cnbc.com/2020/05/20/ford-closes-chicago-plant-after-two-workers-test-positive-for-covid-19.html},
  2020.
\newblock Accessed: 2020-07-18.

\bibitem[CNN(2020)]{12yearold}
CNN.
\newblock 12-year-old boy 3d prints masks for frontline workers.
\newblock
  \url{https://www.cnn.com/videos/us/2020/04/25/coronavirus-3d-print-ppe-12-year-old-pkg-whitfield-vpx.cnn},
  2020.
\newblock Accessed: 2020-07-18.

\bibitem[Combettes(2018)]{monotoneoperator}
P.~L. Combettes.
\newblock Monotone operator theory in convex optimization.
\newblock \emph{Math. Program.}, 2018.

\bibitem[Conn et~al.(2009)Conn, Scheinberg, and Vicente]{conn2009introduction}
Andrew~R Conn, Katya Scheinberg, and Luis~N Vicente.
\newblock \emph{Introduction to derivative-free optimization}.
\newblock SIAM, 2009.

\bibitem[Copeland(2019)]{wallst}
Rob Copeland.
\newblock Google's ``project nightingale'' gathers personal health data on
  millions of americans.
\newblock
  \url{https://www.wsj.com/articles/google-s-secret-project-nightingale-gathers-personal-health-data-on-millions-of-americans-11573496790},
  2019.
\newblock Accessed: 2021-04-25.

\bibitem[Cordeau and Laporte(2007)]{decomposition_4}
Jean-Fran{\c{c}}ois Cordeau and Gilbert Laporte.
\newblock The dial-a-ride problem: models and algorithms.
\newblock \emph{Annals of operations research}, 153\penalty0 (1):\penalty0
  29--46, 2007.

\bibitem[Correa et~al.(2018)Correa, Toro, and Ferreira]{correa2018new}
Jorge~E Correa, Ricardo Toro, and Placid~M Ferreira.
\newblock A new paradigm for organizing networks of computer numerical control
  manufacturing resources in cloud manufacturing.
\newblock \emph{Procedia Manufacturing}, 26:\penalty0 1318--1329, 2018.

\bibitem[Culver and Westcott(2020)]{civilian3d}
David Culver and Ben Westcott.
\newblock 3d printing enthusiasts are working from home to help hospitals fight
  coronavirus.
\newblock
  \url{https://www.cnn.com/2020/04/18/tech/us-coronavirus-ventilator-3d-printer-intl-hnk/index.html},
  2020.
\newblock Accessed: 2020-07-18.

\bibitem[Delange et~al.(2021)Delange, Aljundi, Masana, Parisot, Jia, Leonardis,
  Slabaugh, and Tuytelaars]{delange2021continual}
Matthias Delange, Rahaf Aljundi, Marc Masana, Sarah Parisot, Xu~Jia, Ales
  Leonardis, Greg Slabaugh, and Tinne Tuytelaars.
\newblock A continual learning survey: Defying forgetting in classification
  tasks.
\newblock \emph{IEEE Transactions on Pattern Analysis and Machine
  Intelligence}, 2021.

\bibitem[Deng et~al.(2020)Deng, Kamani, and Mahdavi]{deng2020adaptive}
Yuyang Deng, Mohammad~Mahdi Kamani, and Mehrdad Mahdavi.
\newblock Adaptive personalized federated learning.
\newblock \emph{arXiv preprint arXiv:2003.13461}, 2020.

\bibitem[Denson et~al.(2014)Denson, Crowell, Jaworski, and Mahar]{denson2014}
William Denson, William Crowell, Paul Jaworski, and David Mahar.
\newblock \emph{Electronic Parts Reliability Data 2014}.
\newblock Reliability Information Analysis Center, Rome, NY, 2014.

\bibitem[Differential Privacy~Team(2017)]{apple-dp}
Apple Differential Privacy~Team.
\newblock Learning with privacy at scale.
\newblock In \emph{Apple Machine Learning Journal}, 2017.

\bibitem[Ding et~al.(2017)Ding, Kulkarni, and Yekhanin]{ding2017collecting}
Bolin Ding, Janardhan Kulkarni, and Sergey Yekhanin.
\newblock Collecting telemetry data privately.
\newblock In \emph{Advances in Neural Information Processing Systems}, pages
  3571--3580, 2017.

\bibitem[Ding et~al.(2020)Ding, Minhas, Yu, Wang, Do, Li, Zhang, Chandramouli,
  Gehrke, Kossmann, et~al.]{alex}
Jialin Ding, Umar~Farooq Minhas, Jia Yu, Chi Wang, Jaeyoung Do, Yinan Li,
  Hantian Zhang, Badrish Chandramouli, Johannes Gehrke, Donald Kossmann, et~al.
\newblock {ALEX}: an updatable adaptive learned index.
\newblock In \emph{ACM SIGMOD}, 2020.

\bibitem[Dinh et~al.(2020)Dinh, Tran, and Nguyen]{pfedme}
Canh~T. Dinh, Nguyen~H. Tran, and Tuan~Dung Nguyen.
\newblock Personalized federated learning with moreau envelopes.
\newblock In \emph{34th Conference on Neural Information Processing Systems},
  2020.

\bibitem[Dolezal(2020)]{cant3dprint}
Jakub Dolezal.
\newblock 3d printed face shields for medics and professionals - join us!
\newblock
  \url{https://forum.prusaprinters.org/forum/coronavirus-covid-19/3d-printed-face-shields-for-medics-and-professionals-join-us/},
  2020.
\newblock Accessed: 2020-07-18.

\bibitem[Dong et~al.(2015)Dong, Li, Zarchy, Godfrey, and Schapira]{pcc}
Mo~Dong, Qingxi Li, Doron Zarchy, P~Brighten Godfrey, and Michael Schapira.
\newblock {PCC}: Re-architecting congestion control for consistent high
  performance.
\newblock In \emph{USENIX NSDI}, 2015.

\bibitem[Drineas and Mahoney(2005)]{DrinMah05}
P.~Drineas and M.~W. Mahoney.
\newblock On the {N}ystr\"{o}m method for approximating a {G}ram matrix for
  improved kernel-based learning.
\newblock \emph{Journal of Machine Learning Research}, 6:\penalty0 2153--2175,
  2005.

\bibitem[Drineas et~al.(2011)Drineas, Mahoney, Muthukrishnan, and
  Sarlos]{DrinMuthuMahSarlos11}
P.~Drineas, M.~W. Mahoney, S.~Muthukrishnan, and T.~Sarlos.
\newblock Faster least squares approximation.
\newblock \emph{Numerical Mathematics}, 117:\penalty0 219--249, 2011.

\bibitem[Du et~al.(2021)Du, Xu, Wu, and Tong]{du2021fairness}
Wei Du, Depeng Xu, Xintao Wu, and Hanghang Tong.
\newblock Fairness-aware agnostic federated learning.
\newblock In \emph{Proceedings of the 2021 SIAM International Conference on
  Data Mining (SDM)}, pages 181--189. SIAM, 2021.

\bibitem[Duan et~al.(2018)Duan, Yoon, and Okwudire]{duan2018limited}
Molong Duan, Deokkyun Yoon, and Chinedum~E Okwudire.
\newblock A limited-preview filtered b-spline approach to tracking
  control--with application to vibration-induced error compensation of a 3d
  printer.
\newblock \emph{Mechatronics}, 56:\penalty0 287--296, 2018.

\bibitem[Duan et~al.(2019)Duan, Liu, Chen, Tan, Ren, Qiao, and
  Liang]{duan2019astraea}
Moming Duan, Duo Liu, Xianzhang Chen, Yujuan Tan, Jinting Ren, Lei Qiao, and
  Liang Liang.
\newblock Astraea: Self-balancing federated learning for improving
  classification accuracy of mobile deep learning applications.
\newblock In \emph{2019 IEEE 37th International Conference on Computer Design
  (ICCD)}, pages 246--254. IEEE, 2019.

\bibitem[Edwards and Storkey(2017)]{edwards2017towards}
Harrison Edwards and Amos Storkey.
\newblock Towards a neural statistician.
\newblock \emph{International Conference on Learning Representations}, 2017.

\bibitem[EIA(2020{\natexlab{a}})]{energyfacts}
EIA.
\newblock {U.S} energy information administration, {U.S} energy facts
  explained.
\newblock \url{https://www.eia.gov/energyexplained/us-energy-facts/},
  2020{\natexlab{a}}.
\newblock Accessed: 2021-04-16.

\bibitem[EIA(2020{\natexlab{b}})]{todayenergy}
EIA.
\newblock {U.S} energy information administration, today in energy.
\newblock \url{https://www.eia.gov/todayinenergy/detail.php?id=42635},
  2020{\natexlab{b}}.
\newblock Accessed: 2021-03-29.

\bibitem[EIA(2021)]{consumption}
EIA.
\newblock {U.S} energy information administration, consumption and efficiency.
\newblock \url{https://www.eia.gov/consumption/}, 2021.
\newblock Accessed: 2021-04-16.

\bibitem[Fallah et~al.(2020)Fallah, Mokhtari, and
  Ozdaglar]{fallah2020personalized}
Alireza Fallah, Aryan Mokhtari, and Asuman Ozdaglar.
\newblock Personalized federated learning with theoretical guarantees: A
  model-agnostic meta-learning approach.
\newblock \emph{Advances in Neural Information Processing Systems}, 33, 2020.

\bibitem[Farooq and Hafeez(2020)]{finetunecovid}
Muhammad Farooq and Abdul Hafeez.
\newblock Covid-resnet: A deep learning framework for screening of covid19 from
  radiographs.
\newblock \emph{arXiv preprint arXiv:2003.14395}, 2020.

\bibitem[FDA(2020)]{3drapidresponse}
FDA.
\newblock 3d printing in fda’s rapid response to covid-19.
\newblock
  \url{https://www.fda.gov/emergency-preparedness-and-response/coronavirus-disease-2019-covid-19/3d-printing-fdas-rapid-response-covid-19},
  2020.
\newblock Accessed: 2020-07-18.

\bibitem[Feldman et~al.(2015)Feldman, Friedler, Moeller, Scheidegger, and
  Venkatasubramanian]{feldman2015certifying}
Michael Feldman, Sorelle~A Friedler, John Moeller, Carlos Scheidegger, and
  Suresh Venkatasubramanian.
\newblock Certifying and removing disparate impact.
\newblock In \emph{proceedings of the 21th ACM SIGKDD international conference
  on knowledge discovery and data mining}, pages 259--268, 2015.

\bibitem[Feng et~al.(2018)Feng, Bao, Jin, Sun, Saigusa, Tahmasbi-Sarvestani,
  and Dsa]{feng2018estimation}
Fred Feng, Shan Bao, Judy Jin, Wenbo Sun, Shigenobu Saigusa, Amin
  Tahmasbi-Sarvestani, and Jovin Dsa.
\newblock Estimation of lead vehicle kinematics using camera-based data for
  driver distraction detection.
\newblock \emph{International Journal of Automotive Engineering}, 9\penalty0
  (3):\penalty0 158--164, 2018.

\bibitem[Field(2000)]{demandsurge}
Karen Field.
\newblock Covid-19: How quickly can manufacturing respond to the surge in
  demand?
\newblock
  \url{https://www.fierceelectronics.com/electronics/how-quickly-can-manufacturing-respond-to-surge-demand},
  2000.
\newblock Accessed: 2020-07-18.

\bibitem[Finn et~al.(2017)Finn, Abbeel, and Levine]{finn2017model}
Chelsea Finn, Pieter Abbeel, and Sergey Levine.
\newblock Model-agnostic meta-learning for fast adaptation of deep networks.
\newblock In \emph{International Conference on Machine Learning}, pages
  1126--1135. PMLR, 2017.

\bibitem[Finn et~al.(2018)Finn, Xu, and Levine]{finn2018probabilistic}
Chelsea Finn, Kelvin Xu, and Sergey Levine.
\newblock Probabilistic model-agnostic meta-learning.
\newblock \emph{Advances In Neural Information Processing Systems}, 2018.

\bibitem[Fitzgerald and Schwinke(2016)]{GMIOT}
Michael Fitzgerald and Steve Schwinke.
\newblock General motors relies on iot to anticipate customers needs.
\newblock
  \url{https://sloanreview.mit.edu/article/general-motors-relies-on-iot-to-keep-its-customers-safe-and-secure/},
  2016.
\newblock Accessed: 2020-07-18.

\bibitem[Foster et~al.(2017)Foster, Burns, Grove, Kathan, Lee, Peirovi, and
  Schilling]{demandresponse}
Ben Foster, David Burns, Jadon Grove, David Kathan, Michael Lee, Samin Peirovi,
  and Cameron Schilling.
\newblock Assessment of demand response and advanced metering.
\newblock Technical report, Federal Energy Regulatory Commission, 2017.

\bibitem[Friedman et~al.(2001)Friedman, Hastie, Tibshirani,
  et~al.]{friedman2001elements}
Jerome Friedman, Trevor Hastie, Robert Tibshirani, et~al.
\newblock \emph{The elements of statistical learning}, volume~1.
\newblock Springer series in statistics New York, 2001.

\bibitem[Garcia et~al.(2019)Garcia, Mozaffar, Ren, Correa, Ehmann, Cao, and
  You]{garcia2019sustainable}
Daniel~J Garcia, Mojtaba Mozaffar, Huaqing Ren, Jorge~E Correa, Kornel Ehmann,
  Jian Cao, and Fengqi You.
\newblock Sustainable manufacturing with cyber-physical discrete manufacturing
  networks: Overview and modeling framework.
\newblock \emph{Journal of Manufacturing Science and Engineering}, 141\penalty0
  (2), 2019.

\bibitem[Garnelo et~al.(2018{\natexlab{a}})Garnelo, Rosenbaum, Maddison,
  Ramalho, Saxton, Shanahan, Teh, Rezende, and Eslami]{garnelo2018conditional}
Marta Garnelo, Dan Rosenbaum, Christopher Maddison, Tiago Ramalho, David
  Saxton, Murray Shanahan, Yee~Whye Teh, Danilo Rezende, and SM~Ali Eslami.
\newblock Conditional neural processes.
\newblock In \emph{International Conference on Machine Learning}, pages
  1704--1713. PMLR, 2018{\natexlab{a}}.

\bibitem[Garnelo et~al.(2018{\natexlab{b}})Garnelo, Schwarz, Rosenbaum, Viola,
  Rezende, Eslami, and Teh]{garnelo2018neural}
Marta Garnelo, Jonathan Schwarz, Dan Rosenbaum, Fabio Viola, Danilo~J Rezende,
  SM~Eslami, and Yee~Whye Teh.
\newblock Neural processes.
\newblock \emph{arXiv preprint arXiv:1807.01622}, 2018{\natexlab{b}}.

\bibitem[Gilad et~al.(2020)Gilad, Schiff, Godfrey, Raiciu, and Schapira]{mpcc}
Tomer Gilad, Neta~Rozen Schiff, P.~Brighten Godfrey, Costin Raiciu, and Michael
  Schapira.
\newblock Online learning multipath transport.
\newblock In \emph{ACM CoNEXT}, 2020.

\bibitem[Gon{\c{c}}alves et~al.(2016)Gon{\c{c}}alves, Von~Zuben, Banerjee,
  et~al.]{gonccalves2016multi}
Andr{\'e}~R Gon{\c{c}}alves, Fernando~J Von~Zuben, Arindam Banerjee, et~al.
\newblock Multi-task sparse structure learning with gaussian copula models.
\newblock \emph{Journal of Machine Learning Research}, 2016.

\bibitem[Goodfellow et~al.(2014)Goodfellow, Pouget-Abadie, Mirza, Xu,
  Warde-Farley, Ozair, Courville, and Bengio]{goodfellow2014generative}
Ian Goodfellow, Jean Pouget-Abadie, Mehdi Mirza, Bing Xu, David Warde-Farley,
  Sherjil Ozair, Aaron Courville, and Yoshua Bengio.
\newblock Generative adversarial nets.
\newblock \emph{Advances in neural information processing systems}, 27, 2014.

\bibitem[Google(2019)]{googlesupport}
Google.
\newblock Your chats stay private while messages improves suggestions.
\newblock
  \url{https://support.google.com/messages/answer/9327902?hl=en#zippy=}, 2019.
\newblock Accessed: 2021-04-21.

\bibitem[GoogleCloud(2021)]{googleiot}
GoogleCloud.
\newblock Googlecloud.
\newblock \url{https://cloud.google.com/solutions/iot}, 2021.
\newblock Accessed: 2020-07-18.

\bibitem[Gordon et~al.(2018)Gordon, Bronskill, Bauer, Nowozin, and
  Turner]{gordon2018meta}
Jonathan Gordon, John Bronskill, Matthias Bauer, Sebastian Nowozin, and
  Richard~E Turner.
\newblock Meta-learning probabilistic inference for prediction.
\newblock \emph{arXiv preprint arXiv:1805.09921}, 2018.

\bibitem[Gordon(1985)]{Gor85}
Y.~Gordon.
\newblock Some inequalities for {G}aussian processes and applications.
\newblock \emph{Israel Journal Math.}, 50:\penalty0 265--289, 1985.

\bibitem[Gramacy(2020)]{gramacy2020surrogates}
Robert~B Gramacy.
\newblock \emph{Surrogates: Gaussian process modeling, design, and optimization
  for the applied sciences}.
\newblock CRC Press, 2020.

\bibitem[Grant et~al.(2018)Grant, Finn, Levine, Darrell, and
  Griffiths]{grant2018recasting}
Erin Grant, Chelsea Finn, Sergey Levine, Trevor Darrell, and Thomas Griffiths.
\newblock Recasting gradient-based meta-learning as hierarchical bayes.
\newblock \emph{arXiv preprint arXiv:1801.08930}, 2018.

\bibitem[Hanzely and Richtarik(2021)]{lgd}
Filip Hanzely and Peter Richtarik.
\newblock Federated learning of a mixture of global and local models.
\newblock \emph{arXiv preprint arXiv:2002.05516}, 2021.

\bibitem[Hard et~al.(2018)Hard, Rao, Mathews, Ramaswamy, Beaufays, Augenstein,
  Eichner, Kiddon, and Ramage]{hard2018federated}
Andrew Hard, Kanishka Rao, Rajiv Mathews, Swaroop Ramaswamy, Fran{\c{c}}oise
  Beaufays, Sean Augenstein, Hubert Eichner, Chlo{\'e} Kiddon, and Daniel
  Ramage.
\newblock Federated learning for mobile keyboard prediction.
\newblock \emph{arXiv preprint arXiv:1811.03604}, 2018.

\bibitem[Hardt et~al.(2016)Hardt, Price, and Srebro]{hardt2016equality}
Moritz Hardt, Eric Price, and Nati Srebro.
\newblock Equality of opportunity in supervised learning.
\newblock \emph{Advances in neural information processing systems},
  29:\penalty0 3315--3323, 2016.

\bibitem[Hardy et~al.(2017)Hardy, Henecka, Ivey-Law, Nock, Patrini, Smith, and
  Thorne]{hardy2017private}
Stephen Hardy, Wilko Henecka, Hamish Ivey-Law, Richard Nock, Giorgio Patrini,
  Guillaume Smith, and Brian Thorne.
\newblock Private federated learning on vertically partitioned data via entity
  resolution and additively homomorphic encryption.
\newblock \emph{arXiv preprint arXiv:1711.10677}, 2017.

\bibitem[Hartmann et~al.(2019)Hartmann, Suh, Komarzewski, Smith, and
  Segall]{firefox}
Florian Hartmann, Sunah Suh, Arkadiusz Komarzewski, Tim~D. Smith, and Ilana
  Segall.
\newblock Federated learning for ranking browser history suggestions.
\newblock In \emph{arxiv.org/abs/1911.11807}, 2019.

\bibitem[Hastie et~al.(2015)Hastie, Tibshirani, and
  Wainwright]{HastieTibshiraniWainwrightBook}
T.~Hastie, R.~Tibshirani, and M.~Wainwright.
\newblock \emph{Statistical Learning with Sparsity: The Lasso and
  Generalizations}.
\newblock Monographs on Statistics and Applied Probability 143. CRC Press, New
  York, 2015.

\bibitem[Horn and Kr{\"u}ger(2016)]{horn2016feasibility}
Christian Horn and J{\"o}rg Kr{\"u}ger.
\newblock Feasibility of connecting machinery and robots to industrial control
  services in the cloud.
\newblock In \emph{2016 IEEE 21st International Conference on Emerging
  Technologies and Factory Automation (ETFA)}, pages 1--4. IEEE, 2016.

\bibitem[Hospedales et~al.(2020)Hospedales, Antoniou, Micaelli, and
  Storkey]{hospedales2020meta}
Timothy Hospedales, Antreas Antoniou, Paul Micaelli, and Amos Storkey.
\newblock Meta-learning in neural networks: A survey.
\newblock \emph{arXiv preprint arXiv:2004.05439}, 2020.

\bibitem[Hossein~Motlagh et~al.(2020)Hossein~Motlagh, Mohammadrezaei, Hunt, and
  Zakeri]{hossein2020internet}
Naser Hossein~Motlagh, Mahsa Mohammadrezaei, Julian Hunt, and Behnam Zakeri.
\newblock Internet of things (iot) and the energy sector.
\newblock \emph{Energies}, 13\penalty0 (2):\penalty0 494, 2020.

\bibitem[Hosseini et~al.(2021)Hosseini, Carli, and Dotoli]{Hosseini2021}
Seyed~Mohsen Hosseini, Raffaele Carli, and Mariagrazia Dotoli.
\newblock Robust optimal energy management of a residential microgrid under
  uncertainties on demand and renewable power generation.
\newblock \emph{IEEE Transactions on Automation Science and Engineering},
  18\penalty0 (2):\penalty0 618--637, 2021.

\bibitem[Hsieh et~al.(2018)Hsieh, Ananthanarayanan, and et~al.]{focus}
Kevin Hsieh, Ganesh Ananthanarayanan, and et~al.
\newblock {Focus}: Querying large video datasets with low latency and low cost.
\newblock In \emph{USENIX OSDI}, 2018.

\bibitem[Hu et~al.(2020)Hu, Shaloudegi, Zhang, and Yu]{hu2020fedmgda+}
Zeou Hu, Kiarash Shaloudegi, Guojun Zhang, and Yaoliang Yu.
\newblock Fedmgda+: Federated learning meets multi-objective optimization.
\newblock \emph{arXiv preprint arXiv:2006.11489}, 2020.

\bibitem[Huang et~al.(2020{\natexlab{a}})Huang, Yin, Fu, Zhang, Deng, and
  Liu]{huang2020loadaboost}
Li~Huang, Yifeng Yin, Zeng Fu, Shifa Zhang, Hao Deng, and Dianbo Liu.
\newblock Loadaboost: Loss-based adaboost federated machine learning with
  reduced computational complexity on iid and non-iid intensive care data.
\newblock \emph{Plos one}, 15\penalty0 (4):\penalty0 e0230706,
  2020{\natexlab{a}}.

\bibitem[Huang et~al.(2020{\natexlab{b}})Huang, Yin, Fu, Zhang, Deng, and
  Liu]{loadaboost}
Li~Huang, Yifeng Yin, Zeng Fu, Shifa Zhang, Hao Deng, and Dianbo Liu.
\newblock Loadaboost: loss-based adaboost federated machine learning with
  reduced computational complexity on iid and non-iid intensive care data.
\newblock \emph{arXiv preprint arXiv:1811.12629}, 2020{\natexlab{b}}.

\bibitem[Huang et~al.(2020{\natexlab{c}})Huang, Li, Wang, Du, and
  Zhang]{huang2020fairness}
Wei Huang, Tianrui Li, Dexian Wang, Shengdong Du, and Junbo Zhang.
\newblock Fairness and accuracy in federated learning.
\newblock \emph{arXiv preprint arXiv:2012.10069}, 2020{\natexlab{c}}.

\bibitem[Hung et~al.(2015)Hung, Joseph, and Melkote]{hung2015analysis}
Ying Hung, V~Roshan Joseph, and Shreyes~N Melkote.
\newblock Analysis of computer experiments with functional response.
\newblock \emph{Technometrics}, 57\penalty0 (1):\penalty0 35--44, 2015.

\bibitem[Huo et~al.(2018)Huo, Gu, and Huang]{huo2018training}
Zhouyuan Huo, Bin Gu, and Heng Huang.
\newblock Training neural networks using features replay.
\newblock In \emph{Advances in Neural Information Processing Systems}, pages
  6659--6668, 2018.

\bibitem[Imani et~al.(2019)Imani, Ghoreishi, Allaire, and
  Braga-Neto]{imani2019mfbo}
Mahdi Imani, Seyede~Fatemeh Ghoreishi, Douglas Allaire, and Ulisses~M
  Braga-Neto.
\newblock Mfbo-ssm: Multi-fidelity bayesian optimization for fast inference in
  state-space models.
\newblock In \emph{Proceedings of the AAAI Conference on Artificial
  Intelligence}, volume~33, pages 7858--7865, 2019.

\bibitem[Iyengar(2020)]{fillgap}
Rishi Iyengar.
\newblock Can 3d printing plug the coronavirus equipment gap?
\newblock
  \url{https://www.cnn.com/2020/04/16/tech/coronavirus-medical-equipment-3d-printing/index.html},
  2020.
\newblock Accessed: 2020-07-18.

\bibitem[Izmailov et~al.(2018)Izmailov, Podoprikhin, Garipov, Vetrov, and
  Wilson]{izmailov2018averaging}
Pavel Izmailov, Dmitrii Podoprikhin, Timur Garipov, Dmitry Vetrov, and
  Andrew~Gordon Wilson.
\newblock Averaging weights leads to wider optima and better generalization.
\newblock \emph{uncertainty in artificial intelligence}, 2018.

\bibitem[Jahn et~al.(2019)Jahn, Gallus, Nguyen, Pan, Cetin, Byon, Manuel, Zhou,
  and Jahani]{jahn2019projecting}
David~E Jahn, William~A Gallus, Phong~TT Nguyen, Qiyun Pan, Kristen Cetin,
  Eunshin Byon, Lance Manuel, Yuyu Zhou, and Elham Jahani.
\newblock Projecting the most likely annual urban heat extremes in the central
  united states.
\newblock \emph{Atmosphere}, 10\penalty0 (12):\penalty0 727, 2019.

\bibitem[Jang and Byon(2020)]{jang2020probabilistic}
Youngchan Jang and Eunshin Byon.
\newblock Probabilistic characterization of wind diurnal variability for wind
  resource assessment.
\newblock \emph{IEEE Transactions on Sustainable Energy}, 11\penalty0
  (4):\penalty0 2535--2544, 2020.

\bibitem[Jang et~al.(2020)Jang, Byon, Jahani, and Cetin]{jang2020long}
Youngchan Jang, Eunshin Byon, Elham Jahani, and Kristen Cetin.
\newblock On the long-term density prediction of peak electricity load with
  demand side management in buildings.
\newblock \emph{Energy and Buildings}, 228:\penalty0 110450, 2020.

\bibitem[Jay et~al.(2019)Jay, Rotman, Godfrey, Schapira, and Tamar]{drl-cc}
Nathan Jay, Noga Rotman, P~Brighten Godfrey, Michael Schapira, and Aviv Tamar.
\newblock A deep reinforcement learning perspective on {Internet} congestion
  control.
\newblock In \emph{International Conference on Machine Learning}, 2019.

\bibitem[Jiang et~al.(2018)Jiang, Ananthanarayanan, Bodik, Sen, and
  Stoica]{chameleon}
Junchen Jiang, Ganesh Ananthanarayanan, Peter Bodik, Siddhartha Sen, and Ion
  Stoica.
\newblock Chameleon: scalable adaptation of video analytics.
\newblock In \emph{ACM SIGCOMM}, pages 253--266, 2018.

\bibitem[Jiang et~al.(2019{\natexlab{a}})Jiang, Zhou, Ananthanarayanan, Shu,
  and Chien]{camera-cluster}
Junchen Jiang, Yuhao Zhou, Ganesh Ananthanarayanan, Yuanchao Shu, and Andrew~A.
  Chien.
\newblock Networked cameras are the new big data clusters.
\newblock In \emph{HotEdgeVideo}, 2019{\natexlab{a}}.

\bibitem[Jiang et~al.(2019{\natexlab{b}})Jiang, Kone{\v{c}}n{\`y}, Rush, and
  Kannan]{jiang2019improving}
Yihan Jiang, Jakub Kone{\v{c}}n{\`y}, Keith Rush, and Sreeram Kannan.
\newblock Improving federated learning personalization via model agnostic meta
  learning.
\newblock \emph{arXiv preprint arXiv:1909.12488}, 2019{\natexlab{b}}.

\bibitem[Joseph et~al.(2019)Joseph, Gu, Ba, and Myers]{joseph2019space}
V~Roshan Joseph, Li~Gu, Shan Ba, and William~R Myers.
\newblock Space-filling designs for robustness experiments.
\newblock \emph{Technometrics}, 61\penalty0 (1):\penalty0 24--37, 2019.

\bibitem[Kairouz et~al.(2019)Kairouz, McMahan, Avent, Bellet, Bennis, Bhagoji,
  Bonawitz, Charles, Cormode, Cummings, et~al.]{kairouz2019advances}
Peter Kairouz, H~Brendan McMahan, Brendan Avent, Aur{\'e}lien Bellet, Mehdi
  Bennis, Arjun~Nitin Bhagoji, Keith Bonawitz, Zachary Charles, Graham Cormode,
  Rachel Cummings, et~al.
\newblock Advances and open problems in federated learning.
\newblock \emph{arXiv preprint arXiv:1912.04977}, 2019.

\bibitem[Kang et~al.(2019)Kang, Xiong, Niyato, Xie, and
  Zhang]{kang2019incentive}
Jiawen Kang, Zehui Xiong, Dusit Niyato, Shengli Xie, and Junshan Zhang.
\newblock Incentive mechanism for reliable federated learning: A joint
  optimization approach to combining reputation and contract theory.
\newblock \emph{IEEE Internet of Things Journal}, 6\penalty0 (6):\penalty0
  10700--10714, 2019.

\bibitem[Kang et~al.(2011{\natexlab{a}})Kang, Roshan~Joseph, and
  Brenneman]{kang2011design}
Lulu Kang, V~Roshan~Joseph, and William~A Brenneman.
\newblock Design and modeling strategies for mixture-of-mixtures experiments.
\newblock \emph{Technometrics}, 53\penalty0 (2):\penalty0 125--136,
  2011{\natexlab{a}}.

\bibitem[Kang et~al.(2011{\natexlab{b}})Kang, Grauman, and
  Sha]{kang2011learning}
Zhuoliang Kang, Kristen Grauman, and Fei Sha.
\newblock Learning with whom to share in multi-task feature learning.
\newblock In \emph{ICML}, 2011{\natexlab{b}}.

\bibitem[Karimireddy et~al.(2020)Karimireddy, Kale, Mohri, Reddi, Stich, and
  Suresh]{karimireddy2020scaffold}
Sai~Praneeth Karimireddy, Satyen Kale, Mehryar Mohri, Sashank Reddi, Sebastian
  Stich, and Ananda~Theertha Suresh.
\newblock Scaffold: Stochastic controlled averaging for federated learning.
\newblock In \emph{International Conference on Machine Learning}, pages
  5132--5143. PMLR, 2020.

\bibitem[Katharopoulos and Fleuret(2018)]{nonconvexadaptivesampling}
Angelos Katharopoulos and Franc¸ois Fleuret.
\newblock Not all samples are created equal: Deep learning with importance
  sampling.
\newblock \emph{Proceedings of the 35 th International Conference on Machine
  Learning}, \penalty0 (80), 2018.

\bibitem[Keane and Topol(2021)]{keane2021ai}
Pearse~A Keane and Eric~J Topol.
\newblock Ai-facilitated health care requires education of clinicians.
\newblock \emph{The Lancet}, 397\penalty0 (10281):\penalty0 1254, 2021.

\bibitem[Kermarrec and Ta{\"\i}ani(2015)]{kermarrec2015want}
Anne-Marie Kermarrec and Fran{\c{c}}ois Ta{\"\i}ani.
\newblock Want to scale in centralized systems? think p2p.
\newblock \emph{Journal of Internet Services and Applications}, 6\penalty0
  (1):\penalty0 1--12, 2015.

\bibitem[Kim et~al.(2019)Kim, Mnih, Schwarz, Garnelo, Eslami, Rosenbaum,
  Vinyals, and Teh]{kim2019attentive}
Hyunjik Kim, Andriy Mnih, Jonathan Schwarz, Marta Garnelo, Ali Eslami, Dan
  Rosenbaum, Oriol Vinyals, and Yee~Whye Teh.
\newblock Attentive neural processes.
\newblock \emph{International Conference on Learning Representations}, 2019.

\bibitem[Kimberly(2019)]{nvidiaclara}
Powell Kimberly.
\newblock Nvidia clara federated learning to deliver ai to hospitals while
  protecting patient data.
\newblock
  \url{https://blogs.nvidia.com/blog/2019/12/01/clara-federated-learning/},
  2019.
\newblock Accessed: 2021-04-21.

\bibitem[Kingma and Ba(2015)]{kingma2014adam}
Diederik~P Kingma and Jimmy Ba.
\newblock Adam: A method for stochastic optimization.
\newblock \emph{International Conference on Learning Representations}, 2015.

\bibitem[Kirkpatrick et~al.(2017)Kirkpatrick, Pascanu, Rabinowitz, Veness,
  Desjardins, Rusu, Milan, Quan, Ramalho, Grabska-Barwinska,
  et~al.]{kirkpatrick2017overcoming}
James Kirkpatrick, Razvan Pascanu, Neil Rabinowitz, Joel Veness, Guillaume
  Desjardins, Andrei~A Rusu, Kieran Milan, John Quan, Tiago Ramalho, Agnieszka
  Grabska-Barwinska, et~al.
\newblock Overcoming catastrophic forgetting in neural networks.
\newblock \emph{Proceedings of the national academy of sciences}, 114\penalty0
  (13):\penalty0 3521--3526, 2017.

\bibitem[Kloim{\"u}llner and Raidl(2017)]{decomposition_3}
Christian Kloim{\"u}llner and G{\"u}nther~R Raidl.
\newblock Full-load route planning for balancing bike sharing systems by
  logic-based benders decomposition.
\newblock \emph{Networks}, 69\penalty0 (3):\penalty0 270--289, 2017.

\bibitem[Koller and Friedman(2009)]{Koller2009}
Daphne Koller and Nir Friedman.
\newblock \emph{Probabilistic graphical models: principles and techniques}.
\newblock MIT press, 2009.

\bibitem[Koloskova et~al.(2019)Koloskova, Lin, Stich, and
  Jaggi]{koloskova2019decentralized}
Anastasia Koloskova, Tao Lin, Sebastian~U Stich, and Martin Jaggi.
\newblock Decentralized deep learning with arbitrary communication compression.
\newblock \emph{arXiv preprint arXiv:1907.09356}, 2019.

\bibitem[Koltchinskii and Yuan(2010)]{KolYua10Journal}
V.~Koltchinskii and M.~Yuan.
\newblock Sparsity in multiple kernel learning.
\newblock \emph{Annals of Statistics}, 38:\penalty0 3660--3695, 2010.

\bibitem[Konecny et~al.(2016)Konecny, McMahan, Ramage, and Richtarik]{fedsvrg}
Jakub Konecny, H.~Brendan McMahan, Daniel Ramage, and Peter Richtarik.
\newblock Federated optimization: Distributed machine learning for on-device
  intelligence.
\newblock \emph{arXiv preprint arXiv:1610.02527}, 2016.

\bibitem[Kone{\v{c}}n{\`y} et~al.(2016)Kone{\v{c}}n{\`y}, McMahan, Yu,
  Richt{\'a}rik, Suresh, and Bacon]{konevcny2016federated}
Jakub Kone{\v{c}}n{\`y}, H~Brendan McMahan, Felix~X Yu, Peter Richt{\'a}rik,
  Ananda~Theertha Suresh, and Dave Bacon.
\newblock Federated learning: Strategies for improving communication
  efficiency.
\newblock \emph{arXiv preprint arXiv:1610.05492}, 2016.

\bibitem[Kontar et~al.()Kontar, Raskutti, and Zhou]{KontarRaskutti}
R.~Kontar, G.~Raskutti, and S.~Zhou.
\newblock Minimizing negative transfer of knowledge in multivariate gaussian
  processes: A scalable and regularized approach.
\newblock \emph{IEEE Transactions of Pattern Analysis and Machine
  Intelligence}.
\newblock To appear.

\bibitem[Kontar et~al.(2017)Kontar, Zhou, Sankavaram, Du, and
  Zhang]{kontar2017nonparametric}
Raed Kontar, Shiyu Zhou, Chaitanya Sankavaram, Xinyu Du, and Yilu Zhang.
\newblock Nonparametric-condition-based remaining useful life prediction
  incorporating external factors.
\newblock \emph{IEEE Transactions on Reliability}, 67\penalty0 (1):\penalty0
  41--52, 2017.

\bibitem[Kontar et~al.(2018)Kontar, Zhou, Sankavaram, Du, and
  Zhang]{kontar2018nonparametric}
Raed Kontar, Shiyu Zhou, Chaitanya Sankavaram, Xinyu Du, and Yilu Zhang.
\newblock Nonparametric modeling and prognosis of condition monitoring signals
  using multivariate gaussian convolution processes.
\newblock \emph{Technometrics}, 60\penalty0 (4):\penalty0 484--496, 2018.

\bibitem[Kontar et~al.(2020)Kontar, Raskutti, and Zhou]{kontar2020minimizing}
Raed Kontar, Garvesh Raskutti, and Shiyu Zhou.
\newblock Minimizing negative transfer of knowledge in multivariate gaussian
  processes: A scalable and regularized approach.
\newblock \emph{IEEE Transactions on Pattern Analysis and Machine
  Intelligence}, 2020.

\bibitem[Kraska et~al.(2018)Kraska, Beutel, Chi, Dean, and
  Polyzotis]{learned-index}
Tim Kraska, Alex Beutel, Ed~H Chi, Jeffrey Dean, and Neoklis Polyzotis.
\newblock The case for learned index structures.
\newblock In \emph{ACM SIGMOD}, 2018.

\bibitem[{Kulkarni} et~al.(2020){Kulkarni}, {Kulkarni}, and
  {Pant}]{surveyonpersonalization}
V.~{Kulkarni}, M.~{Kulkarni}, and A.~{Pant}.
\newblock Survey of personalization techniques for federated learning.
\newblock In \emph{2020 Fourth World Conference on Smart Trends in Systems,
  Security and Sustainability (WorldS4)}, pages 794--797, 2020.
\newblock \doi{10.1109/WorldS450073.2020.9210355}.

\bibitem[Kumar and Daume~III(2012)]{kumar2012learning}
Abhishek Kumar and Hal Daume~III.
\newblock Learning task grouping and overlap in multi-task learning.
\newblock \emph{arXiv preprint arXiv:1206.6417}, 2012.

\bibitem[Laaper et~al.(2020)Laaper, Dollar, Cotteleer, and
  Sniderman]{laaper2020deloitte}
S~Laaper, B~Dollar, M~Cotteleer, and B~Sniderman.
\newblock Implementing the smart factory: New perspectives for driving value.
\newblock \emph{Deloitte Insights, Deloitte, USA}, 2020.

\bibitem[Lacoste et~al.(2017)Lacoste, Boquet, Rostamzadeh, Oreshkin, Chung, and
  Krueger]{lacoste2017deep}
Alexandre Lacoste, Thomas Boquet, Negar Rostamzadeh, Boris Oreshkin, Wonchang
  Chung, and David Krueger.
\newblock Deep prior.
\newblock \emph{arXiv preprint arXiv:1712.05016}, 2017.

\bibitem[Lacoste et~al.(2018)Lacoste, Oreshkin, Chung, Boquet, Rostamzadeh, and
  Krueger]{lacoste2018uncertainty}
Alexandre Lacoste, Boris Oreshkin, Wonchang Chung, Thomas Boquet, Negar
  Rostamzadeh, and David Krueger.
\newblock Uncertainty in multitask transfer learning.
\newblock \emph{arXiv preprint arXiv:1806.07528}, 2018.

\bibitem[Laffont and Martimort(2009)]{laffont2009theory}
Jean-Jacques Laffont and David Martimort.
\newblock \emph{The theory of incentives: the principal-agent model}.
\newblock Princeton University press, Princeton, NJ, 2009.

\bibitem[Lai et~al.(2020{\natexlab{a}})Lai, You, Zhu, Madhyastha, and
  Chowdhury]{sol}
Fan Lai, Jie You, Xiangfeng Zhu, Harsha~V. Madhyastha, and Mosharaf Chowdhury.
\newblock Sol: A federated execution engine for fast distributed computation
  over slow networks.
\newblock In \emph{USENIX NSDI}, 2020{\natexlab{a}}.

\bibitem[Lai et~al.(2020{\natexlab{b}})Lai, Zhu, Madhyastha, and
  Chowdhury]{oort}
Fan Lai, Xiangfeng Zhu, Harsha~V. Madhyastha, and Mosharaf Chowdhury.
\newblock Oort: Informed participant selection for scalable federated learning.
\newblock \emph{arXiv preprint arXiv:2010.06081}, 2020{\natexlab{b}}.

\bibitem[Larson et~al.(2014)Larson, Liang, and
  Johansson]{larson2014distributed}
Jeffrey Larson, Kuo-Yun Liang, and Karl~H Johansson.
\newblock A distributed framework for coordinated heavy-duty vehicle
  platooning.
\newblock \emph{IEEE Transactions on Intelligent Transportation Systems},
  16\penalty0 (1):\penalty0 419--429, 2014.

\bibitem[Leonard(2019{\natexlab{a}})]{predictivemaintanence}
Matt Leonard.
\newblock With predictive maintenance, operators seek improved uptime.
\newblock
  \url{https://www.supplychaindive.com/news/with-predictive-maintenance-operators-seek-improved-uptime/561684/},
  2019{\natexlab{a}}.
\newblock Accessed: 2020-07-18.

\bibitem[Leonard(2019{\natexlab{b}})]{sensorprice}
Matt Leonard.
\newblock Declining price of iot sensors means greater use in manufacturing.
\newblock
  \url{https://www.supplychaindive.com/news/declining-price-iot-sensors-manufacturing/564980/},
  2019{\natexlab{b}}.
\newblock Accessed: 2020-07-18.

\bibitem[Leurent and De~Boer(2018)]{leurent2018next}
H~Leurent and E~De~Boer.
\newblock The next economic growth engine: Scaling fourth industrial revolution
  technologies in production.
\newblock In \emph{World Economic Forum}, 2018.

\bibitem[Li et~al.(2020{\natexlab{a}})Li, Menassa, Kamat, and Byon]{LI2020}
Da~Li, Carol~C. Menassa, Vineet~R. Kamat, and Eunshin Byon.
\newblock Heat - human embodied autonomous thermostat.
\newblock \emph{Building and Environment}, 178:\penalty0 106879,
  2020{\natexlab{a}}.

\bibitem[Li et~al.(2019{\natexlab{a}})Li, Qin, Jiao, Yang, Wang, Wang, Wu, and
  Ye]{Ridesharing_deep}
Minne Li, Zhiwei Qin, Yan Jiao, Yaodong Yang, Jun Wang, Chenxi Wang, Guobin Wu,
  and Jieping Ye.
\newblock Efficient ridesharing order dispatching with mean field multi-agent
  reinforcement learning.
\newblock In \emph{The World Wide Web Conference}, pages 983--994,
  2019{\natexlab{a}}.

\bibitem[Li and Kontar(2020)]{li2020negative}
Moyan Li and Raed Kontar.
\newblock On negative transfer and structure of latent functions in
  multi-output gaussian processes.
\newblock \emph{arXiv preprint arXiv:2004.02382}, 2020.

\bibitem[Li et~al.(2018)Li, Sahu, Zaheer, Sanjabi, and
  Ameet~Talwalkar]{fedprox}
Tian Li, Anit~Kumar Sahu, Manzil Zaheer, Maziar Sanjabi, and Virginia~Smith
  Ameet~Talwalkar.
\newblock Federated optimization in heterogeneous networks.
\newblock \emph{Proceedings of the 3rd MLSys Conference}, 2018.

\bibitem[Li et~al.(2019{\natexlab{b}})Li, Sahu, Zaheer, Sanjabi, Talwalkar, and
  Smithy]{li2019feddane}
Tian Li, Anit~Kumar Sahu, Manzil Zaheer, Maziar Sanjabi, Ameet Talwalkar, and
  Virginia Smithy.
\newblock Feddane: A federated newton-type method.
\newblock In \emph{2019 53rd Asilomar Conference on Signals, Systems, and
  Computers}, pages 1227--1231. IEEE, 2019{\natexlab{b}}.

\bibitem[Li et~al.(2019{\natexlab{c}})Li, Sanjabi, Beirami, and
  Smith]{li2019fair}
Tian Li, Maziar Sanjabi, Ahmad Beirami, and Virginia Smith.
\newblock Fair resource allocation in federated learning.
\newblock \emph{arXiv preprint arXiv:1905.10497}, 2019{\natexlab{c}}.

\bibitem[Li et~al.(2020{\natexlab{b}})Li, Sahu, Talwalkar, and
  Smith]{li2020federated}
Tian Li, Anit~Kumar Sahu, Ameet Talwalkar, and Virginia Smith.
\newblock Federated learning: Challenges, methods, and future directions.
\newblock \emph{IEEE Signal Processing Magazine}, 37\penalty0 (3):\penalty0
  50--60, 2020{\natexlab{b}}.

\bibitem[Li et~al.(2021)Li, Hu, Beirami, and Smith]{ditto}
Tian Li, Shengyuan Hu, Ahmad Beirami, and Virginia Smith.
\newblock Ditto: Fair and robust federated learning through personalization.
\newblock \emph{arXiv preprint arXiv:2012.04221}, 2021.

\bibitem[Li et~al.(2019{\natexlab{d}})Li, Milletari, and Xu]{nvidia-fl}
Wenqi Li, Fausto Milletari, and Daguang Xu.
\newblock Privacy-preserving federated brain tumour segmentation.
\newblock In \emph{Machine Learning in Medical Imaging}, 2019{\natexlab{d}}.

\bibitem[Li et~al.(2017)Li, Zhou, Chen, and Li]{li2017meta}
Zhenguo Li, Fengwei Zhou, Fei Chen, and Hang Li.
\newblock Meta-sgd: Learning to learn quickly for few-shot learning.
\newblock \emph{arXiv preprint arXiv:1707.09835}, 2017.

\bibitem[Lian et~al.(2017)Lian, Zhang, Zhang, Hsieh, Zhang, and
  Liu]{lian2017can}
Xiangru Lian, Ce~Zhang, Huan Zhang, Cho-Jui Hsieh, Wei Zhang, and Ji~Liu.
\newblock Can decentralized algorithms outperform centralized algorithms? a
  case study for decentralized parallel stochastic gradient descent.
\newblock \emph{arXiv preprint arXiv:1705.09056}, 2017.

\bibitem[Liang et~al.(2020)Liang, Liu, Ziyin, Salakhutdinov, and
  Morency]{lg-fedavg}
Paul~Pu Liang, Terrance Liu, Liu Ziyin, Ruslan Salakhutdinov, and
  Louis-Philippe Morency.
\newblock Think locally, act globally: Federated learning with local and global
  representations.
\newblock \emph{arXiv preprint arXiv:2001.01523}, 2020.

\bibitem[Lim et~al.(2020)Lim, Luong, Hoang, Jiao, Liang, Yang, Niyato, and
  Miao]{lim2020federated}
Wei Yang~Bryan Lim, Nguyen~Cong Luong, Dinh~Thai Hoang, Yutao Jiao, Ying-Chang
  Liang, Qiang Yang, Dusit Niyato, and Chunyan Miao.
\newblock Federated learning in mobile edge networks: A comprehensive survey.
\newblock \emph{IEEE Communications Surveys \& Tutorials}, 22\penalty0
  (3):\penalty0 2031--2063, 2020.

\bibitem[Lin et~al.(2020)Lin, Yang, and Zhang]{lin2020real}
Sen Lin, Guang Yang, and Junshan Zhang.
\newblock Real-time edge intelligence in the making: A collaborative learning
  framework via federated meta-learning.
\newblock \emph{arXiv preprint arXiv:2001.03229}, 2020.

\bibitem[Liu et~al.(2009)Liu, Lafferty, and Wasserman]{LiuLafWas09}
H.~Liu, J.~Lafferty, and L.~Wasserman.
\newblock The nonparanormal: Semiparametric estimation of high-dimensional
  undirected graphs.
\newblock \emph{Journal of Machine Learning Research}, 10:\penalty0 1--37,
  2009.

\bibitem[Liu and Jin(2013)]{liu2013diagnosing}
Jian Liu and Jionghua Jin.
\newblock Diagnosing multistage manufacturing processes with engineering-driven
  factor analysis considering sampling uncertainty.
\newblock \emph{Journal of Manufacturing Science and Engineering}, 135\penalty0
  (4), 2013.

\bibitem[Liu et~al.(2013)Liu, Zhang, and Shi]{liu2013adaptive}
Kaibo Liu, Xi~Zhang, and Jianjun Shi.
\newblock Adaptive sensor allocation strategy for process monitoring and
  diagnosis in a bayesian network.
\newblock \emph{IEEE Transactions on Automation Science and Engineering},
  11\penalty0 (2):\penalty0 452--462, 2013.

\bibitem[Liu and Wang(2016)]{liu2016stein}
Qiang Liu and Dilin Wang.
\newblock Stein variational gradient descent: A general purpose bayesian
  inference algorithm.
\newblock \emph{Advances In Neural Information Processing Systems}, 2016.

\bibitem[Liu et~al.(2020{\natexlab{a}})Liu, Wu, and Mozafari]{adambs}
Rui Liu, Tianyi Wu, and Barzan Mozafari.
\newblock Adam with bandit sampling for deep learning.
\newblock \emph{NeurIPS}, 2020{\natexlab{a}}.

\bibitem[Liu et~al.(2020{\natexlab{b}})Liu, Chen, Chen, and
  Zhang]{liu2020accelerating}
Wei Liu, Li~Chen, Yunfei Chen, and Wenyi Zhang.
\newblock Accelerating federated learning via momentum gradient descent.
\newblock \emph{IEEE Transactions on Parallel and Distributed Systems},
  31\penalty0 (8):\penalty0 1754--1766, 2020{\natexlab{b}}.

\bibitem[Liu et~al.(2017)Liu, Ma, Aafer, Lee, Zhai, Wang, and
  Zhang]{liu2017trojaning}
Yingqi Liu, Shiqing Ma, Yousra Aafer, Wen-Chuan Lee, Juan Zhai, Weihang Wang,
  and Xiangyu Zhang.
\newblock Trojaning attack on neural networks.
\newblock 2017.

\bibitem[Liu et~al.(2019)Liu, Zhang, Zhuo, and Wang]{LIU2019}
Zifa Liu, Zhe Zhang, Ranqun Zhuo, and Xuyang Wang.
\newblock Optimal operation of independent regional power grid with multiple
  wind-solar-hydro-battery power.
\newblock \emph{Applied Energy}, 235:\penalty0 1541--1550, 2019.
\newblock ISSN 0306-2619.
\newblock \doi{https://doi.org/10.1016/j.apenergy.2018.11.072}.
\newblock URL
  \url{https://www.sciencedirect.com/science/article/pii/S0306261918317781}.

\bibitem[Louizos et~al.(2019)Louizos, Shi, Schutte, and
  Welling]{louizos2019functional}
Christos Louizos, Xiahan Shi, Klamer Schutte, and Max Welling.
\newblock The functional neural process.
\newblock \emph{Advances in Neural Information Processing Systems}, 2019.

\bibitem[Lu et~al.(2018)Lu, Jiang, and Kot]{finetuneface}
Ze~Lu, Xudong Jiang, and Alex Kot.
\newblock Deep coupled resnet for low-resolution face recognition.
\newblock \emph{IEEE Signal Processing Letters}, 25\penalty0 (4):\penalty0
  526--530, 2018.
\newblock \doi{10.1109/LSP.2018.2810121}.

\bibitem[Luo et~al.(2019)Luo, Wu, Luo, Huang, Huang, Liu, and Yang]{fed-oj}
Jiahuan Luo, Xueyang Wu, Yun Luo, Anbu Huang, Yunfeng Huang, Yang Liu, and
  Qiang Yang.
\newblock Real-world image datasets for federated learning.
\newblock In \emph{arxiv.org/abs/1910.11089}, 2019.

\bibitem[Luo and Saigal(2021)]{qisaigal2021multiagent}
Qi~Luo and Romesh Saigal.
\newblock Dynamic multiagent incentive contracts - existence, uniqueness and
  implementation.
\newblock \emph{Mathematics}, 9\penalty0 (1), 2021.
\newblock \doi{10.3390/math9010019}.

\bibitem[Lyu et~al.(2020)Lyu, Yu, and Yang]{lyu2020threats}
Lingjuan Lyu, Han Yu, and Qiang Yang.
\newblock Threats to federated learning: A survey.
\newblock \emph{arXiv preprint arXiv:2003.02133}, 2020.

\bibitem[Ma et~al.(2019)Ma, Li, and Hern{\'a}ndez-Lobato]{ma2019variational}
Chao Ma, Yingzhen Li, and Jos{\'e}~Miguel Hern{\'a}ndez-Lobato.
\newblock Variational implicit processes.
\newblock In \emph{International Conference on Machine Learning}, pages
  4222--4233. PMLR, 2019.

\bibitem[Madakam et~al.(2015)Madakam, Lake, Lake, Lake,
  et~al.]{madakam2015internet}
Somayya Madakam, Vihar Lake, Vihar Lake, Vihar Lake, et~al.
\newblock Internet of things (iot): A literature review.
\newblock \emph{Journal of Computer and Communications}, 3\penalty0
  (05):\penalty0 164, 2015.

\bibitem[Maddox et~al.(2019)Maddox, Izmailov, Garipov, Vetrov, and
  Wilson]{maddox2019simple}
Wesley~J Maddox, Pavel Izmailov, Timur Garipov, Dmitry~P Vetrov, and
  Andrew~Gordon Wilson.
\newblock A simple baseline for bayesian uncertainty in deep learning.
\newblock \emph{Advances in Neural Information Processing Systems},
  32:\penalty0 13153--13164, 2019.

\bibitem[Madhyastha and Okwudire(2020)]{madhyastha2020remotely}
Harsha~V Madhyastha and Chinedum Okwudire.
\newblock Remotely controlled manufacturing: A new frontier for systems
  research.
\newblock In \emph{Proceedings of the 21st International Workshop on Mobile
  Computing Systems and Applications}, pages 62--67, 2020.

\bibitem[Madry et~al.(2017)Madry, Makelov, Schmidt, Tsipras, and
  Vladu]{madry2017towards}
Aleksander Madry, Aleksandar Makelov, Ludwig Schmidt, Dimitris Tsipras, and
  Adrian Vladu.
\newblock Towards deep learning models resistant to adversarial attacks.
\newblock \emph{arXiv preprint arXiv:1706.06083}, 2017.

\bibitem[Mahar et~al.(2011)Mahar, Fields, Reade, Zarubin, and
  McCombie]{nprd2011}
David Mahar, William Fields, John Reade, Peter Zarubin, and Scott McCombie.
\newblock \emph{Nonelectronic Parts Reliability Data 2011}.
\newblock Reliability Information Analysis Center, Rome, NY, 2011.

\bibitem[Mahoney and Drineas(2009)]{DrinMah09}
M.~W. Mahoney and P.~Drineas.
\newblock {CUR} matrix decompositions for improved data analysis.
\newblock \emph{Proc. Natl. Acad. Sci. USA}, 106\penalty0 (3):\penalty0
  697--702, 2009.

\bibitem[Mai et~al.(2012)Mai, Sandor, Wiser, and Schneider]{mai2012renewable}
Trieu Mai, Debra Sandor, Ryan Wiser, and Thomas Schneider.
\newblock Renewable electricity futures study. executive summary.
\newblock Technical report, National Renewable Energy Lab.(NREL), Golden, CO
  (United States), 2012.

\bibitem[Malings et~al.(2017)Malings, Pozzi, Klima, Berg{\'e}s, Bou-Zeid, and
  Ramamurthy]{malings2017surface}
Carl Malings, Matteo Pozzi, Kelly Klima, Mario Berg{\'e}s, Elie Bou-Zeid, and
  Prathap Ramamurthy.
\newblock Surface heat assessment for developed environments: Probabilistic
  urban temperature modeling.
\newblock \emph{Computers, Environment and Urban Systems}, 66:\penalty0 53--64,
  2017.

\bibitem[Mansour et~al.(2020)Mansour, Mohri, Ro, and
  Suresh]{threepersonalizationapp}
Yishay Mansour, Mehryar Mohri, Jae Ro, and Ananda~Theertha Suresh.
\newblock Three approaches for personalization with applications to federated
  learning.
\newblock \emph{CoRR}, abs/2002.10619, 2020.
\newblock URL \url{https://arxiv.org/abs/2002.10619}.

\bibitem[Mao et~al.(2017)Mao, Netravali, and Alizadeh]{pensieve}
Hongzi Mao, Ravi Netravali, and Mohammad Alizadeh.
\newblock Neural adaptive video streaming with {Pensieve}.
\newblock In \emph{ACM SIGCOMM}, 2017.

\bibitem[Masoud and Jayakrishnan(2017)]{decomposition_1}
Neda Masoud and R~Jayakrishnan.
\newblock A decomposition algorithm to solve the multi-hop peer-to-peer
  ride-matching problem.
\newblock \emph{Transportation Research Part B: Methodological}, 99:\penalty0
  1--29, 2017.

\bibitem[Mcdonald et~al.(2009)Mcdonald, Mohri, Silberman, Walker, and
  Mann]{fedsgd}
Ryan Mcdonald, Mehryar Mohri, Nathan Silberman, Dan Walker, and Gideon Mann.
\newblock Efficient large-scale distributed training of conditional maximum
  entropy models.
\newblock In Y.~Bengio, D.~Schuurmans, J.~Lafferty, C.~Williams, and
  A.~Culotta, editors, \emph{Advances in Neural Information Processing
  Systems}, volume~22. Curran Associates, Inc., 2009.
\newblock URL
  \url{https://proceedings.neurips.cc/paper/2009/file/d81f9c1be2e08964bf9f24b15f0e4900-Paper.pdf}.

\bibitem[McKinsey(2016)]{analytics2016age}
McKinsey.
\newblock The age of analytics: competing in a data-driven world, 2016.

\bibitem[McMahan et~al.(2017{\natexlab{a}})McMahan, Moore, Ramage, Hampson, and
  y~Arcas]{mcmahan2017communication}
Brendan McMahan, Eider Moore, Daniel Ramage, Seth Hampson, and Blaise~Aguera
  y~Arcas.
\newblock Communication-efficient learning of deep networks from decentralized
  data.
\newblock In \emph{Artificial Intelligence and Statistics}, pages 1273--1282,
  2017{\natexlab{a}}.

\bibitem[McMahan et~al.(2017{\natexlab{b}})McMahan, Moore, Ramage, Hampson, and
  y~Arcas]{fedavg}
H.~Brendan McMahan, Eider Moore, Daniel Ramage, Seth Hampson, and
  Blaise~Agüera y~Arcas.
\newblock Communication-efficient learning of deep networks from decentralized
  data.
\newblock In \emph{Proceedings of the 20th International Conference on
  Artificial Intelligence and Statistics}. JMLR, 2017{\natexlab{b}}.

\bibitem[McMahan et~al.(2021)]{mcmahan2021advances}
H~Brendan McMahan et~al.
\newblock Advances and open problems in federated learning.
\newblock \emph{Foundations and Trends{\textregistered} in Machine Learning},
  14\penalty0 (1), 2021.

\bibitem[McPherson and Stoll(2020)]{MCPHERSON2020}
Madeleine McPherson and Brady Stoll.
\newblock Demand response for variable renewable energy integration: A proposed
  approach and its impacts.
\newblock \emph{Energy}, 197:\penalty0 117205, 2020.

\bibitem[Meeker and Escobar(1998)]{meeker1998}
William~Q Meeker and Luis~A Escobar.
\newblock \emph{Statistical Methods for Reliability Data}.
\newblock John Wiley \& Sons, 1998.

\bibitem[Microsoft(2019)]{manufcaturingtrends}
Microsoft.
\newblock 2019 manufacturing trends report.
\newblock
  \url{https://info.microsoft.com/rs/157-GQE-382/images/EN-US-CNTNT-Report-2019-Manufacturing-Trends.pdf},
  2019.
\newblock Accessed: 2020-07-18.

\bibitem[Mohebifard and Hajbabaie(2019)]{Intersection_2}
Rasool Mohebifard and Ali Hajbabaie.
\newblock Optimal network-level traffic signal control: A benders
  decomposition-based solution algorithm.
\newblock \emph{Transportation Research Part B: Methodological}, 121:\penalty0
  252--274, 2019.

\bibitem[Mohri et~al.(2019{\natexlab{a}})Mohri, Sivek, and Suresh]{agnosticfl}
Mehryar Mohri, Gary Sivek, and Ananda~Theertha Suresh.
\newblock Agnostic federated learning.
\newblock In Kamalika Chaudhuri and Ruslan Salakhutdinov, editors,
  \emph{Proceedings of the 36th International Conference on Machine Learning},
  volume~97 of \emph{Proceedings of Machine Learning Research}, pages
  4615--4625. PMLR, 09--15 Jun 2019{\natexlab{a}}.

\bibitem[Mohri et~al.(2019{\natexlab{b}})Mohri, Sivek, and
  Suresh]{mohri2019agnostic}
Mehryar Mohri, Gary Sivek, and Ananda~Theertha Suresh.
\newblock Agnostic federated learning.
\newblock In \emph{International Conference on Machine Learning}, pages
  4615--4625. PMLR, 2019{\natexlab{b}}.

\bibitem[Moore(2017)]{UMtech}
Nicole~Casal Moore.
\newblock 3-d printing gets a turbo boost from u-m technology.
\newblock
  \url{https://news.umich.edu/3-d-printing-gets-a-turbo-boost-from-u-m-technology/},
  November 2017.
\newblock Accessed: 2019-02-19.

\bibitem[Mubeen et~al.(2017)Mubeen, Nikolaidis, Didic, Pei-Breivold,
  Sandstr{\"o}m, and Behnam]{mubeen2017delay}
Saad Mubeen, Pavlos Nikolaidis, Alma Didic, Hongyu Pei-Breivold, Kristian
  Sandstr{\"o}m, and Moris Behnam.
\newblock Delay mitigation in offloaded cloud controllers in industrial iot.
\newblock \emph{IEEE Access}, 5:\penalty0 4418--4430, 2017.

\bibitem[Munkhdalai and Yu(2017)]{munkhdalai2017meta}
Tsendsuren Munkhdalai and Hong Yu.
\newblock Meta networks.
\newblock In \emph{International Conference on Machine Learning}, pages
  2554--2563. PMLR, 2017.

\bibitem[Nagy et~al.(2018)Nagy, Ol{\'a}h, Erdei, M{\'a}t{\'e}, and
  Popp]{nagy2018role}
Judit Nagy, Judit Ol{\'a}h, Edina Erdei, Domici{\'a}n M{\'a}t{\'e}, and
  J{\'o}zsef Popp.
\newblock The role and impact of industry 4.0 and the internet of things on the
  business strategy of the value chain—the case of hungary.
\newblock \emph{Sustainability}, 10\penalty0 (10):\penalty0 3491, 2018.

\bibitem[Namkoong et~al.(2017)Namkoong, Sinha, Yadlowsky, and
  Duchi]{convexadaptivesampling}
Hongseok Namkoong, Aman Sinha, Steve Yadlowsky, and John~C. Duchi.
\newblock Radaptive sampling probabilities for non-smooth optimizatio.
\newblock \emph{Proceedings of the 34 th International Conference on Machine
  Learning}, \penalty0 (70), 2017.

\bibitem[Nan et~al.(2018)Nan, Zhou, and Li]{NAN2018}
Sibo Nan, Ming Zhou, and Gengyin Li.
\newblock Optimal residential community demand response scheduling in smart
  grid.
\newblock \emph{Applied Energy}, 210:\penalty0 1280--1289, 2018.

\bibitem[Neukomm et~al.(2019)Neukomm, Nubbe, and Fares]{neukomm2019grid}
Monica Neukomm, Valerie Nubbe, and Robert Fares.
\newblock Grid-interactive efficient buildings.
\newblock Technical report, US Dept. of Energy (USDOE), Washington DC (United
  States); Navigant~…, 2019.

\bibitem[Nguyen et~al.(2020{\natexlab{a}})Nguyen, Do, and
  Carneiro]{nguyen2020pac}
Cuong Nguyen, Thanh-Toan Do, and Gustavo Carneiro.
\newblock Pac-bayesian meta-learning with implicit prior.
\newblock \emph{IEEE TRANSACTION ON PATTERN ANALYSIS AND MACHINE INTELLIGENCE},
  2020{\natexlab{a}}.

\bibitem[Nguyen et~al.(2018)Nguyen, Kieu, Wen, and Cai]{Transportation_deep}
Hoang Nguyen, Le-Minh Kieu, Tao Wen, and Chen Cai.
\newblock Deep learning methods in transportation domain: a review.
\newblock \emph{IET Intelligent Transport Systems}, 12\penalty0 (9):\penalty0
  998--1004, 2018.

\bibitem[Nguyen et~al.(2020{\natexlab{b}})Nguyen, Luong, Zhao, Yuen, and
  Niyato]{nguyen2020resource}
Huy~T Nguyen, Nguyen~Cong Luong, Jun Zhao, Chau Yuen, and Dusit Niyato.
\newblock Resource allocation in mobility-aware federated learning networks: a
  deep reinforcement learning approach.
\newblock In \emph{2020 IEEE 6th World Forum on Internet of Things (WF-IoT)},
  pages 1--6. IEEE, 2020{\natexlab{b}}.

\bibitem[Nichol and Schulman(2018)]{nichol2018reptile}
Alex Nichol and John Schulman.
\newblock Reptile: a scalable metalearning algorithm.
\newblock \emph{arXiv preprint arXiv:1803.02999}, 2\penalty0 (2):\penalty0 1,
  2018.

\bibitem[Nichol et~al.(2018)Nichol, Achiam, and Schulman]{nichol2018first}
Alex Nichol, Joshua Achiam, and John Schulman.
\newblock On first-order meta-learning algorithms.
\newblock \emph{arXiv preprint arXiv:1803.02999}, 2018.

\bibitem[Niknam et~al.(2020)Niknam, Dhillon, and Reed]{niknam2020federated}
Solmaz Niknam, Harpreet~S Dhillon, and Jeffrey~H Reed.
\newblock Federated learning for wireless communications: Motivation,
  opportunities, and challenges.
\newblock \emph{IEEE Communications Magazine}, 58\penalty0 (6):\penalty0
  46--51, 2020.

\bibitem[Ning et~al.(2017)Ning, Byon, Wu, and Li]{Ning2017}
Shuluo Ning, Eunshin Byon, Teresa Wu, and Jing Li.
\newblock A sparse partitioned-regression model for nonlinear
  system–environment interactions.
\newblock \emph{IISE Transactions}, 49\penalty0 (8):\penalty0 814--826, 2017.

\bibitem[{NOAA}(2021)]{ocean}
{NOAA}.
\newblock Climate change: Global temperature.
\newblock
  \url{https://www.climate.gov/news-features/understanding-climate/climate-change-global-temperature},
  2021.
\newblock Accessed: 2021-04-17.

\bibitem[Nock et~al.(2018)Nock, Hardy, Henecka, Ivey-Law, Patrini, Smith, and
  Thorne]{nock2018entity}
Richard Nock, Stephen Hardy, Wilko Henecka, Hamish Ivey-Law, Giorgio Patrini,
  Guillaume Smith, and Brian Thorne.
\newblock Entity resolution and federated learning get a federated resolution.
\newblock \emph{arXiv preprint arXiv:1803.04035}, 2018.

\bibitem[OCR(2009)]{ocr}
OCR.
\newblock Office for civil rights, research.
\newblock
  \url{https://www.hhs.gov/hipaa/for-professionals/privacy/guidance/research/index.html},
  2009.
\newblock Accessed: 2021-04-25.

\bibitem[Okwudire et~al.(2020)Okwudire, Lu, Kumaravelu, and
  Madhyastha]{okwudire2020three}
CE~Okwudire, Xiang Lu, Giridharan Kumaravelu, and Harsha Madhyastha.
\newblock A three-tier redundant architecture for safe and reliable cloud-based
  cnc over public internet networks.
\newblock \emph{Robotics and Computer-Integrated Manufacturing}, 62:\penalty0
  101880, 2020.

\bibitem[Okwudire and Madhyastha(2021)]{MDM}
Chinedum~E. Okwudire and Harsha~V. Madhyastha.
\newblock Distributed manufacturing for and by the masses.
\newblock In \emph{Science}, 2021.

\bibitem[Okwudire et~al.(2018)Okwudire, Huggi, Supe, Huang, and
  Zeng]{okwudire2018low}
Chinedum~E Okwudire, Sharankumar Huggi, Sagar Supe, Chengyang Huang, and Bowen
  Zeng.
\newblock Low-level control of 3d printers from the cloud: A step toward 3d
  printer control as a service.
\newblock \emph{Inventions}, 3\penalty0 (3):\penalty0 56, 2018.

\bibitem[OnStar(2021)]{onstar}
OnStar.
\newblock Welcome to onstar.
\newblock \url{https://www.onstar.com/}, 2021.
\newblock Accessed: 2020-07-18.

\bibitem[OREDA(2009)]{oreda2009}
OREDA.
\newblock \emph{OREDA Offshore Reliability Data Handbook}.
\newblock Det Norske Veritas (DNV), 2009.

\bibitem[Pallonetto et~al.(2020)Pallonetto, {De Rosa}, D’Ettorre, and
  Finn]{PALLONETTO2020}
Fabiano Pallonetto, Mattia {De Rosa}, Francesco D’Ettorre, and Donal~P. Finn.
\newblock On the assessment and control optimisation of demand response
  programs in residential buildings.
\newblock \emph{Renewable and Sustainable Energy Reviews}, 127:\penalty0
  109861, 2020.

\bibitem[Pan and Yang(2009)]{pan2009survey}
Sinno~Jialin Pan and Qiang Yang.
\newblock A survey on transfer learning.
\newblock \emph{IEEE Transactions on knowledge and data engineering},
  22\penalty0 (10):\penalty0 1345--1359, 2009.

\bibitem[Pan and Yang(2010)]{pan2010survey}
Sinno~Jialin Pan and Qiang Yang.
\newblock A survey on transfer learning ieee transactions on knowledge and data
  engineering.
\newblock \emph{22 (10): 1345}, 1359, 2010.

\bibitem[Park and Raskutti(2015)]{park2015learning}
Gunwoong Park and Garvesh Raskutti.
\newblock Learning large-scale poisson dag models based on overdispersion
  scoring.
\newblock In \emph{Advances in Neural Information Processing Systems}, pages
  631--639, 2015.

\bibitem[Patacchiola et~al.(2019)Patacchiola, Turner, Crowley, O'Boyle, and
  Storkey]{patacchiola2019deep}
Massimiliano Patacchiola, Jack Turner, Elliot~J Crowley, Michael O'Boyle, and
  Amos Storkey.
\newblock Deep kernel transfer in gaussian processes for few-shot learning.
\newblock \emph{arXiv preprint arXiv:1910.05199}, 2019.

\bibitem[Patacchiola et~al.(2020)Patacchiola, Turner, Crowley, O'Boyle, and
  Storkey]{patacchiola2020bayesian}
Massimiliano Patacchiola, Jack Turner, Elliot~J Crowley, Michael O'Boyle, and
  Amos Storkey.
\newblock Bayesian meta-learning for the few-shot setting via deep kernels.
\newblock \emph{Advances in Neural Information Processing Systems}, 2020.

\bibitem[Pathak and Wainwright(2020)]{fedsplit}
Reese Pathak and Martin~J. Wainwright.
\newblock Fedsplit: an algorithmic framework for fast federated optimization.
\newblock In \emph{34th Conference on Neural Information Processing Systems},
  2020.
\newblock URL
  \url{https://proceedings.neurips.cc/paper/2020/file/4ebd440d99504722d80de606ea8507da-Paper.pdf}.

\bibitem[Peaceman and Rachford(1955)]{prsplit}
D.~W. Peaceman and Jr. H.~H. Rachford.
\newblock The numerical solution of parabolic and elliptic differential
  equations.
\newblock \emph{Journal of the SIAM}, 1955.

\bibitem[Pearl(2000)]{Pearl2000}
Judea Pearl.
\newblock \emph{Causality: models, reasoning and inference}, volume~29.
\newblock Cambridge Univ Press, 2000.

\bibitem[Pelzer et~al.(2015)Pelzer, Xiao, Zehe, Lees, Knoll, and
  Aydt]{partitioning_2}
Dominik Pelzer, Jiajian Xiao, Daniel Zehe, Michael~H Lees, Alois~C Knoll, and
  Heiko Aydt.
\newblock A partition-based match making algorithm for dynamic ridesharing.
\newblock \emph{IEEE Transactions on Intelligent Transportation Systems},
  16\penalty0 (5):\penalty0 2587--2598, 2015.

\bibitem[Peng et~al.(2019)Peng, Ye, and Chen]{peng2019bayesian}
Weiwen Peng, Zhi-Sheng Ye, and Nan Chen.
\newblock Bayesian deep-learning-based health prognostics toward prognostics
  uncertainty.
\newblock \emph{IEEE Transactions on Industrial Electronics}, 67\penalty0
  (3):\penalty0 2283--2293, 2019.

\bibitem[Platt et~al.(2018)Platt, Jacobson, and Kardia]{platt2018public}
Jodyn~E Platt, Peter~D Jacobson, and Sharon~LR Kardia.
\newblock Public trust in health information sharing: a measure of system
  trust.
\newblock \emph{Health services research}, 53\penalty0 (2):\penalty0 824--845,
  2018.

\bibitem[Plumlee(2017)]{plumlee2017bayesian}
Matthew Plumlee.
\newblock Bayesian calibration of inexact computer models.
\newblock \emph{Journal of the American Statistical Association}, 112\penalty0
  (519):\penalty0 1274--1285, 2017.

\bibitem[Plumlee(2019)]{plumlee2019computer}
Matthew Plumlee.
\newblock Computer model calibration with confidence and consistency.
\newblock \emph{Journal of the Royal Statistical Society: Series B (Statistical
  Methodology)}, 81\penalty0 (3):\penalty0 519--545, 2019.

\bibitem[Pontile et~al.(2007)Pontile, Evgeniou, and
  Argyriou]{pontile2007convex}
M~Pontile, T~Evgeniou, and A~Argyriou.
\newblock Convex multi-task feature learning.
\newblock \emph{Journal of Machine Learning}, 10:\penalty0 243--272, 2007.

\bibitem[Porter and Heppelmann(2014)]{porter2014smart}
Michael~E Porter and James~E Heppelmann.
\newblock How smart, connected products are transforming competition.
\newblock \emph{Harvard business review}, 92\penalty0 (11):\penalty0 64--88,
  2014.

\bibitem[Pu et~al.(2015)Pu, Ananthanarayanan, Bodik, Kandula, Akella, Bahl, and
  Stoica]{iridium}
Qifan Pu, Ganesh Ananthanarayanan, Peter Bodik, Srikanth Kandula, Aditya
  Akella, Victor Bahl, and Ion Stoica.
\newblock Low latency geo-distributed data analytics.
\newblock In \emph{ACM SIGCOMM}, 2015.

\bibitem[Qiu(2020)]{qiu2020big}
Peihua Qiu.
\newblock Big data? statistical process control can help!
\newblock \emph{The American Statistician}, 74\penalty0 (4):\penalty0 329--344,
  2020.

\bibitem[Radford et~al.(2019)Radford, Wu, Child, Luan, Amodei, and
  Sutskever]{gpt2}
Alec Radford, Jeffrey Wu, Rewon Child, David Luan, Dario Amodei, and Ilya
  Sutskever.
\newblock Language models are unsupervised multitask learners.
\newblock \emph{OpenAI}, 2019.

\bibitem[Rahman et~al.(2020)Rahman, Tout, Ould-Slimane, Mourad, Talhi, and
  Guizani]{rahman2020survey}
Sawsan~Abdul Rahman, Hanine Tout, Hakima Ould-Slimane, Azzam Mourad,
  Chamseddine Talhi, and Mohsen Guizani.
\newblock A survey on federated learning: The journey from centralized to
  distributed on-site learning and beyond.
\newblock \emph{IEEE Internet of Things Journal}, 2020.

\bibitem[Ramaswamy et~al.(2019)Ramaswamy, Mathews, Rao, and
  Beaufays]{ramaswamy2019federated}
Swaroop Ramaswamy, Rajiv Mathews, Kanishka Rao, and Fran{\c{c}}oise Beaufays.
\newblock Federated learning for emoji prediction in a mobile keyboard.
\newblock \emph{arXiv preprint arXiv:1906.04329}, 2019.

\bibitem[Raskutti and Mahoney(2016)]{RaskuttiMahoney}
G.~Raskutti and M.~Mahoney.
\newblock A statistical perspective on randomized sketching for ordinary
  least-squares.
\newblock \emph{Journal of Machine Learning Research}, 17\penalty0
  (213):\penalty0 1--31, 2016.

\bibitem[Raskutti et~al.(2012)Raskutti, Wainwright, and Yu]{RasWaiYu12}
G.~Raskutti, M.~J. Wainwright, and B.~Yu.
\newblock Minimax-optimal rates for sparse additive models over kernel classes
  via convex programming.
\newblock \emph{Journal of Machine Learning Research}, 13:\penalty0 398--427,
  2012.

\bibitem[Raskutti et~al.(2014)Raskutti, Wainwright, and Yu]{RasWaiYu14}
G.~Raskutti, M.~J. Wainwright, and B.~Yu.
\newblock Early stopping and non-parametric regression: An optimal
  data-dependent stopping rule.
\newblock \emph{Journal of Machine Learning Research}, 15:\penalty0 335--366,
  2014.

\bibitem[Rasmussen and Williams(2006)]{RasWil06}
C.~E. Rasmussen and C.~Williams.
\newblock \emph{Gaussian Processes for Machine Learning}.
\newblock MIT Press, 2006.

\bibitem[Ravi and Beatson(2019)]{ravi2019amortized}
Sachin Ravi and Alex Beatson.
\newblock Amortized bayesian meta-learning.
\newblock In \emph{ICLR (Poster)}, 2019.

\bibitem[Ravi and Larochelle(2017)]{ravi2017optimization}
Sachin Ravi and Hugo Larochelle.
\newblock Optimization as a model for few-shot learning.
\newblock \emph{International Conference on Learning Representations}, 2017.

\bibitem[Ravikumar et~al.(2011)Ravikumar, Wainwright, Raskutti, and
  Yu]{Ravetal11}
P.~Ravikumar, M.~J. Wainwright, G.~Raskutti, and B.~Yu.
\newblock High-dimensional covariance estimation by minimizing
  $\ell_1$-penalized log-determinant divergence.
\newblock \emph{Electronic Journal of Statistics}, 5:\penalty0 935--980, 2011.

\bibitem[Reddi et~al.(2018)Reddi, Zaheer, Sachan, Kale, and
  Kumar]{reddi2018adaptive}
S~Reddi, Manzil Zaheer, Devendra Sachan, Satyen Kale, and Sanjiv Kumar.
\newblock Adaptive methods for nonconvex optimization.
\newblock In \emph{Proceeding of 32nd Conference on Neural Information
  Processing Systems (NIPS 2018)}, 2018.

\bibitem[Reddi et~al.(2021)Reddi, Charles, Zaheer, Garrett, Rush,
  Kone{\v{c}}n{\'y}, Kumar, and McMahan]{fedadam}
Sashank~J. Reddi, Zachary Charles, Manzil Zaheer, Zachary Garrett, Keith Rush,
  Jakub Kone{\v{c}}n{\'y}, Sanjiv Kumar, and Hugh~Brendan McMahan.
\newblock Adaptive federated optimization.
\newblock In \emph{International Conference on Learning Representations}, 2021.
\newblock URL \url{https://openreview.net/forum?id=LkFG3lB13U5}.

\bibitem[Reisizadeh et~al.(2019)Reisizadeh, Taheri, Mokhtari, Hassani, and
  Pedarsani]{reisizadeh2019robust}
Amirhossein Reisizadeh, Hossein Taheri, Aryan Mokhtari, Hamed Hassani, and
  Ramtin Pedarsani.
\newblock Robust and communication-efficient collaborative learning.
\newblock In \emph{Advances in Neural Information Processing Systems}, pages
  8388--8399, 2019.

\bibitem[Rios and Sahinidis(2013)]{rios2013derivative}
Luis~Miguel Rios and Nikolaos~V Sahinidis.
\newblock Derivative-free optimization: a review of algorithms and comparison
  of software implementations.
\newblock \emph{Journal of Global Optimization}, 56\penalty0 (3):\penalty0
  1247--1293, 2013.

\bibitem[Rockwell(2021)]{rockwell}
Rockwell.
\newblock The connected enterprise.
\newblock
  \url{https://www.rockwellautomation.com/en-us/capabilities/connected-enterprise.html},
  2021.
\newblock Accessed: 2020-07-18.

\bibitem[Rusu et~al.(2019)Rusu, Rao, Sygnowski, Vinyals, Pascanu, Osindero, and
  Hadsell]{rusu2019meta}
Andrei~A Rusu, Dushyant Rao, Jakub Sygnowski, Oriol Vinyals, Razvan Pascanu,
  Simon Osindero, and Raia Hadsell.
\newblock Meta-learning with latent embedding optimization.
\newblock \emph{International Conference on Learning Representations}, 2019.

\bibitem[Ryan et~al.(2014)Ryan, Ela, Flynn, and OMalley]{ryan2014variable}
James Ryan, Erik Ela, Damian Flynn, and Mark OMalley.
\newblock Variable generation, reserves, flexibility and policy interactions.
\newblock In \emph{2014 47th Hawaii International Conference on System
  Sciences}, pages 2426--2434. IEEE, 2014.

\bibitem[RYU and BOYD(16)]{consensusoptimization}
ERNEST~K. RYU and STEPHEN BOYD.
\newblock Primer on monotone operator methods.
\newblock In \emph{Appl. Comput. Math.}, 16.
\newblock URL \url{https://stanford.edu/~boyd/papers/pdf/monotone_primer.pdf}.

\bibitem[Salimans et~al.(2017)Salimans, Ho, Chen, Sidor, and Sutskever]{ZO-RL}
Tim Salimans, Jonathan Ho, Xi~Chen, Szymon Sidor, and Ilya Sutskever.
\newblock Evolution strategies as a scalable alternative to reinforcement
  learning.
\newblock \emph{arXiv preprint arXiv:1703.03864}, 2017.

\bibitem[Salmi et~al.(2020)Salmi, Akmal, Pei, Wolff, Jaribion, and
  Khajavi]{salmi20203d}
Mika Salmi, Jan~Sher Akmal, Eujin Pei, Jan Wolff, Alireza Jaribion, and
  Siavash~H Khajavi.
\newblock 3d printing in covid-19: Productivity estimation of the most
  promising open source solutions in emergency situations.
\newblock \emph{Applied Sciences}, 10\penalty0 (11):\penalty0 4004, 2020.

\bibitem[Samarakoon et~al.(2019)Samarakoon, Bennis, Saad, and
  Debbah]{samarakoon2019distributed}
Sumudu Samarakoon, Mehdi Bennis, Walid Saad, and M{\'e}rouane Debbah.
\newblock Distributed federated learning for ultra-reliable low-latency
  vehicular communications.
\newblock \emph{IEEE Transactions on Communications}, 68\penalty0 (2):\penalty0
  1146--1159, 2019.

\bibitem[Samsung(2021)]{sumsung}
Samsung.
\newblock Samsung galaxy watch active 2.
\newblock
  \url{https://www.samsung.com/us/mobile/wearables/galaxy-watch-active-2/},
  2021.
\newblock Accessed: 2020-07-18.

\bibitem[Sannikov(2008)]{sannikov2008continuous}
Yuliy Sannikov.
\newblock A continuous-time version of the principal-agent problem.
\newblock \emph{Rev. Econ. Stud.}, 75\penalty0 (3):\penalty0 957--984, 2008.

\bibitem[Sarma et~al.(2021)Sarma, Harmon, Sanford, Roth, Xu, Tetreault, Xu,
  Flores, Raman, Kulkarni, et~al.]{sarma2021federated}
Karthik~V Sarma, Stephanie Harmon, Thomas Sanford, Holger~R Roth, Ziyue Xu,
  Jesse Tetreault, Daguang Xu, Mona~G Flores, Alex~G Raman, Rushikesh Kulkarni,
  et~al.
\newblock Federated learning improves site performance in multicenter deep
  learning without data sharing.
\newblock \emph{Journal of the American Medical Informatics Association}, 2021.

\bibitem[Sattler et~al.(2019)Sattler, Müller, and Samek]{clusterfed}
Felix Sattler, Klaus-Robert Müller, and Wojciech Samek.
\newblock Clustered federated learning: Model-agnostic distributed multi-task
  optimization under privacy constraints.
\newblock \emph{arXiv preprint arXiv:1910.01991}, 2019.

\bibitem[Scarabaggio et~al.(2021)Scarabaggio, Grammatico, Carli, and
  Dotoli]{Scarabaggio2021}
Paolo Scarabaggio, Sergio Grammatico, Raffaele Carli, and Mariagrazia Dotoli.
\newblock Distributed demand side management with stochastic wind power
  forecasting.
\newblock \emph{IEEE Transactions on Control Systems Technology}, pages 1--16,
  2021.

\bibitem[Schmidt et~al.(2015)Schmidt, Roux, and Bach.]{sag}
Mark Schmidt, Nicolas~Le Roux, and Francis Bach.
\newblock Minimizing finite sums with the stochastic average gradient.
\newblock \emph{arXiv preprint arXiv:1309.2388}, 2015.

\bibitem[Seward et~al.(2018)Seward, Unterthiner, Bergmann, Jetchev, and
  Hochreiter]{gantraining2020}
Calvin Seward, Thomas Unterthiner, Urs Bergmann, Nikolay Jetchev, and Sepp
  Hochreiter.
\newblock First order generative adversarial networks.
\newblock \emph{International Conference on Machine Learning}, 2018.

\bibitem[Shamir et~al.(2014)Shamir, Srebro, and Zhang]{dane}
Ohad Shamir, Nati Srebro, and Tong Zhang.
\newblock Communication-efficient distributed optimization using an approximate
  newton-type method.
\newblock In Eric~P. Xing and Tony Jebara, editors, \emph{Proceedings of the
  31st International Conference on Machine Learning}, volume~32 of
  \emph{Proceedings of Machine Learning Research}, pages 1000--1008. PMLR,
  2014.

\bibitem[Sheller et~al.(2020)Sheller, Edwards, Reina, Martin, Pati, Kotrotsou,
  Milchenko, Xu, Marcus, Colen, et~al.]{sheller2020federated}
Micah~J Sheller, Brandon Edwards, G~Anthony Reina, Jason Martin, Sarthak Pati,
  Aikaterini Kotrotsou, Mikhail Milchenko, Weilin Xu, Daniel Marcus, Rivka~R
  Colen, et~al.
\newblock Federated learning in medicine: facilitating multi-institutional
  collaborations without sharing patient data.
\newblock \emph{Scientific reports}, 10\penalty0 (1):\penalty0 1--12, 2020.

\bibitem[Shi et~al.(2021)Shi, Lai, Kontar, and Chowdhury]{Fedensemble}
Naichen Shi, Fan Lai, Raed~Al Kontar, and Mosharaf Chowdhury.
\newblock Fed-ensemble: Ensemble models in federated learning for improved
  generalization and uncertainty quantification.
\newblock In \emph{Arxiv:2107.10663}, 2021.

\bibitem[Singh et~al.(2019)Singh, Vepakomma, Gupta, and
  Raskar]{singh2019detailed}
Abhishek Singh, Praneeth Vepakomma, Otkrist Gupta, and Ramesh Raskar.
\newblock Detailed comparison of communication efficiency of split learning and
  federated learning.
\newblock \emph{arXiv preprint arXiv:1909.09145}, 2019.

\bibitem[Singh et~al.(2020)Singh, Valley, Tang, Li, Kamran, Sjoding, Wiens,
  Otles, Donnelly, Wei, et~al.]{singh2020evaluating}
Karandeep Singh, Thomas~S Valley, Shengpu Tang, Benjamin~Y Li, Fahad Kamran,
  Michael~W Sjoding, Jenna Wiens, Erkin Otles, John~P Donnelly, Melissa~Y Wei,
  et~al.
\newblock Evaluating a widely implemented proprietary deterioration index model
  among hospitalized covid-19 patients.
\newblock \emph{Annals of the American Thoracic Society}, \penalty0 (ja), 2020.

\bibitem[Smith et~al.(2017{\natexlab{a}})Smith, Chiang, Sanjabi, and
  Talwalkar]{mocha}
Virginia Smith, Chao-Kai Chiang, Maziar Sanjabi, and Ameet Talwalkar.
\newblock Federated multi-task learning.
\newblock In \emph{31st Conference on Neural Information Processing Systems},
  2017{\natexlab{a}}.

\bibitem[Smith et~al.(2017{\natexlab{b}})Smith, Chiang, Sanjabi, and
  Talwalkar]{multitask-fl}
Virginia Smith, Chao-Kai Chiang, Maziar Sanjabi, and Ameet Talwalkar.
\newblock Federated multi-task learning.
\newblock \emph{NeurIPS}, 2017{\natexlab{b}}.

\bibitem[Smith et~al.(2017{\natexlab{c}})Smith, Chiang, Sanjabi, and
  Talwalkar]{smith2017federated}
Virginia Smith, Chao-Kai Chiang, Maziar Sanjabi, and Ameet Talwalkar.
\newblock Federated multi-task learning.
\newblock \emph{Conference on Neural Information Processing Systems},
  2017{\natexlab{c}}.

\bibitem[Snell et~al.(2017)Snell, Swersky, and Zemel]{snell2017prototypical}
Jake Snell, Kevin Swersky, and Richard~S Zemel.
\newblock Prototypical networks for few-shot learning.
\newblock \emph{arXiv preprint arXiv:1703.05175}, 2017.

\bibitem[Spear and Srivastava(1987)]{spear1987repeated}
Stephen~E Spear and Sanjay Srivastava.
\newblock On repeated moral hazard with discounting.
\newblock \emph{Rev. Econ. Stud.}, 54\penalty0 (4):\penalty0 599--617, 1987.

\bibitem[Srai et~al.(2016)Srai, Kumar, Graham, Phillips, Tooze, Ford, Beecher,
  Raj, Gregory, Tiwari, et~al.]{srai2016distributed}
Jagjit~Singh Srai, Mukesh Kumar, Gary Graham, Wendy Phillips, James Tooze,
  Simon Ford, Paul Beecher, Baldev Raj, Mike Gregory, Manoj~Kumar Tiwari,
  et~al.
\newblock Distributed manufacturing: scope, challenges and opportunities.
\newblock \emph{International Journal of Production Research}, 54\penalty0
  (23):\penalty0 6917--6935, 2016.

\bibitem[Stevens(2020)]{defectivemasks}
Robest Stevens.
\newblock Why diy 3d-printed face masks and shields are so risky.
\newblock
  \url{https://slate.com/technology/2020/04/diy-3d-printed-face-masks-shields-coronavirus.html},
  2020.
\newblock Accessed: 2020-07-18.

\bibitem[Su(2020)]{faultymasks}
Alice Su.
\newblock Faulty masks. flawed tests. china’s quality control problem in
  leading global covid-19 fight.
\newblock
  \url{https://www.latimes.com/world-nation/story/2020-04-10/china-beijing-supply-world-coronavirus-fight-quality-control},
  2020.
\newblock Accessed: 2020-07-18.

\bibitem[Sugiyama and
  Kawanabe(2012)]{machinelearninginnonstationaryenvironments}
Masashi Sugiyama and Motoaki Kawanabe.
\newblock \emph{Machine Learning in Non-Stationary Environments}.
\newblock MIT Press, 2012.

\bibitem[Sun et~al.(2019)Sun, Qiu, Xu, and Huang]{finetunebert}
Chi Sun, Xipeng Qiu, Yige Xu, and Xuanjing Huang.
\newblock How to fine-tune bert for text classification?
\newblock \emph{Chinese Computational Linguistics}, 2019.

\bibitem[Tafreshian and Masoud(2020)]{partitioning_1}
Amirmahdi Tafreshian and Neda Masoud.
\newblock Trip-based graph partitioning in dynamic ridesharing.
\newblock \emph{Transportation Research Part C: Emerging Technologies},
  114:\penalty0 532--553, 2020.

\bibitem[Tang et~al.(2019)Tang, Yu, Lian, Zhang, and
  Liu]{tang2019doublesqueeze}
Hanlin Tang, Chen Yu, Xiangru Lian, Tong Zhang, and Ji~Liu.
\newblock Doublesqueeze: Parallel stochastic gradient descent with double-pass
  error-compensated compression.
\newblock In \emph{International Conference on Machine Learning}, pages
  6155--6165, 2019.

\bibitem[Tossou et~al.(2019)Tossou, Dura, Laviolette, Marchand, and
  Lacoste]{tossou2019adaptive}
Prudencio Tossou, Basile Dura, Francois Laviolette, Mario Marchand, and
  Alexandre Lacoste.
\newblock Adaptive deep kernel learning.
\newblock \emph{arXiv preprint arXiv:1905.12131}, 2019.

\bibitem[Valadarsky et~al.(2017)Valadarsky, Schapira, Shahaf, and
  Tamar]{learning-routes}
Asaf Valadarsky, Michael Schapira, Dafna Shahaf, and Aviv Tamar.
\newblock Learning to route.
\newblock In \emph{ACM HotNets}, 2017.

\bibitem[Vanschoren(2018)]{vanschoren2018meta}
Joaquin Vanschoren.
\newblock Meta-learning: A survey.
\newblock \emph{arXiv preprint arXiv:1810.03548}, 2018.

\bibitem[Vaswani et~al.(2017)Vaswani, Shazeer, Parmar, Uszkoreit, Jones, Gomez,
  Kaiser, and Polosukhin]{attention}
Ashish Vaswani, Noam Shazeer, Niki Parmar, Jakob Uszkoreit, Llion Jones,
  Aidan~N Gomez, \L~ukasz Kaiser, and Illia Polosukhin.
\newblock Attention is all you need.
\newblock In I.~Guyon, U.~V. Luxburg, S.~Bengio, H.~Wallach, R.~Fergus,
  S.~Vishwanathan, and R.~Garnett, editors, \emph{Advances in Neural
  Information Processing Systems}, volume~30. Curran Associates, Inc., 2017.
\newblock URL
  \url{https://proceedings.neurips.cc/paper/2017/file/3f5ee243547dee91fbd053c1c4a845aa-Paper.pdf}.

\bibitem[Walter(2019)]{futurefl}
De~Brouwer Walter.
\newblock The federated future is ready for shipping.
\newblock
  \url{https://medium.com/@_doc_ai/the-federated-future-is-ready-for-shipping-d17ff40f43e3},
  2019.
\newblock Accessed: 2021-04-21.

\bibitem[Wang et~al.(2020)Wang, Yurochkin, Sun, Papailiopoulos, and
  Khazaeni]{wang2020federated}
Hongyi Wang, Mikhail Yurochkin, Yuekai Sun, Dimitris Papailiopoulos, and
  Yasaman Khazaeni.
\newblock Federated learning with matched averaging.
\newblock \emph{International Conference on Learning Representation}, 2020.

\bibitem[Wang et~al.(2021{\natexlab{a}})Wang, Chung, AlShelahi, Kontar, Byon,
  and Saigal]{wang2021look}
Jingxing Wang, Seokhyun Chung, Abdullah AlShelahi, Raed Kontar, Eunshin Byon,
  and Romesh Saigal.
\newblock Look-ahead planning for renewable energy: A dynamic ``predict and
  store'' approach.
\newblock \emph{submitted}, 2021{\natexlab{a}}.

\bibitem[Wang et~al.(2021{\natexlab{b}})Wang, Li, and
  Tsung]{wang2021distribution}
Kai Wang, Jian Li, and Fugee Tsung.
\newblock Distribution inference from early-stage stationary data streams by
  transfer learning.
\newblock \emph{IISE Transactions}, pages 1--25, 2021{\natexlab{b}}.

\bibitem[Wang et~al.(2019{\natexlab{a}})Wang, Mathews, Kiddon, Eichner,
  Beaufays, and Ramage]{fedper}
Kangkang Wang, Rajiv Mathews, Chloe Kiddon, Hubert Eichner, Francoise Beaufays,
  and Daniel Ramage.
\newblock Federated learning with personalization layers.
\newblock \emph{arXiv preprint arXiv:1910.10252}, 2019{\natexlab{a}}.

\bibitem[Wang et~al.(2019{\natexlab{b}})Wang, Mathews, Kiddon, Eichner,
  Beaufays, and Ramage]{wang2019federated}
Kangkang Wang, Rajiv Mathews, Chlo{\'e} Kiddon, Hubert Eichner, Fran{\c{c}}oise
  Beaufays, and Daniel Ramage.
\newblock Federated evaluation of on-device personalization.
\newblock \emph{arXiv preprint arXiv:1910.10252}, 2019{\natexlab{b}}.

\bibitem[Wang et~al.(2019{\natexlab{c}})Wang, Han, Wang, Zhao, Chen, and
  Chen]{wang2019edge}
Xiaofei Wang, Yiwen Han, Chenyang Wang, Qiyang Zhao, Xu~Chen, and Min Chen.
\newblock In-edge ai: Intelligentizing mobile edge computing, caching and
  communication by federated learning.
\newblock \emph{IEEE Network}, 33\penalty0 (5):\penalty0 156--165,
  2019{\natexlab{c}}.

\bibitem[Wang et~al.(2019{\natexlab{d}})Wang, Zhao, Yu, Zhang, and
  Chen]{wang2019bayesian}
Zhenyi Wang, Yang Zhao, Ping Yu, Ruiyi Zhang, and Changyou Chen.
\newblock Bayesian meta sampling for fast uncertainty adaptation.
\newblock In \emph{International Conference on Learning Representations},
  2019{\natexlab{d}}.

\bibitem[Weber(2015)]{weber2015federated}
Griffin~M Weber.
\newblock Federated queries of clinical data repositories: scaling to a
  national network.
\newblock \emph{Journal of biomedical informatics}, 55:\penalty0 231--236,
  2015.

\bibitem[Weber et~al.(2009)Weber, Murphy, McMurry, MacFadden, Nigrin,
  Churchill, and Kohane]{weber2009shared}
Griffin~M Weber, Shawn~N Murphy, Andrew~J McMurry, Douglas MacFadden, Daniel~J
  Nigrin, Susanne Churchill, and Isaac~S Kohane.
\newblock The shared health research information network (shrine): a prototype
  federated query tool for clinical data repositories.
\newblock \emph{Journal of the American Medical Informatics Association},
  16\penalty0 (5):\penalty0 624--630, 2009.

\bibitem[Wei et~al.(2019{\natexlab{a}})Wei, Xu, Zhang, Zheng, Zang, Chen,
  Zhang, Zhu, Xu, and Li]{Intersection_1}
Hua Wei, Nan Xu, Huichu Zhang, Guanjie Zheng, Xinshi Zang, Chacha Chen, Weinan
  Zhang, Yanmin Zhu, Kai Xu, and Zhenhui Li.
\newblock Colight: Learning network-level cooperation for traffic signal
  control.
\newblock In \emph{Proceedings of the 28th ACM International Conference on
  Information and Knowledge Management}, pages 1913--1922, 2019{\natexlab{a}}.

\bibitem[Wei et~al.(2019{\natexlab{b}})Wei, Li, Ding, Ma, Yang, Farhad, Jin,
  Quek, and Poor]{differentialprivacyreview}
Kang Wei, Jun Li, Ming Ding, Chuan Ma, Howard~H. Yang, Farokhi Farhad, Shi Jin,
  Tony Q.~S. Quek, and H.~Vincent Poor.
\newblock Federated learning with differential privacy: Algorithms and
  performance analysis.
\newblock \emph{arXiv preprint arXiv:1911.00222}, 2019{\natexlab{b}}.

\bibitem[Wei et~al.(2018)Wei, Liu, and Tang]{wei2018reliability}
Pengfei Wei, Fuchao Liu, and Chenghu Tang.
\newblock Reliability and reliability-based importance analysis of structural
  systems using multiple response gaussian process model.
\newblock \emph{Reliability Engineering \& System Safety}, 175:\penalty0
  183--195, 2018.

\bibitem[Wellener et~al.(2019)]{wellener2019deloitte}
Paul Wellener et~al.
\newblock Deloitte and mapi smart factory study: capturing value through the
  digital journey.
\newblock \emph{Deloitte Insights and MAPI, Deloitte, USA}, 2019.

\bibitem[Welling and Teh(2011)]{welling2011bayesian}
Max Welling and Yee~W Teh.
\newblock Bayesian learning via stochastic gradient langevin dynamics.
\newblock In \emph{Proceedings of the 28th international conference on machine
  learning (ICML-11)}, pages 681--688, 2011.

\bibitem[Wessler et~al.(2021)Wessler, Nelson, Park, McGinnes, Gulati, Brazil,
  Van~Calster, van Klaveren, Venema, Steyerberg, et~al.]{wessler2021external}
Benjamin~S Wessler, Jason Nelson, Jinny~G Park, Hannah McGinnes, Gaurav Gulati,
  Riley Brazil, Ben Van~Calster, David van Klaveren, Esmee Venema, Ewout
  Steyerberg, et~al.
\newblock External validations of cardiovascular clinical prediction models: A
  large-scale review of the literature.
\newblock \emph{medRxiv}, 2021.

\bibitem[Williams and Rasmussen(2006)]{williams2006gaussian}
Christopher~K Williams and Carl~Edward Rasmussen.
\newblock \emph{Gaussian processes for machine learning}, volume~2.
\newblock MIT press Cambridge, MA, 2006.

\bibitem[Winstein and Balakrishnan(2013)]{remy}
Keith Winstein and Hari Balakrishnan.
\newblock {TCP} ex machina: Computer-generated congestion control.
\newblock In \emph{ACM SIGCOMM}, 2013.

\bibitem[Wolff et~al.(2019)Wolff, Moons, Riley, Whiting, Westwood, Collins,
  Reitsma, Kleijnen, and Mallett]{wolff2019probast}
Robert~F Wolff, Karel~GM Moons, Richard~D Riley, Penny~F Whiting, Marie
  Westwood, Gary~S Collins, Johannes~B Reitsma, Jos Kleijnen, and Sue Mallett.
\newblock Probast: a tool to assess the risk of bias and applicability of
  prediction model studies.
\newblock \emph{Annals of internal medicine}, 170\penalty0 (1):\penalty0
  51--58, 2019.

\bibitem[Wong(2019)]{aichipsbillion}
Christina Wong.
\newblock Ai chips for self driving cars will a be 10 billion market by 2024.
\newblock
  \url{https://www.nextbigfuture.com/2019/03/ai-chips-for-self-driving-cars-will-a-be-10-billion-market-by-2024.html},
  2019.
\newblock Accessed: 2020-07-18.

\bibitem[Woodall and Del~Castillo(2014)]{woodall2014overview}
William~H Woodall and Enrique Del~Castillo.
\newblock An overview of george box's contributions to process monitoring and
  feedback adjustment.
\newblock \emph{Applied Stochastic Models in Business and Industry},
  30\penalty0 (1):\penalty0 53--61, 2014.

\bibitem[Wu and Hamada(2011)]{wu2011experiments}
CF~Jeff Wu and Michael~S Hamada.
\newblock \emph{Experiments: planning, analysis, and optimization}, volume 552.
\newblock John Wiley \& Sons, 2011.

\bibitem[Wu and Wang(2020)]{clientsamplewiththeta}
Hongda Wu and Ping Wang.
\newblock Fast-convergent federated learning with adaptive weighting.
\newblock \emph{arXiv preprint arXiv:2012.00661}, 2020.

\bibitem[Wu et~al.(2018)Wu, Peurifoy, Chuang, and Tegmark]{wu2018meta}
Tailin Wu, John Peurifoy, Isaac~L Chuang, and Max Tegmark.
\newblock Meta-learning autoencoders for few-shot prediction.
\newblock \emph{arXiv preprint arXiv:1807.09912}, 2018.

\bibitem[Xiong et~al.(2009)Xiong, Xiong, Chen, and Yang]{xiong2009optimizing}
F~Xiong, Y~Xiong, W~Chen, and S~Yang.
\newblock Optimizing latin hypercube design for sequential sampling of computer
  experiments.
\newblock \emph{Engineering Optimization}, 41\penalty0 (8):\penalty0 793--810,
  2009.

\bibitem[Xu and Wang(2020)]{xu2020client}
Jie Xu and Heqiang Wang.
\newblock Client selection and bandwidth allocation in wireless federated
  learning networks: A long-term perspective.
\newblock \emph{IEEE Transactions on Wireless Communications}, 2020.

\bibitem[Yampikulsakul et~al.(2014)Yampikulsakul, Byon, Huang, Sheng, and
  You]{yampikulsakul2014condition}
Nattavut Yampikulsakul, Eunshin Byon, Shuai Huang, Shuangwen Sheng, and Mingdi
  You.
\newblock Condition monitoring of wind power system with nonparametric
  regression analysis.
\newblock \emph{IEEE Transactions on Energy Conversion}, 29\penalty0
  (2):\penalty0 288--299, 2014.

\bibitem[Yan et~al.(2020)Yan, Ayers, and et~al.]{puffer}
Francis~Y. Yan, Hudson Ayers, and et~al.
\newblock Learning in situ: a randomized experiment in video streaming.
\newblock In \emph{USENIX NSDI}, 2020.

\bibitem[Yang et~al.(2019{\natexlab{a}})Yang, Liu, Chen, and
  Tong]{yang2019federated}
Qiang Yang, Yang Liu, Tianjian Chen, and Yongxin Tong.
\newblock Federated machine learning: Concept and applications.
\newblock \emph{ACM Transactions on Intelligent Systems and Technology (TIST)},
  10\penalty0 (2):\penalty0 1--19, 2019{\natexlab{a}}.

\bibitem[Yang et~al.(2019{\natexlab{b}})Yang, Ren, Zhou, and
  Liu]{yang2019parallel}
Shengwen Yang, Bing Ren, Xuhui Zhou, and Liping Liu.
\newblock Parallel distributed logistic regression for vertical federated
  learning without third-party coordinator.
\newblock \emph{arXiv preprint arXiv:1911.09824}, 2019{\natexlab{b}}.

\bibitem[Yang et~al.(2018{\natexlab{a}})Yang, Andrew, Eichner, Sun, Li, Kong,
  Ramage, and Beaufays]{ggkeyboard}
Timothy Yang, Galen Andrew, Hubert Eichner, Haicheng Sun, Wei Li, Nicholas
  Kong, Daniel Ramage, and Fran{\c{c}}oise Beaufays.
\newblock Applied federated learning: Improving {Google} keyboard query
  suggestions.
\newblock In \emph{arxiv.org/abs/1812.02903}, 2018{\natexlab{a}}.

\bibitem[Yang et~al.(2018{\natexlab{b}})Yang, Andrew, Eichner, Sun, Li, Kong,
  Ramage, and Beaufays]{yang2018applied}
Timothy Yang, Galen Andrew, Hubert Eichner, Haicheng Sun, Wei Li, Nicholas
  Kong, Daniel Ramage, and Fran{\c{c}}oise Beaufays.
\newblock Applied federated learning: Improving google keyboard query
  suggestions.
\newblock \emph{arXiv preprint arXiv:1812.02903}, 2018{\natexlab{b}}.

\bibitem[Yazıcı et~al.(2019)Yazıcı, Foo, Winkler, Yap, Piliouras, and
  Chandrasekhar]{gantraining2019}
Yasin Yazıcı, Chuan-Sheng Foo, Stefan Winkler, Kim-Hui Yap, Georgios
  Piliouras, and Vijay Chandrasekhar.
\newblock The unusual effectiveness of averaging in gan training.
\newblock \emph{International Conference on Learning Representations}, 2019.

\bibitem[Ye et~al.(2013)Ye, Hong, and Xie]{ye2013heterogeneities}
Zhi-Sheng Ye, Yili Hong, and Yimeng Xie.
\newblock How do heterogeneities in operating environments affect field failure
  predictions and test planning?
\newblock \emph{The Annals of Applied Statistics}, 7:\penalty0 2249--2271,
  2013.

\bibitem[Yoon et~al.(2018)Yoon, Kim, Dia, Kim, Bengio, and
  Ahn]{yoon2018bayesian}
Jaesik Yoon, Taesup Kim, Ousmane Dia, Sungwoong Kim, Yoshua Bengio, and Sungjin
  Ahn.
\newblock Bayesian model-agnostic meta-learning.
\newblock In \emph{Proceedings of the 32nd International Conference on Neural
  Information Processing Systems}, pages 7343--7353, 2018.

\bibitem[You et~al.(2017)You, Byon, Jin, and Lee]{you2017wind}
Mingdi You, Eunshin Byon, Jionghua Jin, and Giwhyun Lee.
\newblock When wind travels through turbines: A new statistical approach for
  characterizing heterogeneous wake effects in multi-turbine wind farms.
\newblock \emph{IISE Transactions}, 49\penalty0 (1):\penalty0 84--95, 2017.

\bibitem[Yu et~al.(2020)Yu, Bagdasaryan, and Shmatikov]{local-adaptation}
Tao Yu, Eugene Bagdasaryan, and Vitaly Shmatikov.
\newblock Salvaging federated learning by local adaptation.
\newblock \emph{arXiv preprint arXiv:2002.04758}, 2020.

\bibitem[Yuan and Ma(2020)]{fedac}
Honglin Yuan and Tengyu Ma.
\newblock Federated accelerated stochastic gradient descent.
\newblock In \emph{34th Conference on Neural Information Processing Systems},
  2020.
\newblock URL
  \url{https://papers.nips.cc/paper/2020/file/39d0a8908fbe6c18039ea8227f827023-Supplemental.pdf}.

\bibitem[Yuan and Lin(2007)]{YuaLin07}
M.~Yuan and Y.~Lin.
\newblock Model selection and estimation in the {G}aussian graphical model.
\newblock \emph{Biometrika}, 94\penalty0 (1):\penalty0 19--35, 2007.

\bibitem[Yuan et~al.(2012)Yuan, Liu, and Yan]{yuan2012visual}
Xiao-Tong Yuan, Xiaobai Liu, and Shuicheng Yan.
\newblock Visual classification with multitask joint sparse representation.
\newblock \emph{IEEE Transactions on Image Processing}, 21\penalty0
  (10):\penalty0 4349--4360, 2012.

\bibitem[Yue and Kontar(2019{\natexlab{a}})]{yue2019renyi}
Xubo Yue and Raed Kontar.
\newblock The renyi gaussian process: Towards improved generalization.
\newblock \emph{arXiv preprint arXiv:1910.06990}, 2019{\natexlab{a}}.

\bibitem[Yue and Kontar(2019{\natexlab{b}})]{yue2019variational}
Xubo Yue and Raed Kontar.
\newblock Variational inference of joint models using multivariate gaussian
  convolution processes.
\newblock \emph{arXiv preprint arXiv:1903.03867}, 2019{\natexlab{b}}.

\bibitem[Yue et~al.(2021)Yue, Nouiehed, and Kontar]{yue2021gifair}
Xubo Yue, Maher Nouiehed, and Raed~Al Kontar.
\newblock Gifair-fl: An approach for group and individual fairness in federated
  learning.
\newblock \emph{arXiv preprint arXiv:2108.02741}, 2021.

\bibitem[Yurochkin et~al.(2019)Yurochkin, Agarwal, Ghosh, Greenewald, Hoang,
  and Khazaeni]{yurochkin2019bayesian}
Mikhail Yurochkin, Mayank Agarwal, Soumya Ghosh, Kristjan Greenewald, Nghia
  Hoang, and Yasaman Khazaeni.
\newblock Bayesian nonparametric federated learning of neural networks.
\newblock In \emph{International Conference on Machine Learning}, pages
  7252--7261. PMLR, 2019.

\bibitem[Zafar et~al.(2017)Zafar, Valera, Rogriguez, and
  Gummadi]{zafar2017fairness}
Muhammad~Bilal Zafar, Isabel Valera, Manuel~Gomez Rogriguez, and Krishna~P
  Gummadi.
\newblock Fairness constraints: Mechanisms for fair classification.
\newblock In \emph{Artificial Intelligence and Statistics}, pages 962--970.
  PMLR, 2017.

\bibitem[Zeng et~al.(2021)Zeng, Chen, and Lee]{zeng2021improving}
Yuchen Zeng, Hongxu Chen, and Kangwook Lee.
\newblock Improving fairness via federated learning.
\newblock \emph{arXiv preprint arXiv:2110.15545}, 2021.

\bibitem[Zhang et~al.(2018{\natexlab{a}})Zhang, Jin, Ratnasamy, Wawrzynek, and
  Lee]{awstream}
Ben Zhang, Xin Jin, Sylvia Ratnasamy, John Wawrzynek, and Edward~A. Lee.
\newblock {AWS}tream: Adaptive wide-area streaming analytics.
\newblock In \emph{ACM SIGCOMM}, 2018{\natexlab{a}}.

\bibitem[Zhang et~al.(2018{\natexlab{b}})Zhang, B{\"u}tepage, Kjellstr{\"o}m,
  and Mandt]{zhang2018advances}
Cheng Zhang, Judith B{\"u}tepage, Hedvig Kjellstr{\"o}m, and Stephan Mandt.
\newblock Advances in variational inference.
\newblock \emph{IEEE transactions on pattern analysis and machine
  intelligence}, 41\penalty0 (8):\penalty0 2008--2026, 2018{\natexlab{b}}.

\bibitem[Zhang et~al.(2020{\natexlab{a}})Zhang, Kou, and Wang]{zhang2020fairfl}
Daniel~Yue Zhang, Ziyi Kou, and Dong Wang.
\newblock Fairfl: A fair federated learning approach to reducing demographic
  bias in privacy-sensitive classification models.
\newblock In \emph{2020 IEEE International Conference on Big Data (Big Data)},
  pages 1051--1060. IEEE, 2020{\natexlab{a}}.

\bibitem[Zhang et~al.(2018{\natexlab{c}})Zhang, Chan, and
  Zhou]{zhang2018enabling}
Ding Zhang, Ching~Chuen Chan, and George~You Zhou.
\newblock Enabling industrial internet of things (iiot) towards an emerging
  smart energy system.
\newblock \emph{Global energy interconnection}, 1\penalty0 (1):\penalty0
  39--47, 2018{\natexlab{c}}.

\bibitem[Zhang(2019)]{FLfuture}
Mi~Zhang.
\newblock Federated learning: The future of distributed machine learning.
\newblock
  \url{https://medium.com/syncedreview/federated-learning-the-future-of-distributed-machine-learning-eec95242d897},
  2019.
\newblock Accessed: 2020-07-18.

\bibitem[Zhang and Hong(2021)]{equivalencefeddynandfedpd}
Xinwei Zhang and Mingyi Hong.
\newblock On the connection between feddyn and fedpd, 2021.
\newblock URL \url{http://people.ece.umn.edu/~mhong/FedDyn_FedPD.pdf}.

\bibitem[Zhang et~al.(2020{\natexlab{b}})Zhang, Hong, Dhople, Yin, and
  Liu]{zhang2020fedpd}
Xinwei Zhang, Mingyi Hong, Sairaj Dhople, Wotao Yin, and Yang Liu.
\newblock Fedpd: A federated learning framework with optimal rates and
  adaptivity to non-iid data, 2020{\natexlab{b}}.

\bibitem[Zhang and Yeung(2012)]{zhang2012convex}
Yu~Zhang and Dit-Yan Yeung.
\newblock A convex formulation for learning task relationships in multi-task
  learning.
\newblock \emph{arXiv preprint arXiv:1203.3536}, 2012.

\bibitem[Zhao et~al.(2018{\natexlab{a}})Zhao, Yin, An, Wang, and
  Feng]{decomposition_2}
Meng Zhao, Jiateng Yin, Shi An, Jian Wang, and Dejian Feng.
\newblock Ridesharing problem with flexible pickup and delivery locations for
  app-based transportation service: mathematical modeling and decomposition
  methods.
\newblock \emph{Journal of Advanced Transportation}, 2018, 2018{\natexlab{a}}.

\bibitem[Zhao et~al.(2018{\natexlab{b}})Zhao, Li, Lai, Suda, Civin, and
  Chandra]{zhao2018federated}
Yue Zhao, Meng Li, Liangzhen Lai, Naveen Suda, Damon Civin, and Vikas Chandra.
\newblock Federated learning with non-iid data.
\newblock \emph{arXiv preprint arXiv:1806.00582}, 2018{\natexlab{b}}.

\bibitem[Zhou et~al.(2016)Zhou, Yang, and Shao]{zhou2016energy}
Kaile Zhou, Shanlin Yang, and Zhen Shao.
\newblock Energy internet: the business perspective.
\newblock \emph{Applied Energy}, 178:\penalty0 212--222, 2016.

\bibitem[Zhu et~al.(2019)Zhu, Liu, and Han]{gradientleakage}
Ligeng Zhu, Zhijian Liu, and Song Han.
\newblock Deep leakage from gradients.
\newblock \emph{33rd Conference on Neural Information Processing Systems},
  2019.

\bibitem[Zimmeman et~al.(2006)Zimmeman, Kramer, McNair, and
  Malila]{zimmeman2006acute}
JE~Zimmeman, AA~Kramer, DS~McNair, and FM~Malila.
\newblock Acute physiology and chronic health evaluations (apache) iv: hospital
  mortality assessment for today’s critically ill patients.
\newblock \emph{Crit Care Med}, 34\penalty0 (5):\penalty0 1297--1310, 2006.

\bibitem[Zinkevich et~al.(2010)Zinkevich, Weimer, Li, and Smola]{parallelsgd}
Martin Zinkevich, Markus Weimer, Lihong Li, and Alex Smola.
\newblock Parallelized stochastic gradient descent.
\newblock In J.~Lafferty, C.~Williams, J.~Shawe-Taylor, R.~Zemel, and
  A.~Culotta, editors, \emph{Advances in Neural Information Processing
  Systems}, volume~23. Curran Associates, Inc., 2010.
\newblock URL
  \url{https://proceedings.neurips.cc/paper/2010/file/abea47ba24142ed16b7d8fbf2c740e0d-Paper.pdf}.

\bibitem[Zou et~al.(2016)Zou, Chen, Xia, He, and Kang]{Zou2016}
Peng Zou, Qixin Chen, Qing Xia, Guannan He, and Chongqing Kang.
\newblock Evaluating the contribution of energy storages to support large-scale
  renewable generation in joint energy and ancillary service markets.
\newblock \emph{IEEE Transactions on Sustainable Energy}, 7\penalty0
  (2):\penalty0 808--818, 2016.

\bibitem[Zou and Lu(2020)]{zou2020gradient}
Yayi Zou and Xiaoqi Lu.
\newblock Gradient-em bayesian meta-learning.
\newblock \emph{Advances in Neural Information Processing Systems}, 2020.

\end{thebibliography}
\EOD

\end{document}